\pdfoutput=1
\documentclass{article}

\usepackage[margin=1in]{geometry}  
\usepackage{mathpazo}
\usepackage[backend=biber,style=alphabetic,natbib=true]{biblatex}
\usepackage{amsmath}
\usepackage{array}
\usepackage{enumerate}
\usepackage{amsthm}
\usepackage[textsize=scriptsize]{todonotes}

\usepackage[utf8]{inputenc} 
\usepackage[T1]{fontenc}    
\usepackage{hyperref}       
\usepackage{url}            
\usepackage{booktabs}       
\usepackage{amsfonts}       
\usepackage{nicefrac}       
\usepackage{microtype}      
\usepackage{subcaption}
\usepackage{graphicx}
\usepackage{multirow}
\usepackage{dsfont}
\usepackage{algorithm}
\usepackage[algo2e]{algorithm2e} 

\usepackage{longtable}

\usepackage{graphbox}

\renewcommand{\epsilon}{\varepsilon}

\newcommand{\contains}{\textsc{Contains}}
\newcommand{\classify}{\textsc{Classify}}
\newcommand{\task}{\text{object recognition}}

\title{From ImageNet to Image Classification:\\
	Contextualizing Progress on Benchmarks}

\newcommand\AND{
\end{tabular}\hfil\linebreak[4]\hfil%
\begin{tabular}[t]{c}\ignorespaces%
}
\author{
    Dimitris Tsipras\thanks{Equal contribution.} \\
	MIT \\
	\texttt{tsipras@mit.edu} 
	\and
    Shibani Santurkar\footnotemark[1] \\
	MIT \\
	\texttt{shibani@mit.edu} \\
	\and
	Logan Engstrom\\ 
	MIT \\
	\texttt{engstrom@mit.edu} \\
	\AND
	Andrew Ilyas\\
	MIT \\
	\texttt{ailyas@mit.edu} \\
	\and 
	Aleksander M\k{a}dry \\
	MIT \\
	\texttt{madry@mit.edu}  
}
\date{}

\usepackage[normalem]{ulem}
\usepackage{xcolor}

\newcommand{\scrcmd}{\textsuperscript}
\newcommand{\scr}[1]{\scrcmd{#1}}

\begin{document}
\maketitle

\begin{abstract}
    Building rich machine learning datasets in a scalable manner often 
necessitates a crowd-sourced data collection pipeline.
In this work, we use human studies to investigate the
consequences of employing such a pipeline, focusing on
the popular ImageNet dataset.
We study how specific design choices in
the ImageNet creation process impact 
the fidelity of the resulting dataset---including 
the introduction of biases that state-of-the-art models exploit. 
Our analysis pinpoints how a noisy data collection pipeline can lead to a 
{systematic} misalignment between the resulting benchmark and the
real-world task it serves as a proxy for.
Finally, our findings emphasize the need to augment our current model training
and evaluation toolkit  to take such misalignments into account.
\footnote{To facilitate further research, we release our refined ImageNet
	annotations at \url{https://github.com/MadryLab/ImageNetMultiLabel}.}

\end{abstract}

\section{Introduction}
\label{sec:intro}
\begin{figure}[!t]
	\begin{subfigure}[t]{0.245\textwidth}
		\centering
		\includegraphics[width=1\textwidth]{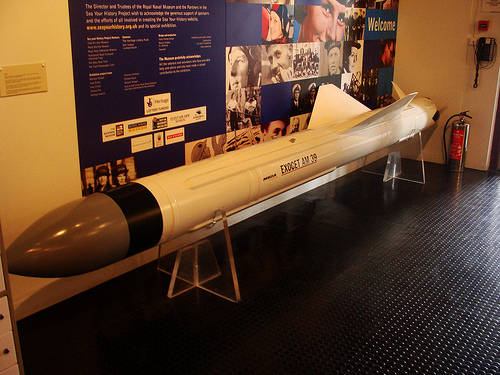}
		\caption{missile}
	\end{subfigure}
	\begin{subfigure}[t]{0.245\textwidth}
		\centering
		\includegraphics[width=1\textwidth]{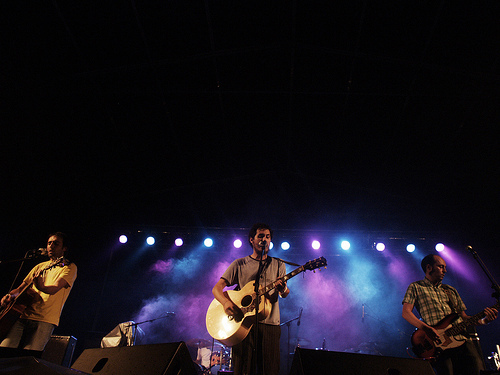}
		\caption{stage}
		\label{fig:anec_multi}
	\end{subfigure}
	\begin{subfigure}[t]{0.245\textwidth}
		\centering
		\includegraphics[width=1\textwidth]{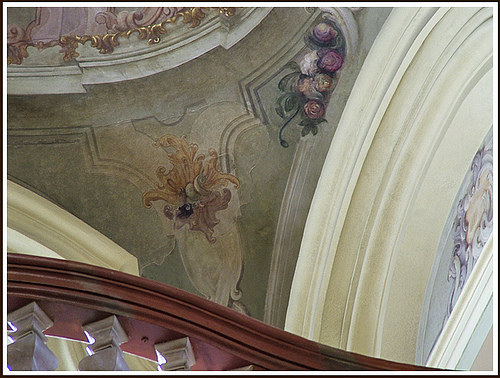}
		\caption{monastery}
		\label{fig:anec_context}
	\end{subfigure}
	\begin{subfigure}[t]{0.245\textwidth}
		\centering
		\includegraphics[width=1\textwidth]{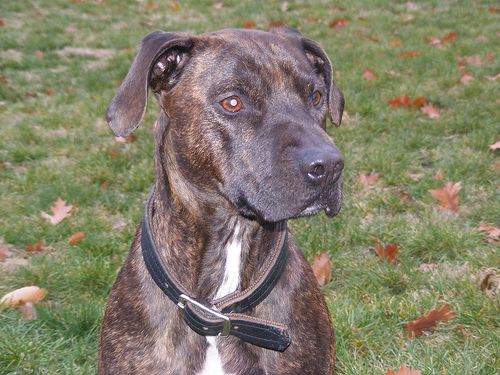}
		\caption{Staffordshire bull terrier}
	\end{subfigure}
	\caption{Judging the correctness of ImageNet labels may not be 
		straightforward. While the labels shown above appear valid for the corresponding
		images, \emph{none} of them actually match the ImageNet
		labels (which are ``projectile'', ``acoustic guitar'', ``church'',
        and ``American Staffordshire terrier'', respectively).}
	\label{fig:anec}
\end{figure}

Large-scale vision datasets and their associated
benchmarks~\citep{everingham2010pascal,deng2009imagenet,lin2014microsoft,
russakovsky2015imagenet} have been instrumental in guiding the development of
machine learning 
models~\citep{krizhevsky2012imagenet,szegedy2016rethinking,he2016deep}.
At the same time, while the progress made on these benchmarks is 
undeniable, they are only proxies for real-world tasks that we actually care about---e.g., 
object recognition or localization in the wild.
Thus, it is natural to wonder:
\begin{center}
	\emph{How aligned are existing benchmarks  
	with their motivating real-world tasks?}
\end{center}
On one hand, significant design effort goes towards ensuring that these benchmarks
accurately model real-world challenges~\citep{ponce2006dataset,torralba2011unbiased,everingham2010pascal,russakovsky2015imagenet}.
On the other hand, the sheer size of machine learning datasets makes
meticulous data curation virtually impossible. 
Dataset creators thus resort to scalable methods such as automated data retrieval and 
crowd-sourced 
annotation~\citep{everingham2010pascal,russakovsky2015imagenet,lin2014microsoft,
zhou2017places}, often at the cost of faithfulness to the task being
modeled.
As a result, the dataset and its corresponding annotations can sometimes be
ambiguous, incorrect, or otherwise misaligned with ground truth (cf. Figure~\ref{fig:anec}).
Still, despite our awareness of these
issues~\citep{russakovsky2015imagenet,recht2018imagenet,hooker2019selective,
northcutt2019confident}, we  lack a precise
characterization of their pervasiveness and impact, even for widely-used
datasets.  

\subsection*{Our contributions}
We develop a methodology for obtaining fine-grained data annotations via large-scale 
human studies.
These annotations allow us to precisely quantify ways in which 
typical \task{} benchmarks fall short of capturing the underlying ground truth. 
We then study how such \emph{benchmark-task misalignment} 
impacts state-of-the-art models---after all, 
models are often developed in a way that treats existing datasets as the ground truth.
We focus our exploration on 
the ImageNet dataset~\cite{deng2009imagenet} (specifically, the ILSVRC2012 \task{} 
task~\cite{russakovsky2015imagenet}), 
one of the most widely used benchmarks in computer vision.

\paragraph{Quantifying benchmark-task alignment.} We find that systematic annotation 
issues pervade ImageNet, and can often be
attributed to design choices in the dataset collection pipeline itself.
For example, during
the ImageNet labeling process, annotators were not asked to
classify images, but rather to validate a specific
automatically-obtained candidate label without  knowledge of other classes in the dataset. 
This leads to:
\begin{itemize}
    \item \emph{Multi-object images} (Section~\ref{sec:multi}): Each image in
      ImageNet is associated with a single label. Yet, we find that more than
      one fifth of ImageNet images contain objects from \emph{multiple}
      classes.  
        In fact, the
        ImageNet label often does not even correspond to {what humans deem} the
        ``main object'' in the image.
        Nevertheless, models
        still achieve significantly-better-than-chance prediction performance on these images, 
        indicating that they must exploit idiosyncrasies of the dataset that
        humans are oblivious to.
    \item \emph{Bias in label validation} (Section~\ref{sec:nonexp}):
        Even when there is only one object in an image, collectively, annotators often end up
        validating several mutually exclusive labels. 
        These correspond, for example, to images that are ambiguous or
        difficult to label, or instances where two or more classes have
        synonymous labels.
         The ImageNet label in these cases is determined not by 
        annotators 
        themselves, but rather by the fidelity of the automated image retrieval process.
        In general, given that annotators are ineffective at filtering out 
       errors under this setup, the automated component of the data pipeline
       may have a disproportionate impact on the quality of ImageNet
       annotations.
\end{itemize}

More broadly, these issues point an inherent tension between the goal of building 
large-scale 
benchmarks that are realistic and the scalable data collection pipelines needed to 
achieve this goal.
Hence, for benchmarks created using such pipelines, the current standard for model 
evaluation---accuracy 
with respect to a single, fixed dataset label---may be insufficient to correctly judge model 
performance.

\paragraph{Human-based performance evaluation.}  In light of this, we use our
annotation pipeline to more directly measure human-model alignment.
We find that more accurate ImageNet models  also
make predictions that annotators are more likely to agree with.
In fact, we find that models have reached a level where 
non-expert annotators are
largely unable to distinguish between predicted labels and 
ImageNet labels (i.e., even model predictions that don't match the dataset
labels are often judged valid as by annotators). 
While reassuring, this finding highlights a
different challenge: non-expert annotations may no longer suffice to tell apart further 
progress from overfitting to idiosyncrasies of the ImageNet distribution.

\section{A Closer Look at the ImageNet Dataset}
\label{sec:pipeline}
We start by briefly describing the original ImageNet data collection and 
annotation 
process. As it
turns out, several---seemingly innocuous---details of this process have a significant
impact on the resulting dataset.

\paragraph{The ImageNet creation pipeline.}
ImageNet is a prototypical example of a large-scale dataset (1000 classes and
more than million images) created through automated data collection and
crowd-sourced filtering.  At a high level, this creation process comprised
two stages~\citep{deng2009imagenet}:

\begin{enumerate}
	\item \emph{Image and label collection:} The ImageNet creators first selected a set of 
	classes using the WordNet hierarchy~\citep{miller1995wordnet}. 
	Then, for each class, they sourced images by querying several 
	search engines with the WordNet synonyms of the class, augmented 
	with synonyms of its WordNet parent node(s). For example, the 
	query for class ``whippet'' (parent: ``dog'') would also include 
	``whippet dog''~\citep{russakovsky2015imagenet}.
	To further expand the image pool, the dataset creators also performed queries in
	several  languages. Note that for each of the retrieved images, the 
	label---which we will 
	refer to as the {\em proposed label}---that will be assigned to this image
	should it be included in the dataset, is already determined. That is, the 
	label 
	is simply given by the WordNet node that was used for the corresponding 
	search query.
	
	\item \emph{Image validation via the} \contains{} \emph{task.}
	To validate  the image-label pairs retrieved in the previous stage, the ImageNet 
	creators
	employed annotators via the Mechanical Turk (MTurk) crowd-sourcing
	platform.
	Specifically, for every class, annotators were presented with its description 
	(along with 
	 links to the relevant Wikipedia pages) and a 
	 grid of candidate images.
	Their task was then to select all images in that grid that contained an
	object of that class (with {\em explicit} instructions to ignore clutter and 
	occlusions).
	The grids were shown to multiple annotators and only images that 
	received a
	 ``convincing majority'' of votes (based on per-class thresholds 
	 estimated using a small pool of annotators)
	 were included in ImageNet.
     In what follows, we will refer to this filtering procedure as the \contains{} task.
\end{enumerate}

\subsection*{Revisiting the ImageNet labels}
\label{sec:pipeline_issues}
The automated process described above is a natural method for creating
a large-scale dataset, especially if it involves a wide range of classes (as is the case for 
ImageNet).
However, even putting aside occasional annotator errors, the resulting 
dataset might not accurately capture the ground truth (see Figure 
\ref{fig:anec}).
Indeed, as we discuss below, this pipeline design itself can lead to certain
\emph{systematic} errors in the dataset. 
The root cause for many of these errors is that the image 
validation stage (i.e., the \contains{} task)
asks annotators only to verify if a \emph{specific} proposed label (i.e., WordNet node 
for which the image was retrieved), shown in isolation, is 
valid for a given image.
Crucially, annotators are never asked to choose among different possible 
labels for
the image and, in fact, have no knowledge of what the other classes even are.
This can introduce discrepancies in the dataset in two ways:

\paragraph{Images with multiple objects.} 
Annotators are instructed to \emph{ignore} the presence of other objects when
validating a particular ImageNet label for an image.
However, these objects could themselves correspond to other ImageNet classes.
This can lead to the selection of images with multiple valid labels or even to	
images where the dataset label does not correspond to the most prominent 
object in the image.

\paragraph{Biases in image filtering.} 
Since annotators have no knowledge of what the other classes are, they do not have a 
sense of the
granularity of image features they should pay attention to (e.g., the labels in 
Figure~\ref{fig:anec} appear reasonable until one becomes aware of the other 
possible classes in ImageNet).
Moreover, the task itself does not necessary account for their expertise (or the
lack of thereof). Indeed, one cannot reasonably expect non-experts to
distinguish, e.g., between all the 24 terrier breeds that are present in
ImageNet. 
As a result, if annotators are shown images containing objects of a different,
yet similar class, they are likely to select them as valid.
This implies that potential errors in the collection process (e.g., automated 
search retrieving images that do not match the query label) are unlikely to be 
corrected during validation (via the \contains{} task) and thus
can propagate to the final dataset. \\
 
In the light of the above, it is clear that eliciting ground truth information 
from annotators	using the ImageNet creation pipeline may not be 
straightforward.	
In the following sections, we present a framework for improving this elicitation (by bootstrapping from and refining the existing labels) and then use that framework to investigate the
discrepancies highlighted above (Section~\ref{sec:method}) and their impact on 
ImageNet-trained
models (Section~\ref{sec:issues}).  	

\section{From Label Validation to Image Classification}
\label{sec:method}

\begin{figure*}[!t]
	\centering
	\includegraphics[width=1\textwidth]{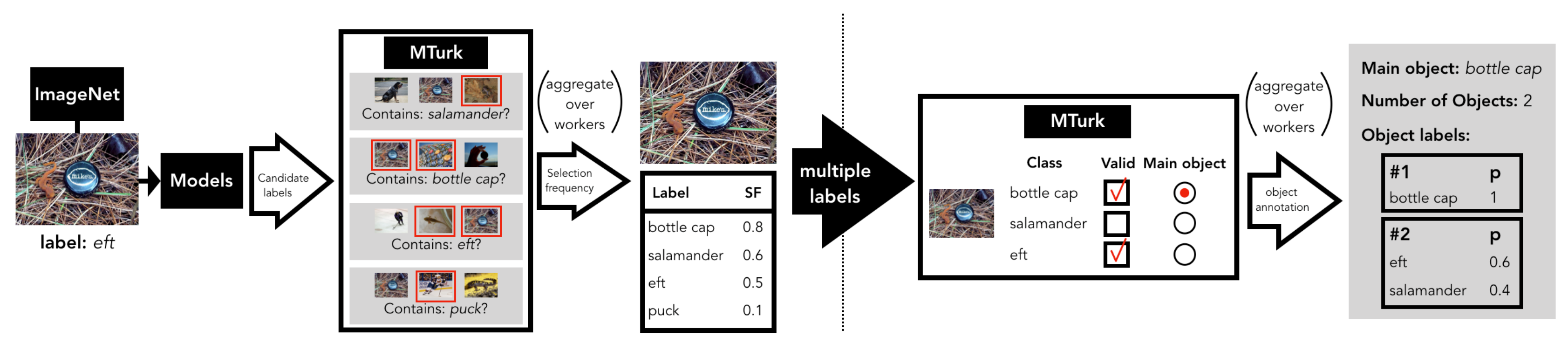}
    \caption{Overview of our data annotation pipeline.
        First, we collect a pool of potential labels for each image using the
        top-5 predictions of multiple models (Section~\ref{sec:contains}).
        Then, we ask annotators to gauge the validity of each label (in
        isolation) using the \contains{} task (described in
        Section~\ref{sec:pipeline}).
        Next, we present all highly selected labels for each image to a
        new set of annotators and ask them to select \emph{one} label for every
        distinct object in the image, as well as a label for the main object
        according to their judgement, i.e., the \classify{} task 
        (Section~\ref{sec:followup}).
        Finally, we aggregate their responses to obtain fine-grained image
        annotations (Section~\ref{sec:post_proc}).  }
    \label{fig:method}
\end{figure*}

We begin our study by obtaining a better understanding of the ground truth for
ImageNet data.
To achieve this, rather than asking annotators to \emph{validate} a single proposed label
for an image (as in the original pipeline), we would like them to
\emph{classify} the image, selecting {\em all} the relevant labels for it.
However, asking (untrained) annotators to choose from among all 1,000 ImageNet classes
is infeasible.

To circumvent this difficulty, our pipeline consists of two phases, illustrated in
Figure~\ref{fig:method}.
First, we obtain a small set of (potentially) relevant \emph{candidate labels} for each image 
(Section~\ref{sec:contains}).
Then, we present these labels to annotators and ask them to select one of them
for {\em each} distinct object using what we call the \classify{} task 
(Section~\ref{sec:followup}).
These phases are described in more detail below, with additional information in 
Appendix~\ref{app:methodology}. 

For our analysis, we use 10,000 images from the ImageNet validation 
set---i.e., 10 randomly selected images per class.
Note that since both ImageNet training and ImageNet validation sets were created using the
same procedure, analyzing the latter is sufficient to understand systematic
issues in that dataset.

\subsection{Obtaining candidate labels}
\label{sec:contains}
To ensure that the per-image annotation task is manageable, we narrow
down the candidate labels to a small set.
To this end, we first obtain \emph{potential labels} for each image by simply
combining the top-5 predictions of 10 models from different parts of the accuracy
spectrum with the existing ImageNet label (yields approximately 14 labels per 
image; cf. Appendix
Figure~\ref{fig:cand_label_dist}).
Then, to prune this set further, we reuse the ImageNet \contains{}
task---asking annotators whether an image contains a particular
class (Section~\ref{sec:pipeline})---but for \emph{all} potential labels.
The outcome of this experiment is a \emph{selection frequency}  for each
image-label pair, i.e., the fraction of annotators that selected the image
as containing the corresponding label.\footnote{Note that this
notion of selection frequency (introduced by~\citet{recht2018imagenet}) essentially 
mimics the majority voting process used to create  
ImageNet (cf. Section~\ref{sec:pipeline}), except that it uses 
a fixed number of 
annotators per grid instead of the original adaptive process.}
We find that, although images were often selected as valid for
many labels, relatively few of these labels had high selection frequency
(typically less than five per image).
Thus, restricting potential labels to this smaller set of \emph{candidate
labels} allows us to hone in on the most likely ones, while ensuring that the
resulting annotation task is still cognitively tractable.

\subsection{Image classification via the \classify{} task}
\label{sec:followup}
\label{sec:followup_specs}
\label{sec:post_proc}
Once we have identified a small set of candidate labels for each image, we 
present them to annotators to obtain fine-grained label information.
Specifically, we ask annotators to identify: (a) \emph{all} labels that correspond
to objects in the image, and (b) the label for the \emph{main} object (according
to their judgment).
Crucially, we explicitly instruct annotators to select only \emph{one label per distinct 
object}---i.e., in case they are confused about the correct label for a specific object, to pick 
the one 
they consider most likely.
Moreover, since ImageNet contains classes that could describe parts or attributes
of a single physical entity (e.g., "car" and "car wheel"), we ask annotators to treat these 
as 
distinct objects, since they are not
mutually exclusive.
We present each image to multiple annotators and then aggregate their responses 
(per-image) as described below.
In the rest of the paper, we refer to this annotation setup as the \classify{}
task.

\paragraph{Identifying the main label and number of objects.}
From each annotator's response, we learn what they consider to be the label of the main
object, as well as how many objects they think are present in the image.
By aggregating these two quantities based on a majority vote over annotators,
we can get an estimate of the number of objects in the image, as
well as of the main label for that image.
    
\paragraph{Partitioning labels into objects.}
Different annotators may choose different labels for the same object and thus we need to
map their selections to a single  set of distinct objects.
To illustrate this, consider an image of a soccer ball and a terrier,
where one annotator has selected ``Scotch terrier'' and ``soccer ball'' and
another ``Toy terrier'' and ``soccer  ball''.
We would like to partition these selections into objects as [``soccer
ball''] and [``Scotch terrier'' or ``Toy terrier''] since both responses indicate
that there are two objects in the image and that the soccer ball and the terrier
are distinct objects.
More generally, we would like to partition selections in a way that avoids grouping labels 
together if annotators identified them as distinct objects.
To this end, we employ exhaustive search to find a partition that optimizes for
this criterion (since there exist only a few possible partitions to begin with).
Finally, we label each distinct object with its most frequently selected label. \\

The resulting annotations characterize the content of an image in a more fine-grained 
manner compared to the original ImageNet labels.
Note that these annotations may still not perfectly match the ground truth.
After all, we also employ untrained, non-expert annotators, that make occasional
errors, and moreover, some images are inherently 
ambiguous without further context (e.g., Figure~\ref{fig:anec_context}).  Nevertheless, as
we will see, these annotations are already sufficient for our examination of ImageNet. 

\section{Quantifying the Benchmark-Task Alignment of ImageNet}
\label{sec:issues}

Our goal in this section is two-fold. First, we would like to examine potential 
sources of discrepancy between ImageNet and the 
motivating \task{} task, using our refined 
image annotations.
\footnote{Our primary focus is on studying systematic issues in ImageNet. 
Thus we defer discussing clear mislabelings to Appendix~\ref{app:mislabeled}.}
Next, we want to assess the 
impact these deviations have on on models developed using this benchmark. 
To this end, we will measure how the accuracy of a diverse set of models
(see Appendix~\ref{app:models} for a list), is affected when they are evaluated
on different sets of images.

\subsection{Multi-object images}
\label{sec:multi}

\begin{figure}[!h]
	\begin{center}
		\begin{subfigure}{0.26\textwidth}
			\centering
			\includegraphics[width=1\textwidth]{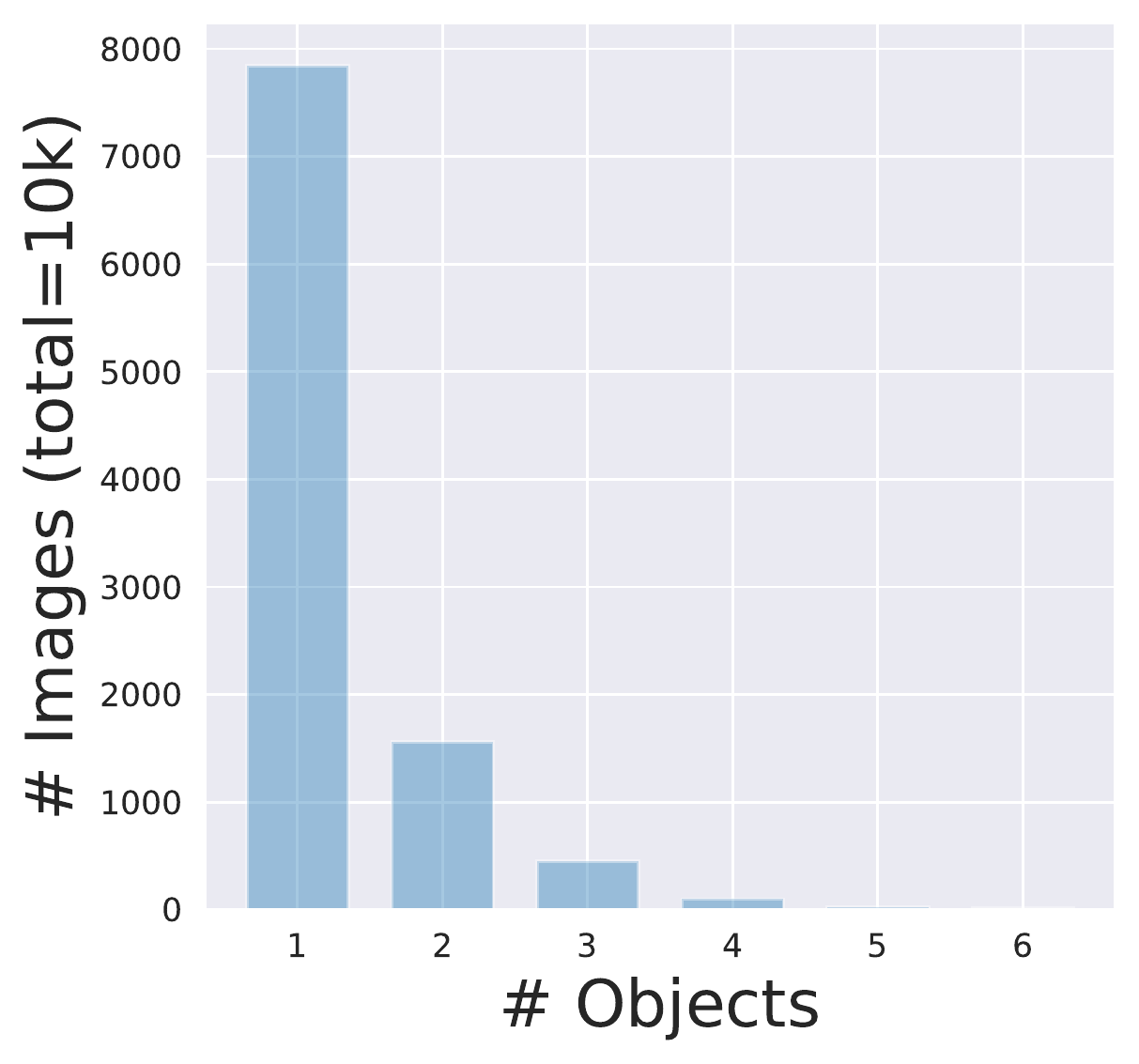}
			\caption{}
			\label{fig:object_hist}
		\end{subfigure}
		\begin{subfigure}{0.7\textwidth}
			\centering
			\includegraphics[width=1\textwidth]{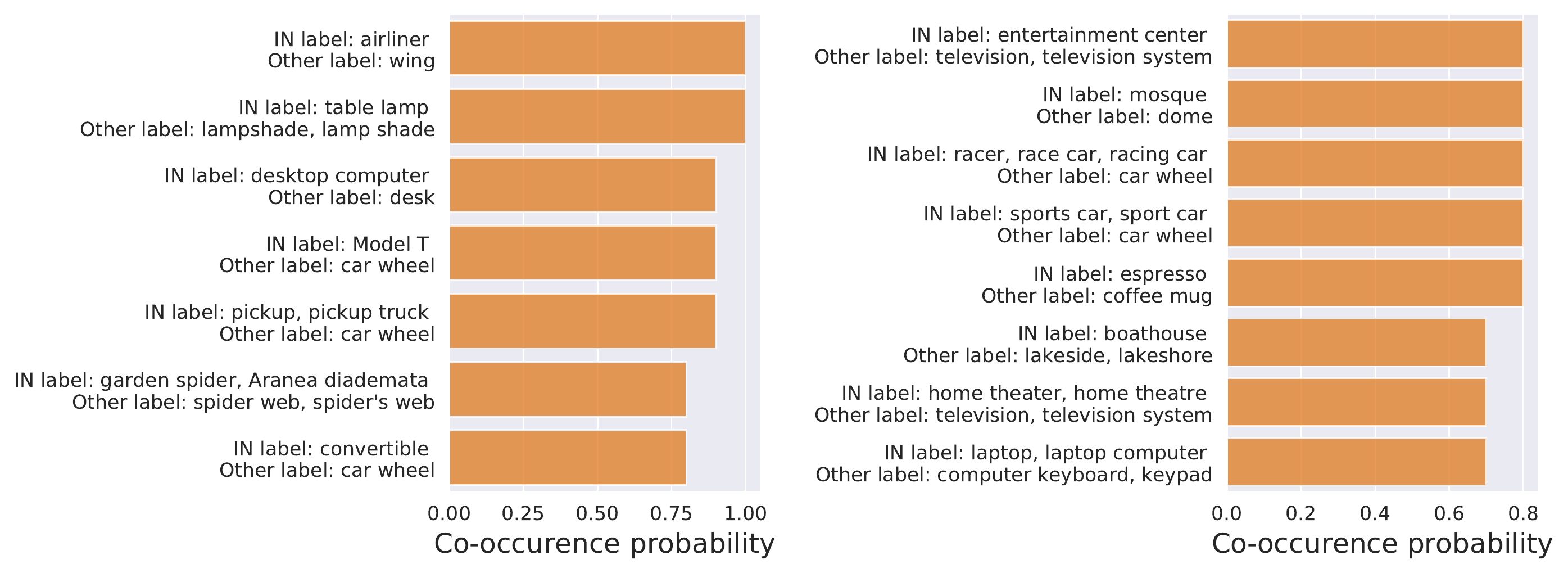}
			\caption{}
			\label{fig:class_coocc}
		\end{subfigure} 
		\hspace{.5em}
		\begin{subfigure}{1\textwidth}
			\centering
			\includegraphics[width=1\textwidth]{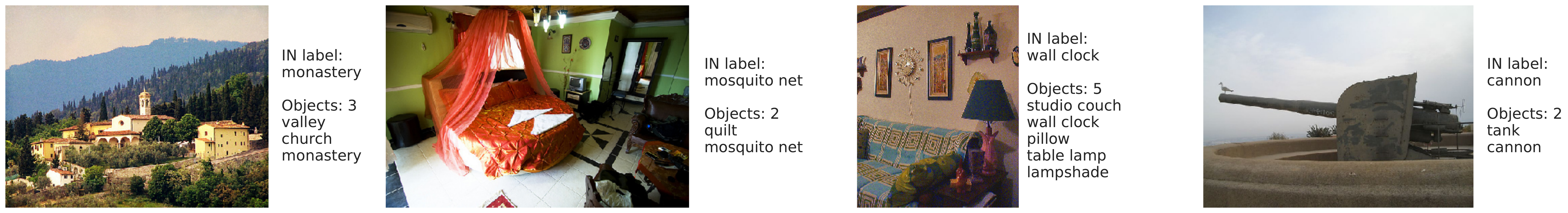}
			\caption{}
			\label{fig:multi_examples_pap}
		\end{subfigure}
	\end{center}
	\caption{(a) Number of objects per image---more than a
		fifth of the images contains two or more objects from ImageNet classes.
		(b) Pairs of classes which consistently co-occur as distinct
		objects. Here, we visualize the top 15 ImageNet classes based on how
		often their images contain another \emph{fixed} object (``Other
		label'').
		(c) Random examples of multi-label ImageNet images (cf. Appendix 
		Figure~\ref{fig:multi_examples} for additional samples).
	}
	\label{fig:multi}
\end{figure}

We start by taking a closer look at images which contain objects from more than
one ImageNet class---how often these additional objects appear and how salient
they are.
(Recall that if two labels are both simultaneously valid for an image---i.e., they
are not mutually exclusive---we refer to them as different objects (e.g.,
``{car}'' and ``{car wheel}'').)
In Figure~\ref{fig:object_hist}, we observe that a significant fraction of
images---more than a fifth---contains \emph{at least two} objects.
In fact, there are pairs of classes which \emph{consistently co-occur} (see 
Figure~\ref{fig:class_coocc}). This indicates that multi-object images in ImageNet are not 
caused  solely by irrelevant clutter, but also due to systemic issues with the 
class selection process. Even though ImageNet classes, in principle, 
correspond to distinct objects, these objects overlap greatly in terms of where they occur 
in the real-world.

\paragraph{Model accuracy on multi-object images.}
Model performance is typically measured using (top-1 or top-5) accuracy with
respect to a \emph{single} ImageNet label, treating it as the ground truth.
However, it is not clear what the right notion of ground truth annotation even
is when classifying multi-object images. 
Indeed, we find that models perform significantly worse on multi-label images
based on top-1 accuracy (measured w.r.t. ImageNet labels): accuracy drops by
more than 10\% across all models---see Figure~\ref{fig:kobjects}.
In fact, we observe in Appendix Figure~\ref{fig:cooccurrence_bars} that model
accuracy is especially low on certain classes that systematically co-occur.
\begin{figure}[!h]
	\centering
	\begin{subfigure}{0.48\textwidth}
		\includegraphics[width=1\textwidth]{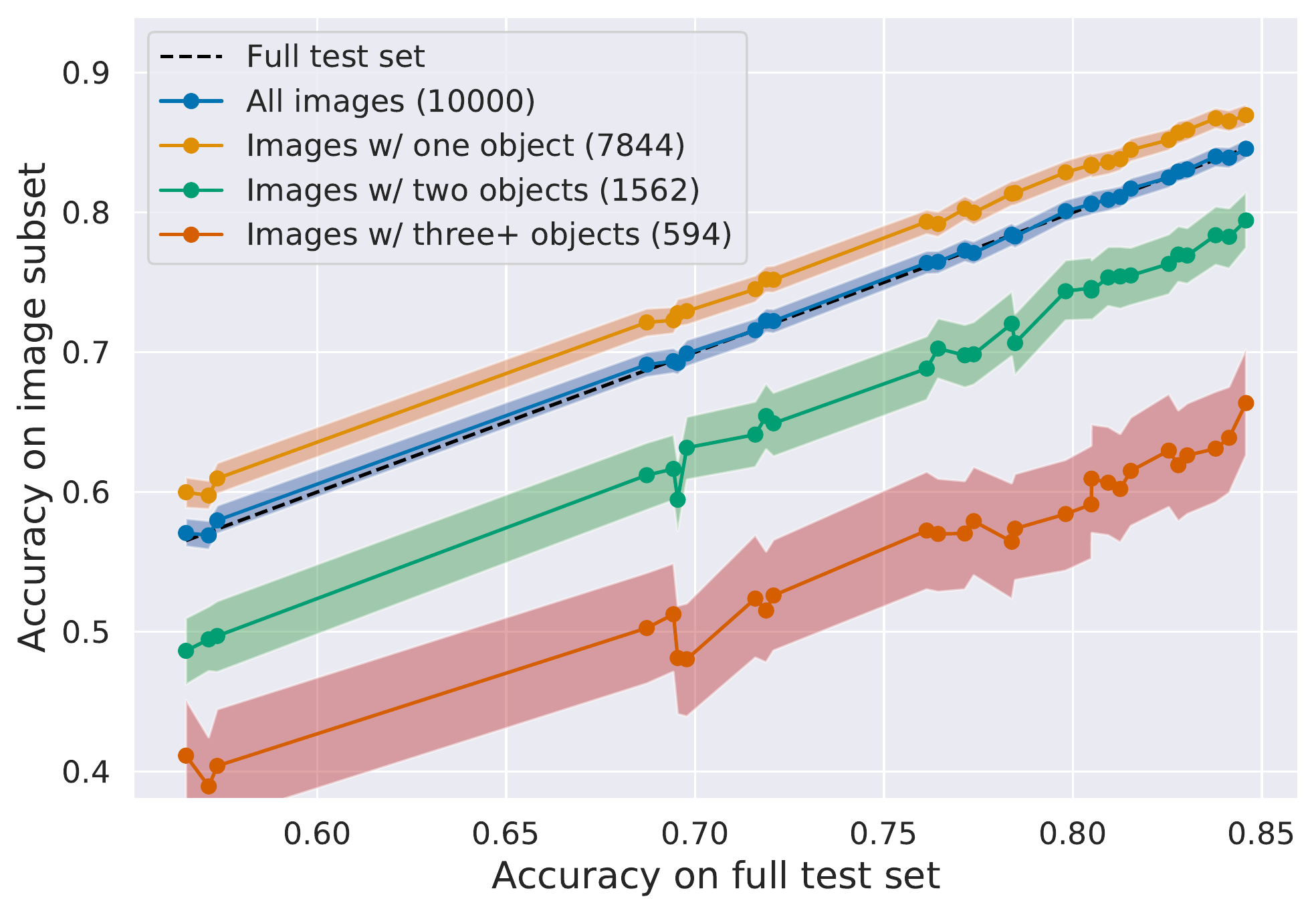}
		\caption{}
		\label{fig:kobjects}
	\end{subfigure}
	\hfil
	\begin{subfigure}{0.48\textwidth}
		\centering
		\includegraphics[width=1\textwidth]{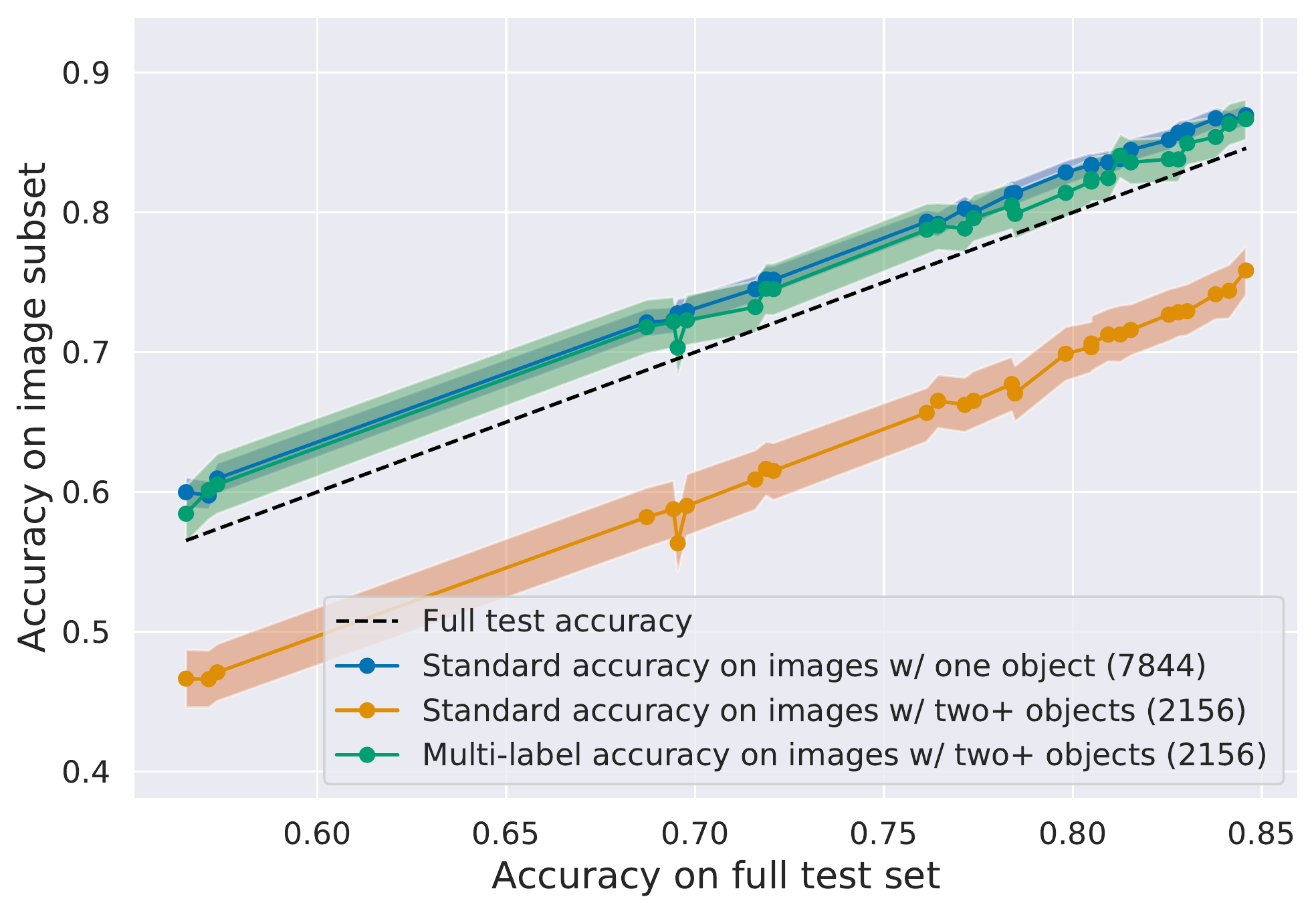}
		\caption{}
		\label{fig:multi_accuracy}
	\end{subfigure}
	\caption{(a) Top-1 model accuracy on multi-object images (as a function
		of overall test accuracy).  Accuracy drops by roughly 10\% across all
		models.
		(b) Evaluating multi-label accuracy on ImageNet: the fraction of images
		where the model predicts the label of \emph{any} object in the image.
		Based on this metric, the performance gap between single- and
		multi-object images virtually vanishes. Confidence intervals: 95\% via
        bootstrap.}
\end{figure}
 
In light of this, a more natural notion of accuracy for multi-object images would be to 
consider a model prediction to be correct if it matches the label of \emph{any} object in 
the image. 
On this metric, we find that the aforementioned performance
drop essentially disappears---models perform similarly on single- and
multi-object images (see Figure~\ref{fig:multi_accuracy}).
This indicates that the way we typically measure accuracy, i.e., with respect to a
single label, can be overly pessimistic, as it penalizes the model for
predicting the label of another valid object in the image.
Note that while evaluating top-5 accuracy also accounts for most of these
multi-object confusions (which was after all its original 
motivation~\citep{russakovsky2015imagenet}), it tends to inflate accuracy on 
\emph{single object} images, by treating several erroneous
predictions as valid (cf.\ Appendix~\ref{app:clutter}).

\begin{figure}[!h]
	\begin{center}
\begin{subfigure}{0.35\textwidth}
	\centering
	\includegraphics[width=0.9\textwidth]{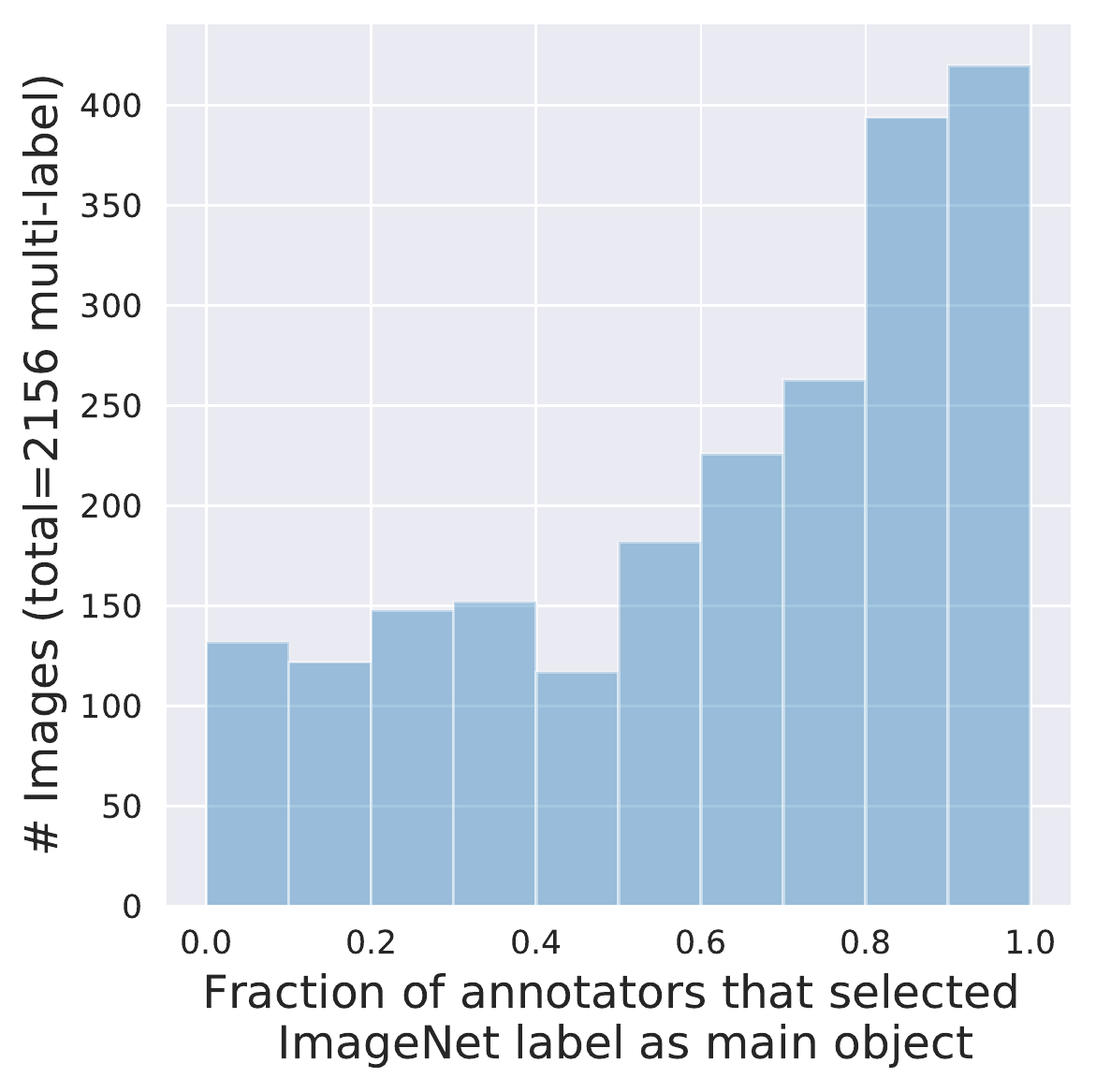}
	\caption{}
	\label{fig:main_gt_dist}
\end{subfigure}
\hfil
\begin{subfigure}{0.62\textwidth}
	\centering
	\includegraphics[width=0.9\textwidth]{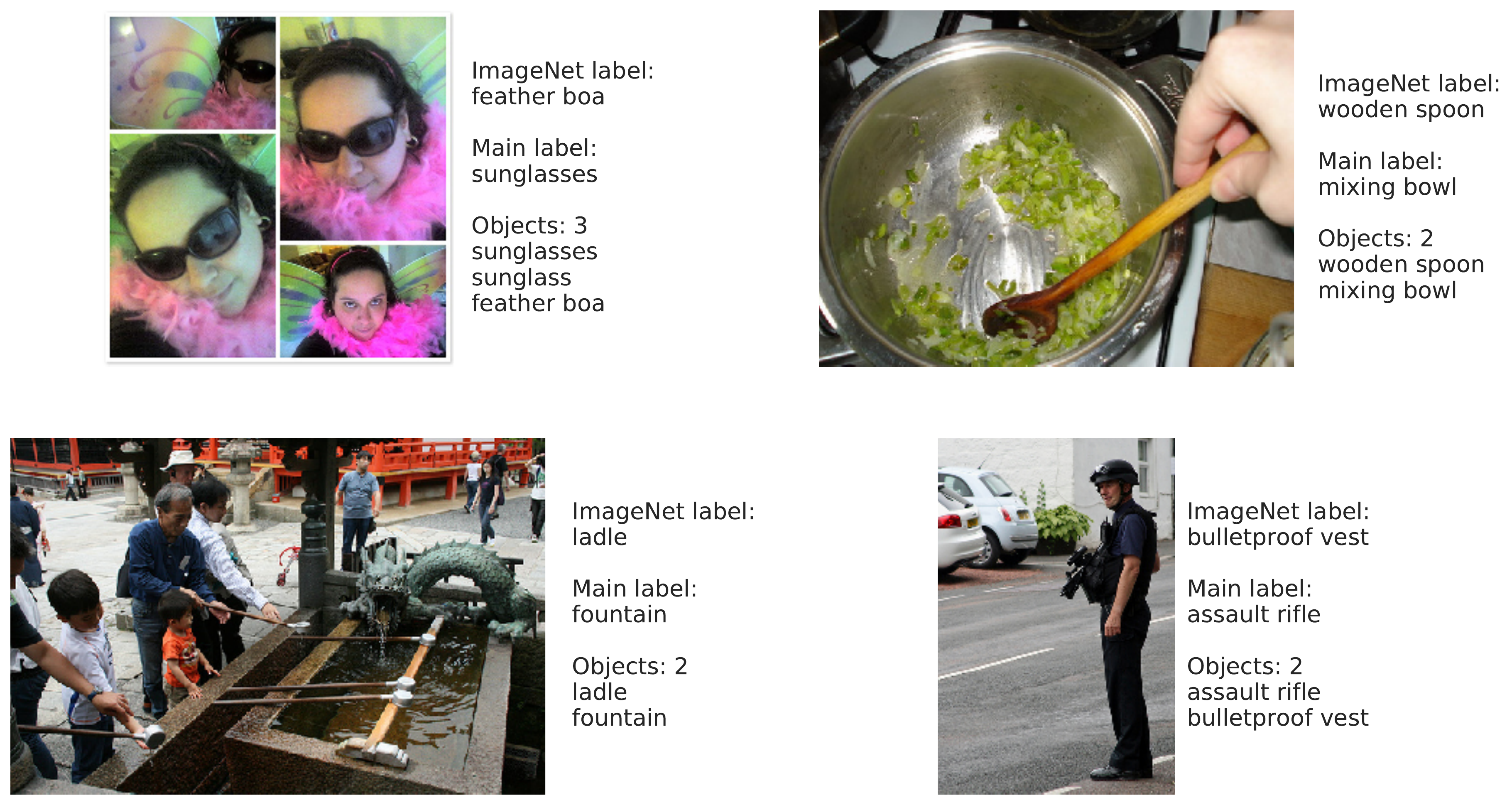}
	\caption{}
	\label{fig:disagreement_exs_pap}
\end{subfigure}
	\end{center}
    \caption{(a) Fraction of annotators that selected the ImageNet label
        as denoting the main image object. For 650 images (out of 2156
        multi-object images), the majority of annotators select a
        label \emph{other} than the ImageNet one as the main object.
        (b) Random examples of images where the main label
        as per annotators contradicts the ImageNet label (cf. Appendix
        Figure~\ref{fig:disagreement} for additional samples).
    }
	\label{fig:disagreement_pap}
\end{figure}

\begin{figure}[!h]
	\begin{subfigure}{0.39\textwidth}
		\centering
		\includegraphics[width=1\textwidth]{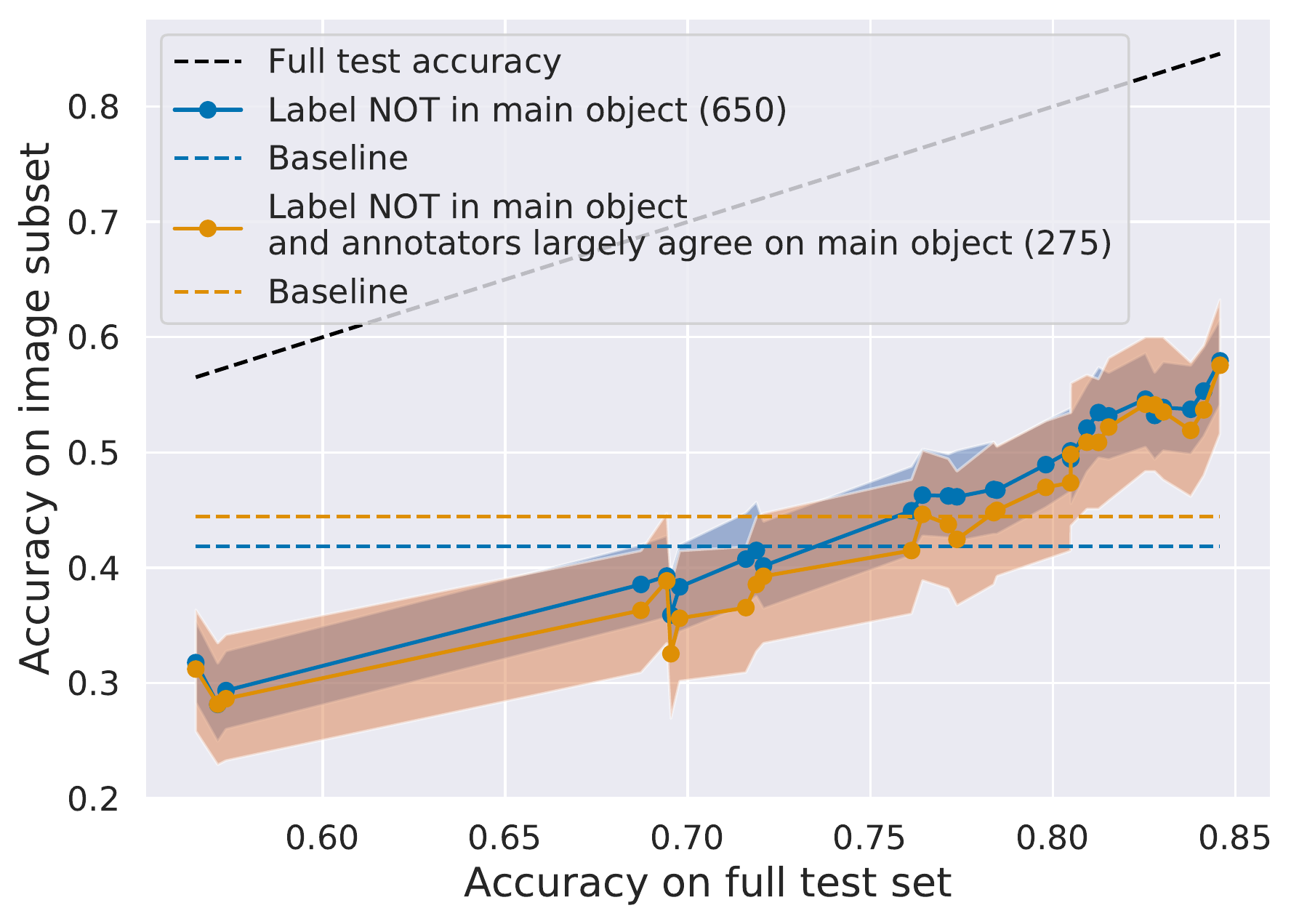}
		\caption{}
		\label{fig:overfitting}
	\end{subfigure}
	\hfil
	\begin{subfigure}{0.57\textwidth}
		\centering
		\includegraphics[width=1\textwidth]{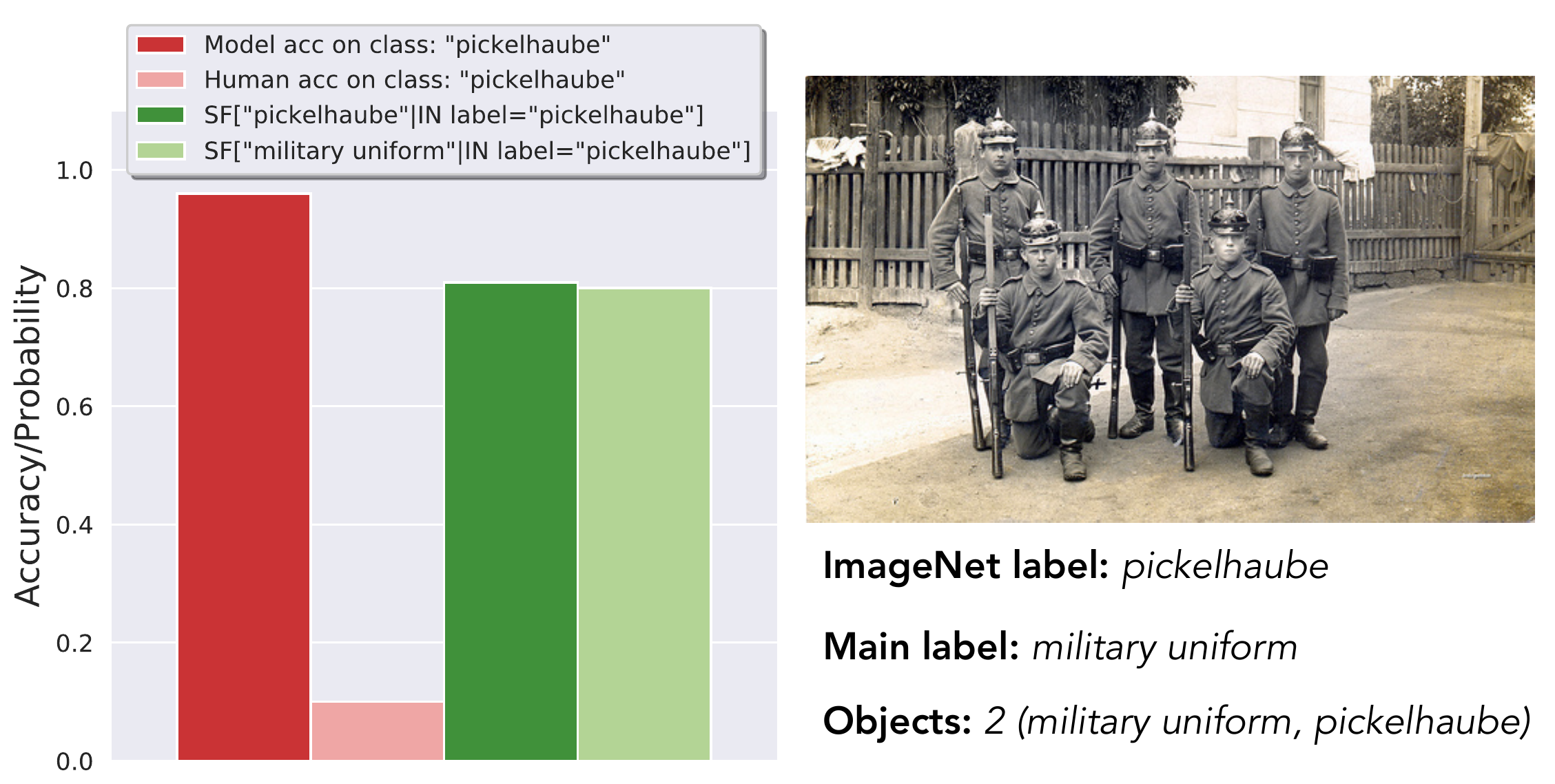}
		\caption{}
		\label{fig:bearskin}
	\end{subfigure}
    \caption{(a) Model accuracy on images where the annotator-selected main
        object does not match the ImageNet label.
        Models perform much better than the baseline of randomly choosing
        one of the objects in the image (dashed line)---potentially by
        exploiting dataset biases.
		(b) Example of a class where humans disagree with the label as 
        to the main object, yet models still predict the ImageNet
        label. Here, for images of that class, we plot both model and annotator
        accuracy as well as the selection frequency (SF) of both labels.
	}
\end{figure}

\paragraph{Human-label disagreement.}
Although models suffer a sizeable accuracy drop on multi-object images, they
are still relatively good at predicting the ImageNet label---much
better than the baseline of choosing the label of one object at random.
This bias could be justified whenever there is a distinct main 
object in the image and it corresponds to the ImageNet label.
However, we find that for nearly a third of the multi-object images, the
ImageNet label \emph{does not} denote the most likely main object as judged by
human annotators---see Figure~\ref{fig:disagreement_pap}.
Nevertheless, model accuracy (w.r.t. the ImageNet label)
on these images is still high---see Figure~\ref{fig:overfitting}.


On these samples, models must 
base their predictions on some features that humans do not 
consider salient.
For instance, we find that these disagreements often arise when the 
ImageNet label corresponds to a very distinctive object 
(e.g., ``pickelhaube''), but the image also contain another more 
prominent object (e.g., ``military uniform'')---see Figure~\ref{fig:bearskin}.
To do well on these classes, the model likely picks up on 
ImageNet-specific biases, e.g., detecting that military uniforms often 
occur in images of class ``pickelhaube'' but not vice versa. 
While this might be a valid strategy for improving ImageNet accuracy (cf.
Appendix Figure~\ref{fig:bearskin_bars}), it causes models to rely on features 
that may not generalize to the real-world.


\subsection{Bias in label validation}
\label{sec:nonexp}
We now turn our attention to assessing the quality of the ImageNet filtering
process discussed in Section~\ref{sec:pipeline}.
Our goal is to understand how likely annotators are to
detect incorrect samples during the image validation process.
Recall that annotators are asked a somewhat leading question, of whether
a specific label is present in the image, making them prone to
answering positively even for images of a different, yet similar, class.

\begin{figure}[!h]
	\centering
	\includegraphics[width=0.85\textwidth]{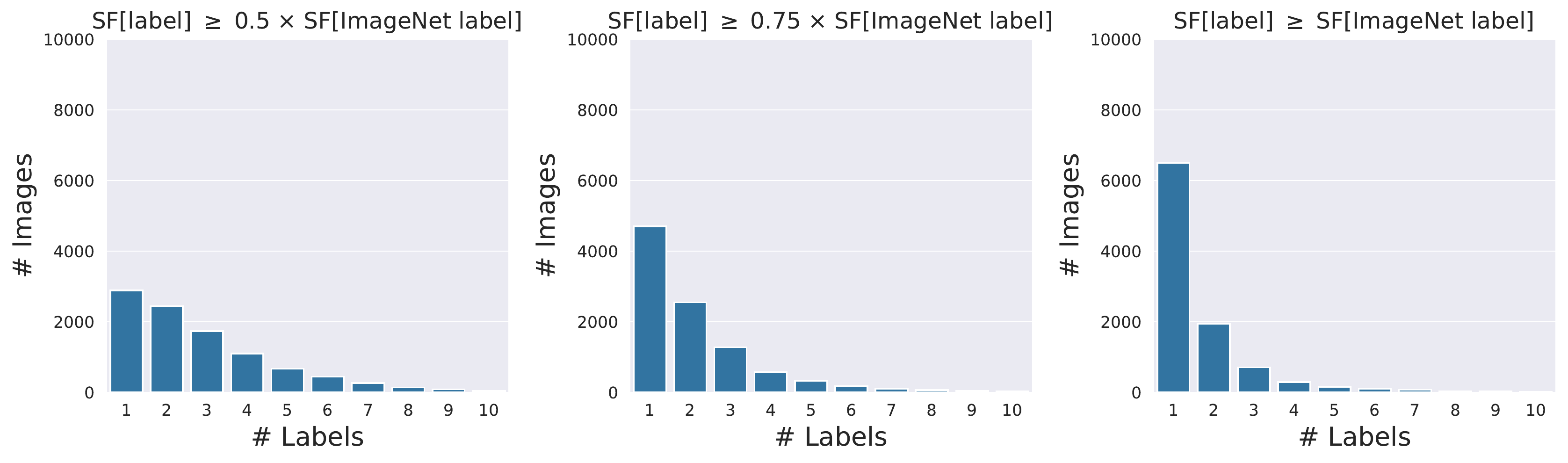}
    \caption{Number of labels per image that annotators selected as valid in
        isolation (determined by the selection frequency of the label relative
        to that of the ImageNet label).
        For more than 70\% of images, annotators select another label
        at least half as often as they select the ImageNet label
        (\emph{leftmost}).
	}
	\label{fig:confident}
\end{figure}

\begin{figure}[!h]
	\begin{center}
	\begin{subfigure}{0.38\textwidth}
	\centering
	\includegraphics[width=0.9\textwidth]{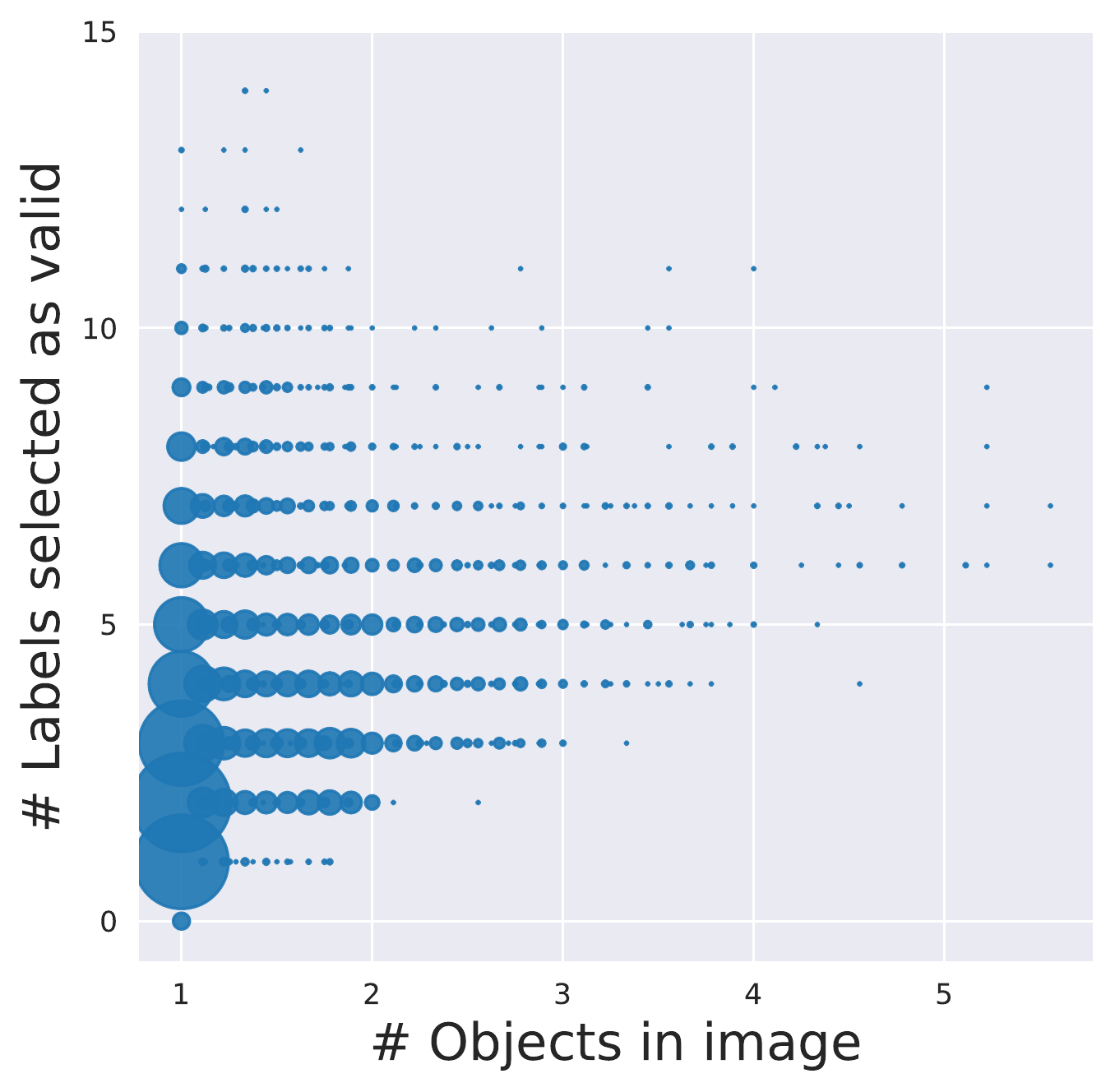}
	\caption{
	}
	\label{fig:obj_contains_scatter}
\end{subfigure}
\hfil
\begin{subfigure}{0.6\textwidth}
	\centering
	\includegraphics[width=0.9\textwidth]{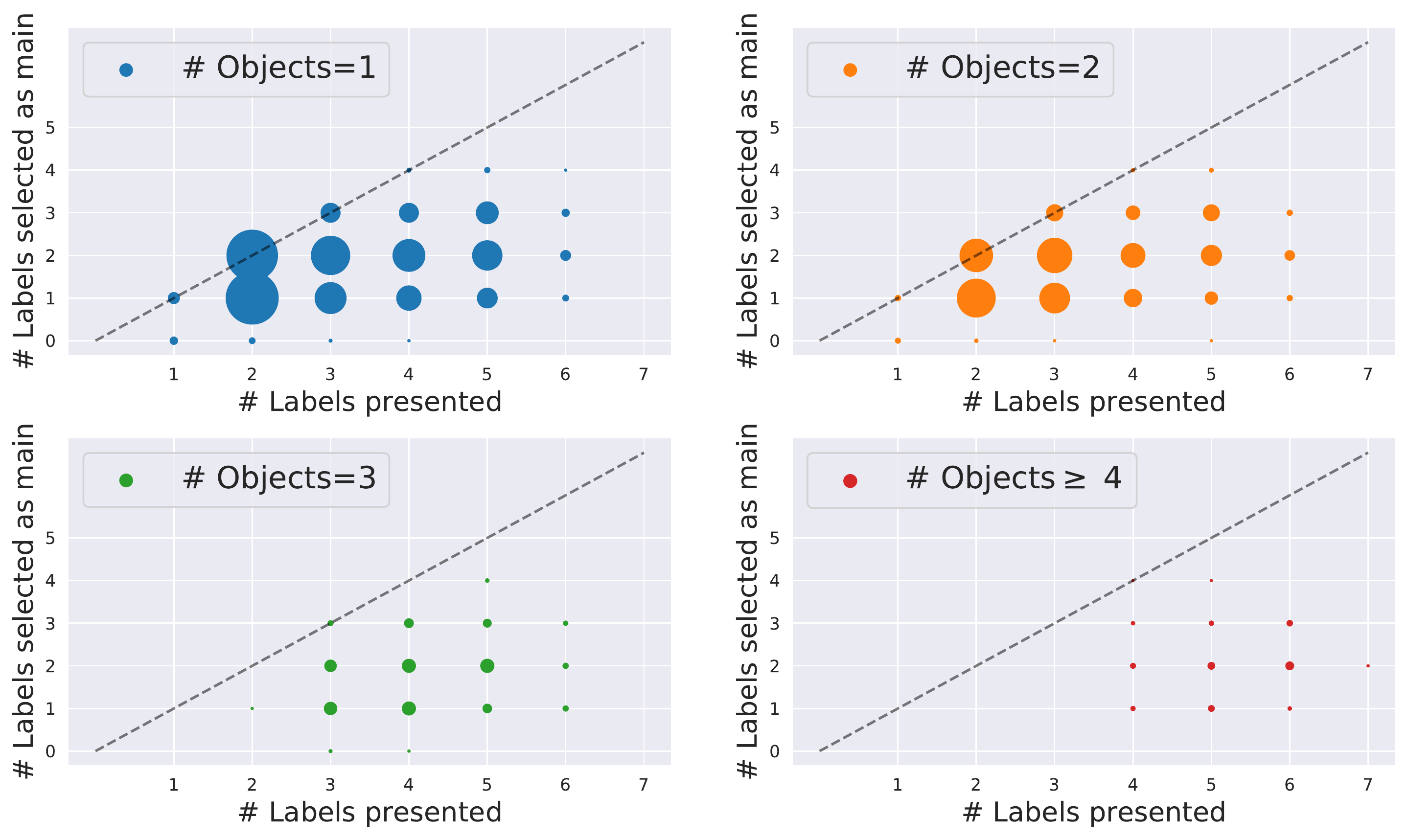}
	\caption{}
	\label{fig:obj_main_scatter}
\end{subfigure}
	\end{center}
    \caption{(a) Number of labels selected in the \contains{} task: 
        the y-axis measures the number of labels that were selected by at least
        two annotators;
        the x-axis measures the average number of objects indicated to be
        present in the image during the \classify{} task;
        the dot size represents the number of images in each 2D bin.
    	Even when annotators perceive an images as depicting a single object,
    	they often select multiple labels as valid. 
        (b) Number of labels that at least two annotators selected for
        the main image object (in the \classify{} task; cf.\ 
        Section~\ref{sec:followup}) as a function of the number of labels
        presented to them.
        Annotator confusion decreases significantly when the task
        setup explicitly involves choosing between multiple labels
         \emph{simultaneously}.}
	\label{fig:obj_scatter}
\end{figure}

Indeed, we find that, under the original task setup
(i.e., the \contains{} task), annotators consistently select multiple labels,
\emph{in addition to} the ImageNet label, as being valid for an image.
In fact, for nearly 40\% of the images, another label is selected \emph{at least
as often} as the ImageNet label (cf.  Figure~\ref{fig:confident}). 
Moreover, this phenomenon does not occur only when multiple 
objects are present in the image---even when annotators perceive a single object, they 
often select as many as 10 classes
(cf.~Figure~\ref{fig:obj_contains_scatter}).
Thus, even for images where a single ground  truth label exists, the ImageNet validation 
process may fail to elicit this label from annotators.\footnote{Note that our estimates for 
the selection frequency of image-ImageNet
    label pairs may be biased (underestimates)~\cite{engstrom2020identifying} as
    these specific pairs have already been filtered during dataset creation
    based on their selection frequency. However, we can effectively ignore this
    bias since: a) our results seem to be robust to varying the number of
    annotators (Appendix~\ref{app:selection_frequency}), b) most of our results
    are based on the follow-up \classify{} task for which this bias
    does not apply.}

In fact, we find that this confusion is not just a consequence of using
non-expert annotators, but also of the \contains{} task setup itself.
If instead of asking annotators to judge the validity of a specific label(s) in isolation, 
we ask them to choose the main object among several possible labels simultaneously (i.e., 
via the \classify{} task), 
they select substantially fewer labels---see Figure~\ref{fig:obj_main_scatter}.

These findings highlight how sensitive annotators are to the data collection and 
validation pipeline setup, even to aspects of it that may not seem significant at first glance.
It also indicates that the existing ImageNet annotations may not have be 
vetted as carefully as one might expect.
ImageNet annotators might have been unable to correct certain errors, and consequently, 
the resulting class distributions may be determined, to a large extent, by the fidelity (and 
biases) of the automated image retrieval process.

\begin{figure}[!h]
	\begin{subfigure}{0.68\textwidth}
		\includegraphics[height=0.6\textwidth]{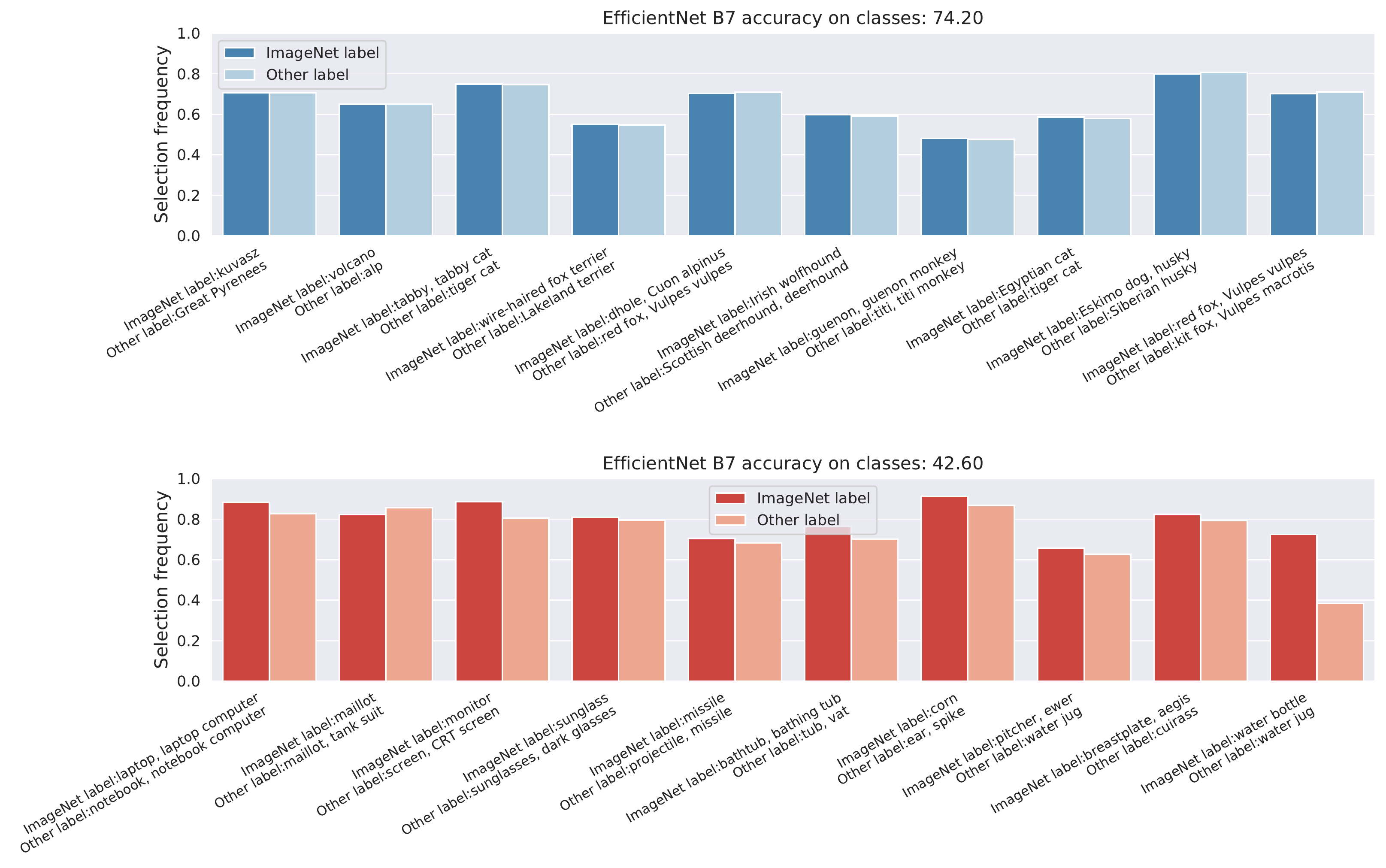}
		\caption{}
		\label{fig:amb_examples_pap}
	\end{subfigure}
	\hfil
	\begin{subfigure}{0.30\textwidth}
		\centering
		\includegraphics[height=0.95\textwidth]{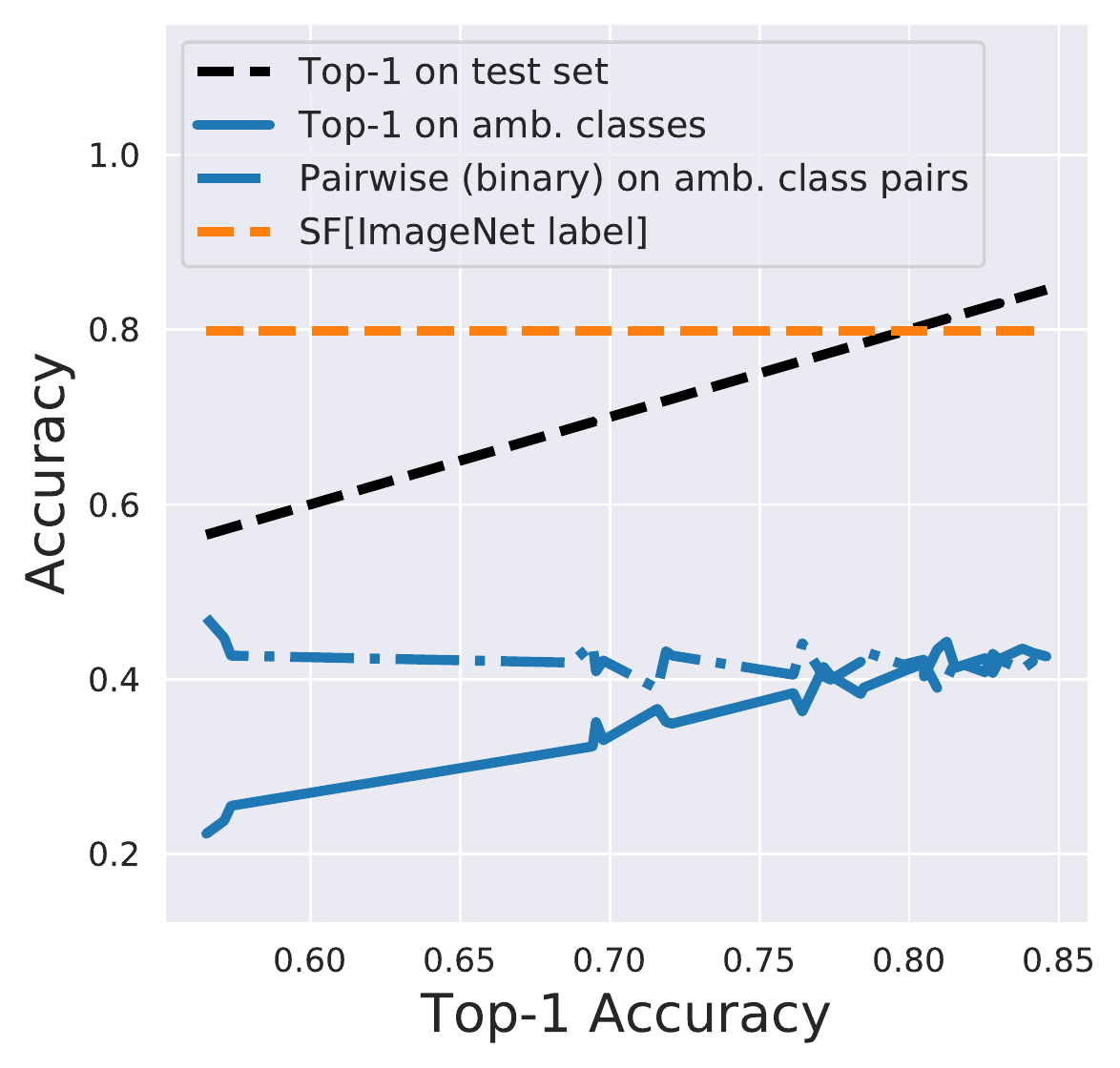}
		\caption{}
		\label{fig:amb}
	\end{subfigure}
	\caption{(a) ImageNet class pairs for which annotators often 
		deem both classes as valid.
		We visualize the top 10 pairs split based on the accuracy of
		EfficientNet B7 on these pairs being high
		(\emph{top}) or low (\emph{bottom}). 
		(b) Model progress on ambiguous class pairs (from (a)
        \emph{bottom}) has been largely stagnant---possibly due to substantial
        overlap in the class distributions.
        In fact, models are unable to distinguish between these pairs better
        than chance (cf. pairwise accuracy).}
\end{figure}

\paragraph{Confusing class pairs.}
We find that there are several pairs of ImageNet classes that
annotators have trouble telling apart---they consistently select both labels 
as valid for images of either class---see Figure~\ref{fig:amb_examples_pap}. 
On some of these pairs we see that models still perform well---likely because the search 
results from the automated image retrieval process are 
sufficiently error-free (i.e., the overlap in the image search results for the two 
classes is not significant) to disambiguate between these pairs.

However, for other pairs, even state-of-the-art models have poor accuracy (below 
40\%)---see Figure~\ref{fig:amb}.
In fact, we can attribute this poor performance to model confusion \emph{within} these  
pairs---\emph{none} of the models we examined do much better than chance at 
distinguishing between the two  classes. 
The apparent performance barrier on these \emph{ambiguous classes}  could thus be due 
to an inherent overlap in their ImageNet distributions.
It is likely that the automated image retrieval caused mixups in 
the images for the two classes, and annotators were unable to rectify these. 
If that is indeed the case, it is natural to wonder whether 
accuracy on ambiguous classes can be improved without overfitting to the ImageNet test 
set. 

The annotators' inability to remedy overlaps in the case of ambiguous class pairs could 
be due to the presence of classes that are semantically quite similar, 
e.g., ``rifle'' and ``assault rifle''---which is problematic under the \contains{} 
task as discussed before. In some cases, we find that there also were errors in the task 
setup.
For instance, there were occasional overlaps in the class
names (e.g.,  ``maillot'' and ``maillot, tank suit'') and Wikipedia links 
(e.g.,
``laptop computer'' and ``notebook computer'') presented to the annotators. 
This highlights that choosing labels that are in principle disjoint (e.g., using 
WordNet) might not be sufficient to ensure that the resulting dataset has 
non-overlapping classes---when using noisy validation pipelines, we need 
to factor human confusion into class selection and description.

\section{Beyond Test Accuracy: Human-In-The-Loop Model Evaluation}
\label{sec:human_eval}
Based on our analysis of the ImageNet dataset so far, it is clear that using top-1 accuracy 
as a standalone performance metric can be problematic---issues such as multi-object 
images and ambiguous classes make ImageNet labels an imperfect proxy for the ground 
truth.
Taking these issues into consideration, we now turn our focus to augmenting the model 
evaluation toolkit with metrics that are better aligned with the underlying goal of \task{}.

\subsection{Human assessment of model predictions}
\label{sec:reasonable}
\label{sec:human_judge}
To gain a broader understanding of model performance we start by \emph{directly} 
employing annotators to assess how good model predictions are.
Intuitively, this should not only help account for imperfections in
ImageNet labels but also to capture improvements in models (e.g., predicting 
a similar dog breed) that are not reflected in accuracy alone.
Specifically, given a model prediction for a specific image, we measure:
\begin{itemize}
    \item \textbf{Selection frequency of the prediction:} How often 
    annotators
    select the predicted label as being present in the image 
    (determined using the \contains{} task).
        Note that this metric accommodates for multiple
        objects or ambiguous classes as annotators will confirm all valid labels.  
        \item \textbf{Accuracy based on main
        label annotation:} How frequently the prediction matches the main
        label for the image, as
        obtained 
        using our \classify{} task (cf. Section~\ref{sec:followup}).  This metric 
        penalizes models that
        exploit dataset biases to predict ImageNet labels even when these 
        do not
        correspond to the main object.  At the same time, it only measures
        accuracy as non-experts perceive it---if annotators cannot distinguish
        between two classes (e.g., different dog breeds), models can do no 
        better than 
        random chance on this metric, even if their predictions actually match the ground 
        truth.
\end{itemize}

\begin{figure}[!h]
\begin{center}
\begin{subfigure}{0.45\textwidth}
	\centering
	\includegraphics[width=1\textwidth]{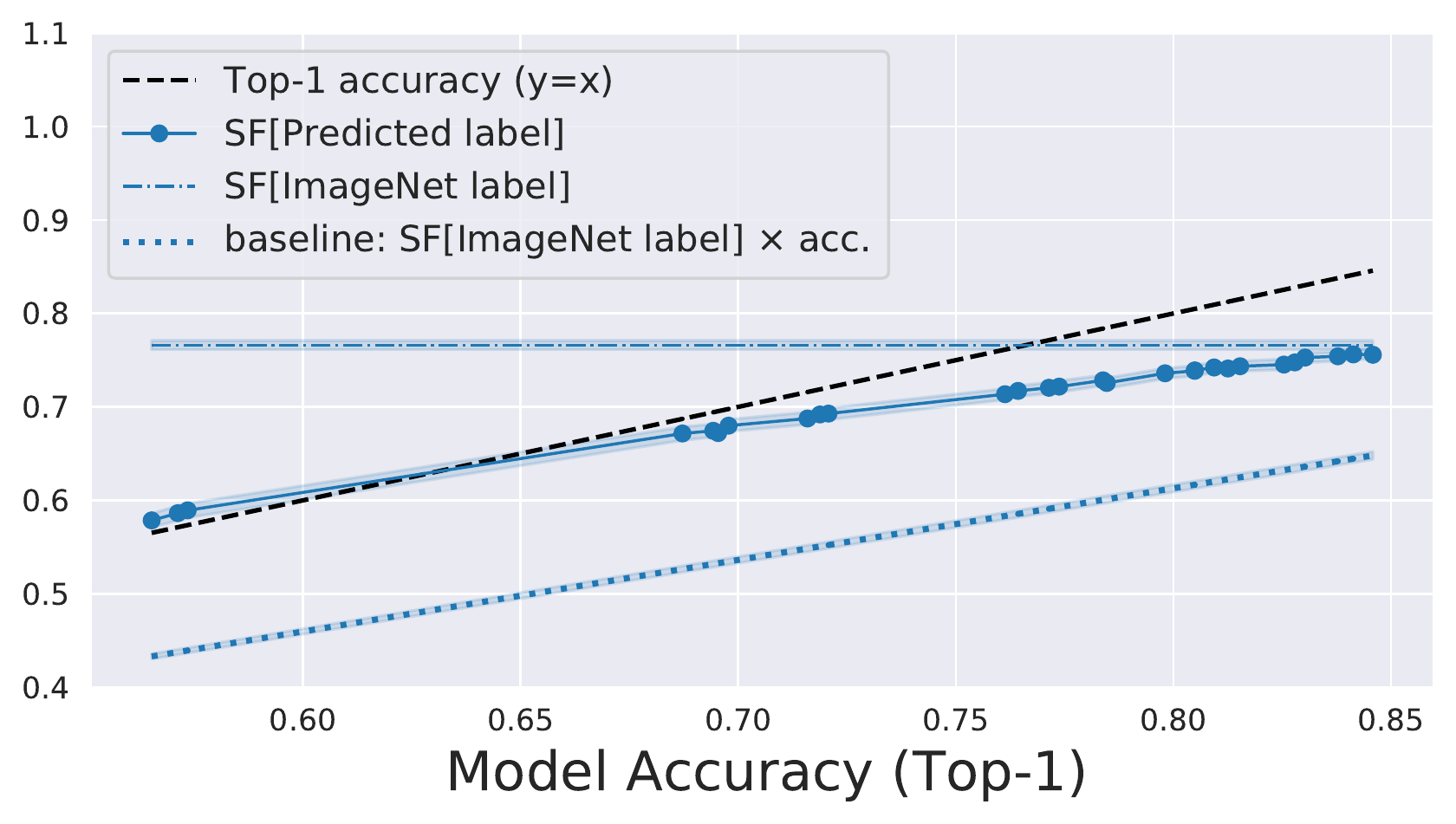}
	\caption{}
	\label{fig:limit_sf}
\end{subfigure}
\hfil
\begin{subfigure}{0.45\textwidth}
	\centering
	\includegraphics[width=1\textwidth]{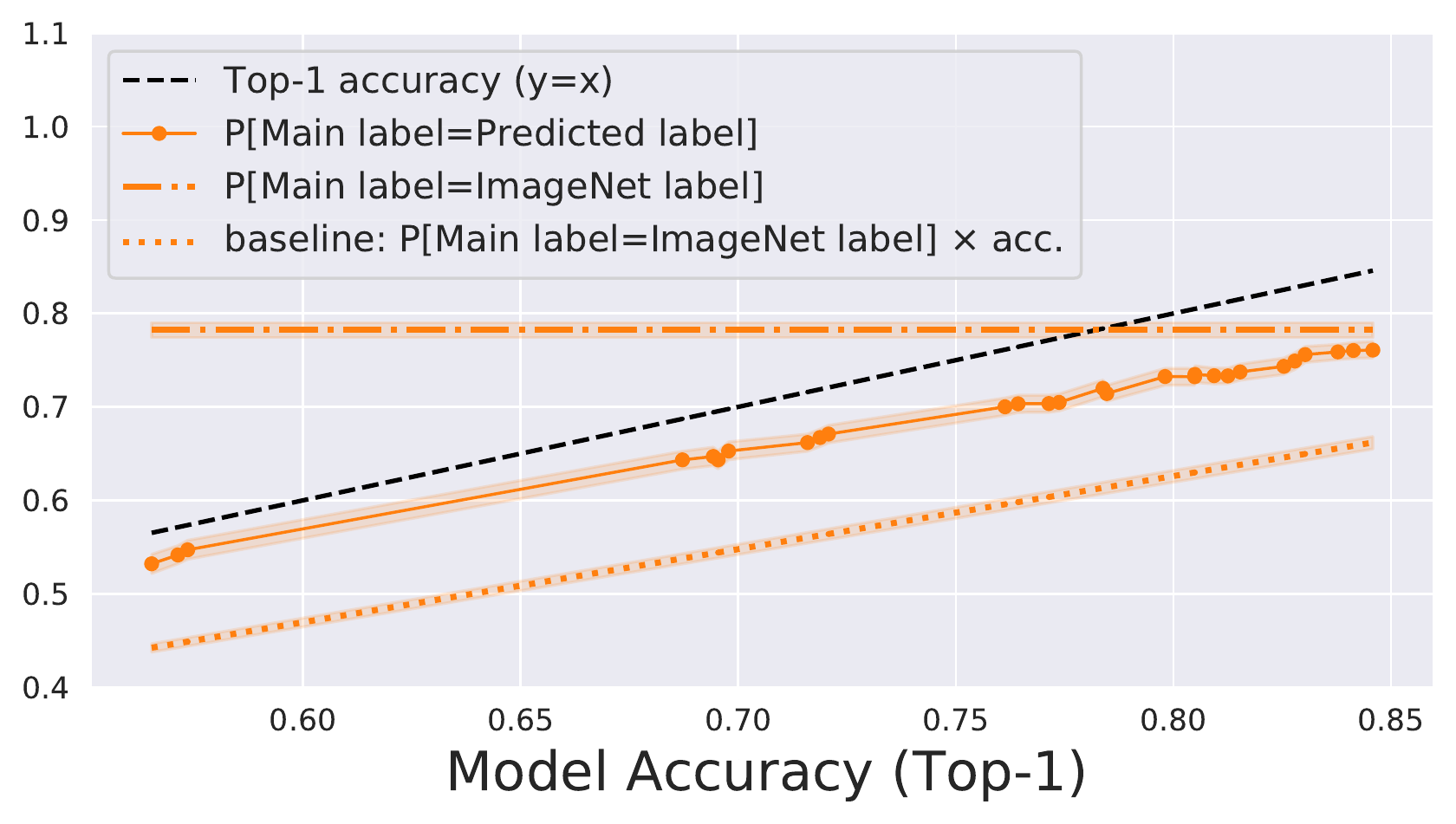}
	\caption{}
	\label{fig:limit_main}
\end{subfigure}
\end{center}
	\caption{Using humans to assess model predictions---we measure
		 how often annotators select the predicted/ImageNet label: (a)
		 to be contained in the image (\emph{selection
         frequency} [SF] from Section~\ref{sec:contains}), and (b) to denote the
         \emph{main} image object (cf. Section~\ref{sec:followup}), along with
         95\% confidence intervals via bootstrap (shaded).  We find that though
         state-of-the-art models have imperfect top-1 accuracy, their
         predictions are, on average, almost indistinguishable according to our
         annotators from the ImageNet labels themselves.}
	\label{fig:limit}
\end{figure}

Comparing model predictions and ImageNet labels along these axes would 
allow us to assess the (top-1 accuracy) gap in model performance from a 
human perspective. Concretely, we want to understand whether more 
accurate models also make higher-quality predictions, i.e., if the labels they 
predict (including  erroneous ones) also appear more reasonable to humans?
We find that models improve consistently along these axes as well (cf.  
Figure~\ref{fig:limit})---faster than  
improvements in accuracy can explain (i.e., more predictions matching the ImageNet 
label).
Moreover, we observe that the predictions of state-of-the-art models  
have gotten, on average, quite close to ImageNet labels with respect to 
these metrics.
That is, annotators are almost \emph{equally likely} to select the predicted 
label
as being present in the image (or being the main object)
as the ImageNet label.
This indicates that  
model predictions might be closer to what non-expert annotators 
can recognize as the ground truth than accuracy alone suggests.

These findings do not imply that all of the remaining gap between 
state-of-the-art model performance and perfect top-1 accuracy is 
inconsequential.
After all, for many images, the labels shown to annotators during the 
ImageNet creation process (based on automated data retrieval) could have  
been the ground truth.
In these cases, striving for higher accuracy on ImageNet would actually 
extend 
to real-world \task{} settings.
However, the results in Figure~\ref{fig:limit} hint at a different issue: we are 
at a point where we may no longer by able to easily identify (e.g., using 
crowd-sourcing) the extent to 
which further gains in
accuracy correspond to such improvements, as opposed to models simply
matching the ImageNet distribution better.

\begin{figure}[!h]
	\includegraphics[width=.49\textwidth]{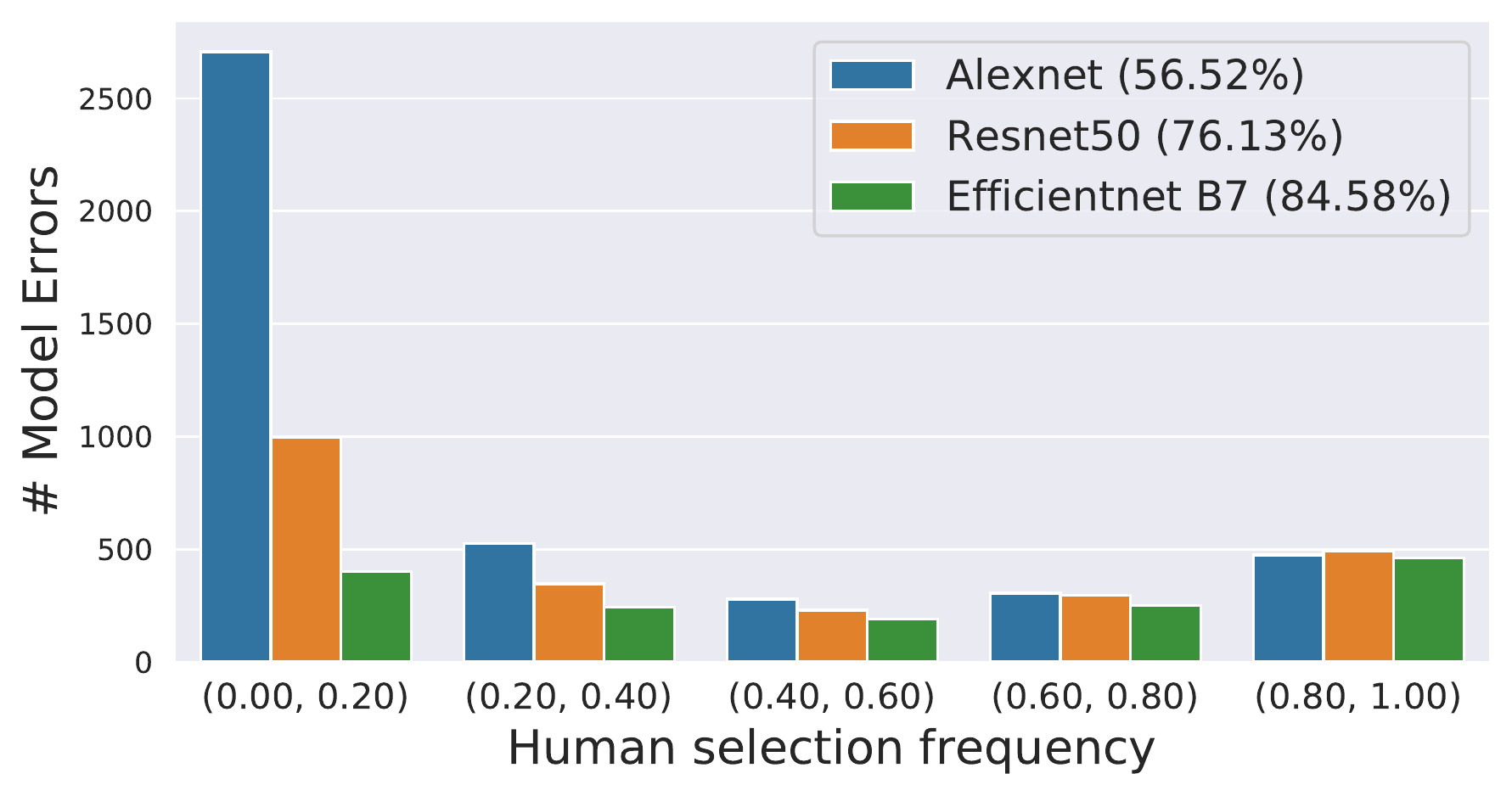}
    \hfill
	\includegraphics[width=.49\textwidth]{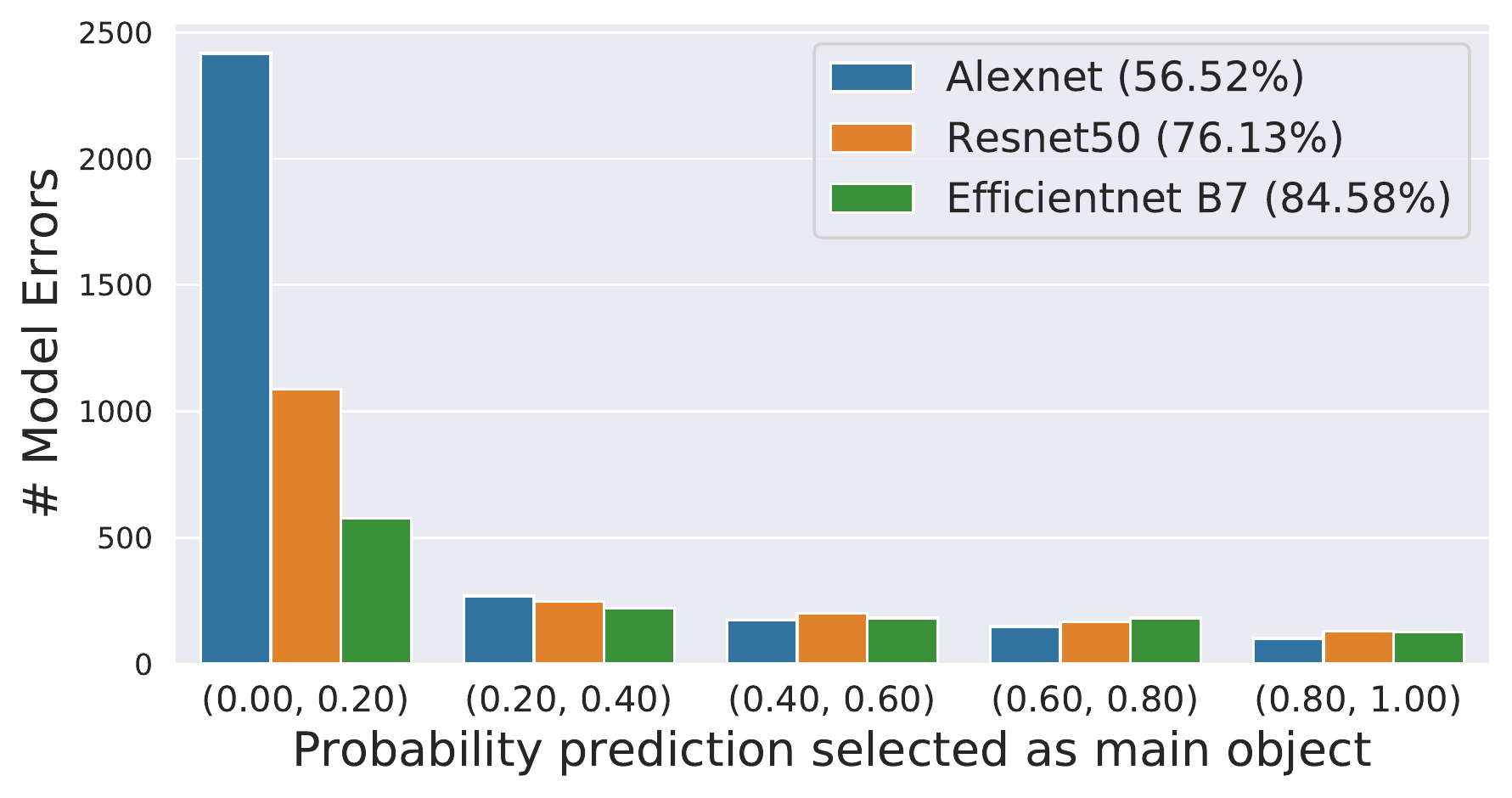}
    \caption{Distribution of annotator selection frequencies (cf.
        Section~\ref{sec:contains}) for model predictions deemed incorrect 
        w.r.t. the 
        ImageNet label.
        Models that are more accurate also seem to make fewer mistakes  that 
        have low
    human selection frequency (for the corresponding image-prediction pair).  
    }
	\label{fig:sf_errors}
\end{figure}

\paragraph{Incorrect predictions.}
We can also use these metrics to examine model mistakes, i.e., 
predictions that deviate from the ImageNet label, more closely.
Specifically, we can treat human assessment of these labels (based on the 
metrics above) as a proxy for how much these predictions deviate from
the ground truth.
We find that recent, more accurate ImageNet models make progressively 
fewer mistakes that would be judged by humans as such (i.e., with
low selection frequency or low probability of being the main object)---see 
Figure~\ref{fig:sf_errors}.
This indicates that models are not only getting better at predicting the
ImageNet label, but are also making fewer blatant mistakes, at least according 
to \emph{non-expert} humans.
At the same time, we observe that for \emph{all} models,
a large fraction of the seemingly incorrect predictions are actually valid
labels
according to human annotators. This suggests that using a single ImageNet
label alone to determine the correctness of these predictions, may, at
times, be pessimistic
(e.g., multi-object images or ambiguous classes).

When viewed from a different perspective, Figure~\ref{fig:sf_errors} also 
highlights the pitfalls of using selection frequency as 
the sole filtering criteria during dataset collection. 
Images with high selection frequency (w.r.t., the dataset label) can still be 
challenging for models (cf.  Appendix Figure~\ref{fig:scatter}).

\subsection{Fine-grained comparison of human vs.\ model performance}
\label{sec:similarity}
Next, we can get a more fine-grained understanding of model predictions by 
comparing them to the ones humans make.
To do this, we use the image annotations we collect (cf. 
Section~\ref{sec:method})---specifically the main label---as a proxy for 
(non-expert) human predictions.
(Note that, the human prediction matches the ImageNet label only
on about 80\% of the images, despite the fact that annotators are 
directly presented with a few relevant labels (including the 
ImageNet label) to choose from---see Figure~\ref{fig:limit_main}.)

\paragraph{Confusion matrix.} We start by comparing the confusion 
matrices for models and humans.
Since it is hard to manually inspect the full 1000-by-1000 matrices (cf. 
Appendix~\ref{app:confusion}), we
partition ImageNet classes into 11 superclasses 
(Appendix~\ref{app:superclasses}).
This allows us to study confusions \emph{between} and \emph{within} superclasses
separately.

\begin{figure*}[!h]
	\begin{center}
	\begin{subfigure}[b]{0.7\textwidth}
	\centering
	\includegraphics[width=0.88\textwidth]{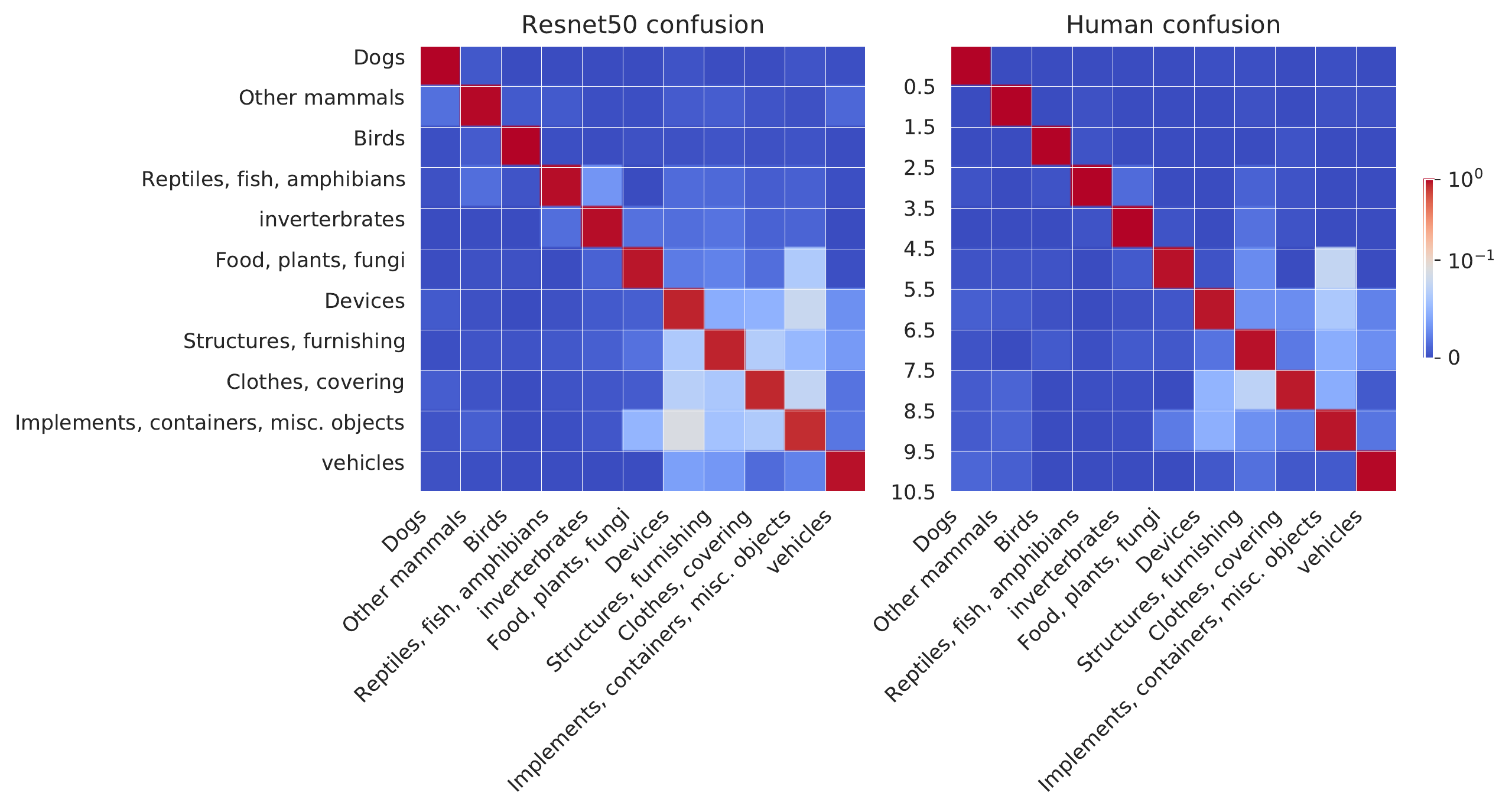}
	\caption{Confusion matrices}
	\label{fig:high_level_conf_r50}
\end{subfigure}
\hfill
\begin{subfigure}[b]{0.295\textwidth}
	\includegraphics[width=0.86\textwidth]{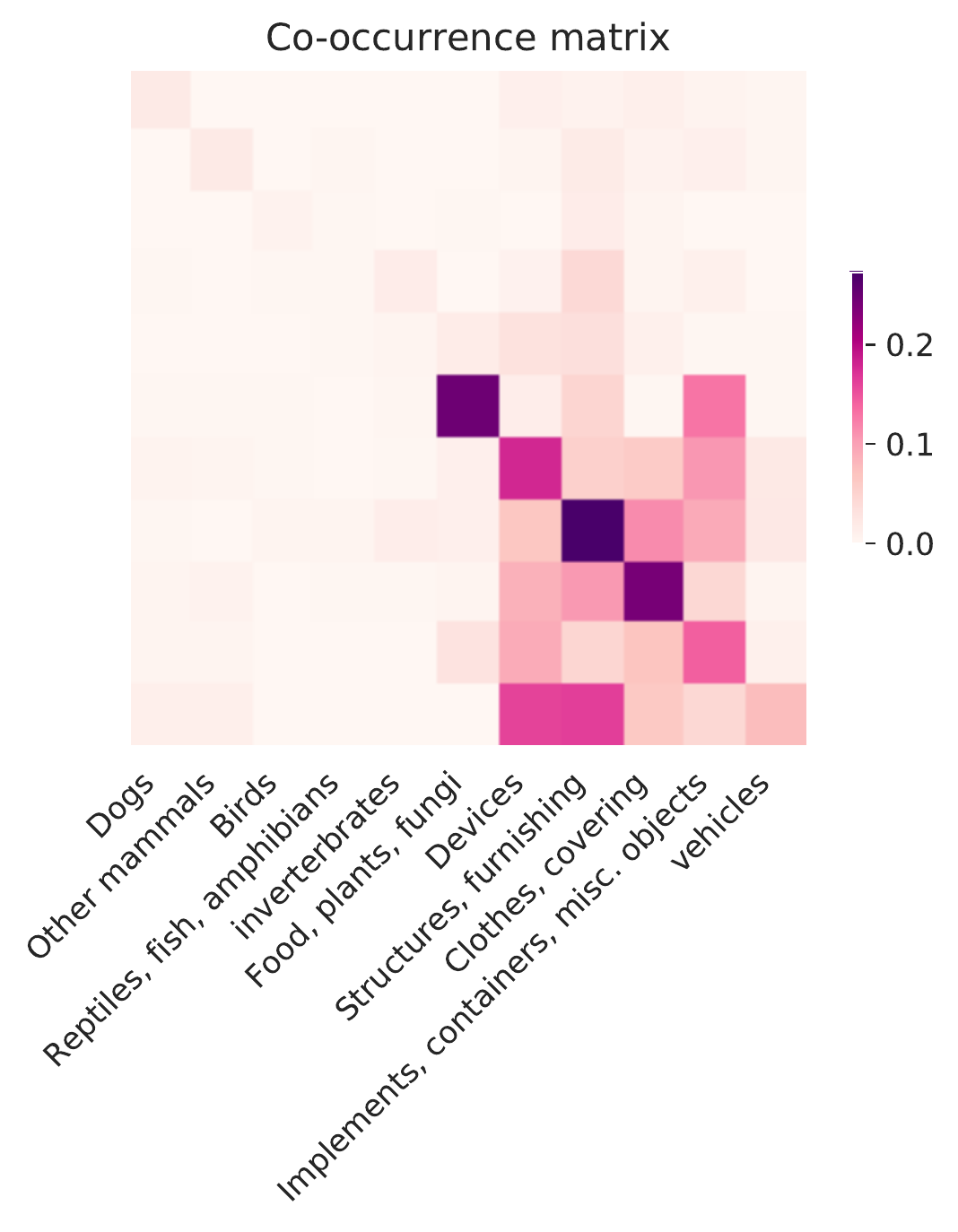}
	\caption{Co-occurrence matrix}
	\label{fig:cooccurrence}
\end{subfigure}
	\end{center}
    \caption{Characterizing similarity between model and human predictions
        (indicated by main object selection in the \contains{} task):
		(a) Model (ResNet-50) and human confusion matrices on 11 ImageNet
		superclasses (cf. Section~\ref{app:superclasses}).
		(b) Superclass co-occurrence matrix: how likely a specific pair of
		superclasses is to occur together (based on annotation from
		Section~\ref{sec:followup}).}
	\label{fig:sc_conf}
\end{figure*}

In the cross-superclass confusion matrix
(cf. Figure~\ref{fig:high_level_conf_r50}), we observe a block where both 
human 
and model is high---this is particularly striking given that 
these superclasses are semantically quite different.
To understand these confusions better, we compare the superclass confusion
matrix with the superclass \emph{co-occurrence} matrix, i.e., how often an 
image of
superclass $i$ (w.r.t. the ImageNet label) also contains an object of
superclass $j$ according to annotators (cf.  Figure~\ref{fig:cooccurrence}).
Indeed, we find that the two matrices are quite aligned---indicating 
that model and human
confusion between superclasses might be stemming from multi-object 
images.
We also observe that the {intra-superclass} confusions are more significant for
humans compared to models (cf. Appendix
Figure~\ref{fig:high_level_conf_b7_intra}),  particularly on  fine-grained
classes (e.g., dog breeds).

%

\section{Related Work}

\paragraph{Identifying ImageNet issues.}
Some of the ImageNet label issues we study have been already been identified in
prior work.
Specifically, \citet{recht2018imagenet,northcutt2019confident,
hooker2019selective} demonstrate the existence of classes that might be
inherently ambiguous (similar to our findings in Section~\ref{sec:nonexp}).
Moreover, the existence of cluttered images was discussed by
\citet{russakovsky2015imagenet} as an indication that the dataset mirrors
real-world conditions---hence being desirable---and  by
\citet{stock2018convnets,northcutt2019confident,hooker2019selective} as a source
of label ambiguity (similar to our findings in Section~\ref{sec:multi}).
Additionally, \citet{stock2018convnets}, motivated by the multi-label nature of
images, perform human-based model evaluation, similar to our experiments
in Section~\ref{sec:reasonable}.
Finally, \citet{stock2018convnets} use manual data collection and saliency maps
to identify racial biases that models rely on to make their predictions.
However, the focus of all these studies is not to characterize ImageNet issues
and hence they only provide coarse estimates for their pervasiveness.
In particular, none of these studies evaluate how these issues affect model
performance, nor do they obtain annotations that are sufficient to rigorously
draw per-class conclusions (either due to granularity or scale).

\paragraph{Human performance on ImageNet.}
Rigorously studying human accuracy on the full ImageNet classification task is
challenging, mainly due to the fact that annotators need to be mindful of all
the 1000 ImageNet classes that are potentially present.
The original ImageNet challenge paper contained results for two
trained annotators~\citep{russakovsky2015imagenet}, while
\citet{karpathy2014what} reports result based on evaluating themselves.
An MTurk study using a subset of the ImageNet classes is presented in
\citet{dodge2017study}.

\paragraph{Generalization beyond the test set.}
The topic of designing large-scale vision datasets that allow generalization
beyond the narrow benchmark task has been an active topic of discussion in the
computer vision
community~\citep{ponce2006dataset,everingham2010pascal,torralba2011unbiased,russakovsky2015imagenet}.
\citet{torralba2011unbiased} propose evaluating cross-dataset
generalization---testing the performance of a model on a different dataset which
shares a similar class structure.
\citet{recht2018imagenet} focus on reproducing the ImageNet validation
set process to measure potential adaptive overfitting or over-reliance on the
exact dataset creation conditions.
\citet{kornblith2019better} investigate the extent to which better ImageNet
performance implies better feature extraction as measured by the suitability of
internal model representations for transfer
learning~\citep{donahue2014decaf,sharif2014cnn}.

\paragraph{Adversarial testing.}
Beyond the benign shift in test distribution discussed above, there has been
significant work on measuring model performance from a worst-case perspective.
A large body of work on adversarial
examples~\citep{biggio2013evasion,szegedy2014intriguing} demonstrates that
models are extremely brittle to imperceptible pixel-wise perturbations---every
image can be arbitrarily misclassified.
Similar, yet less severe, brittleness can be observed for a number of more
natural perturbations, such as: worst-case spatial transformation (e.g,
rotations)~\citep{fawzi2015manitest,engstrom2019rotation}, common image
corruptions~\citep{hendrycks2019benchmarking}, adversarial texture
manipulation~\citep{geirhos2018imagenettrained}, adversarial data
collection~\citep{barbu2019objectnet,hendrycks2019natural}.

\section{Conclusion}
\label{sec:discussion}
\label{sec:discussion_gt}

In this paper, we take a step towards understanding how closely widely-used 
vision benchmarks align with the real-world tasks they are meant to 
approximate, focusing on the ImageNet \task{} dataset.
Our analysis uncovers systemic---and fairly pervasive---ways in which
ImageNet  annotations deviate from the ground truth---such as the wide
presence of images with multiple valid labels, and ambiguous classes. 

Crucially, we find that these deviations significantly impact what ImageNet-trained
models learn (and don't learn), and  how we perceive model progress.
For instance, top-1 accuracy often underestimates the performance of 
models by
unduly penalizing them for predicting the label of a different, but also valid,
image object.
Further, current models seem to derive part of their accuracy by exploiting 
ImageNet-specific features that humans are oblivious to, and hence may not 
generalize to the real world.
Such issues make it clear that measuring accuracy alone may give us only an 
imperfect view of model performance on the underlying \task{} task. 

Taking a step towards evaluation metrics that circumvent these issues, we
utilize human annotators to directly judge the correctness of model predictions.
On the positive side, we find that models that are more accurate on
ImageNet also tend to be more human-aligned in their errors.
In fact, on average, (non-expert) annotators turn out to be unable to
distinguish the (potentially incorrect) predictions of state-of-the-art models from the ImageNet labels
themselves.
While this might be reassuring, it also indicates that we are at a point where we 
cannot easily gauge (e.g., via simple crowd-sourcing) whether further progress on
the ImageNet benchmark is meaningful, or is simply a result of overfitting to
this benchmarks' idiosyncrasies.

More broadly, our findings highlight an inherent conflict between the goal of
building rich datasets that capture complexities of the real world and the 
need
for the annotation process to be scalable.
Indeed, in the context of ImageNet, we found that some of the very
reasons that make the collection pipeline scalable  (e.g., resorting to the
\contains{} task, employing non-expert annotators) were also at the crux of the
afore-mentioned annotation issues.
We believe that developing annotation pipelines that better capture the ground
truth while remaining scalable is an important avenue for future research.

\section*{Acknowledgements}
Work supported in part by the NSF grants CCF-1553428, CNS-1815221, the Google
PhD Fellowship, the Open Phil AI Fellowship, and the Microsoft Corporation.

Research was sponsored by the United States Air Force Research Laboratory and
was accomplished under Cooperative Agreement Number FA8750-19-2-1000. The views
and conclusions contained in this document are those of the authors and should
not be interpreted as representing the official policies, either expressed or
implied, of the United States Air Force or the U.S. Government. The U.S.
Government is authorized to reproduce and distribute reprints for Government
purposes notwithstanding any copyright notation herein.

\small 
\printbibliography

\clearpage
\normalsize 
\appendix
\section{Experimental setup}
\label{app:setup}

\subsection{Datasets}
\label{app:datasets}
We perform our analysis on the ImageNet dataset~\cite{russakovsky2015imagenet}. A full 
description of the data creation process can be found in \citet{deng2009imagenet} and
\citet{russakovsky2015imagenet}. For the purposes of our human studies, we use a 
random subset of  the validation set---10,000 images chosen by sampling 10 random 
images from each of the 1,000 classes. (Model performance on this subset closely mirrors 
overall test accuracy, as shown in Figure~\ref{fig:multi_accuracy}.) In our analysis, we 
refer to the original dataset labels as ``ImageNet labels'' or ``IN labels''.

\subsubsection{ImageNet superclasses} 
\label{app:superclasses}
We construct 11 superclasses by manually grouping together semantically similar 1000 
ImageNet classes (in parenthesis), with the assistance of the WordNet 
hierarchy---\emph{Dogs} (130), 
\emph{Other mammals} (88), 
\emph{Bird} (59), 
\emph{Reptiles, fish, amphibians} (60), 
\emph{Inverterbrates} (61), 
\emph{Food, plants, fungi} (63), 
\emph{Devices} (172), 
\emph{Structures, furnishing} (90), 
\emph{Clothes, covering} (92), 
\emph{Implements, containers, misc. objects} (117), 
\emph{Vehicles} (68).

\subsection{Models}
\label{app:models}

We perform our evaluation on various standard ImageNet-trained models---see Appendix 
Table~\ref{tab:models} for a full list.
We use open-source pre-trained implementations from 
\url{github.com/Cadene/pretrained-models.pytorch} and/or 
\url{github.com/rwightman/pytorch-image-models/tree/master/timm} for all
architectures.

\begin{table}[!h]
	\caption{Models used in our analysis with the corresponding 
		ImageNet top-1/5 accuracies.} 
	\label{tab:models}
	\begin{minipage}{0.5\textwidth}
		\begin{center}
			\begin{tabular}{lccc}
	\toprule
	\textbf{Model} & \phantom{x} & \textbf{Top-1} & \textbf{Top-5} \\
	\midrule
	\texttt{alexnet}~\citep{krizhevsky2012imagenet} && 56.52 & 79.07 \\ 
	\texttt{squeezenet1\_1}~\citep{iandola2016squeezenet} && 57.12 & 80.13 \\ 
	\texttt{squeezeNet1\_0}~\citep{iandola2016squeezenet} && 57.35 & 79.89 \\ 
	\texttt{vgg11}~\citep{simonyan2015very} && 68.72 & 88.66 \\ 
	\texttt{vgg13}~\citep{simonyan2015very} && 69.43 & 89.03 \\ 
	\texttt{inception\_v3}~\citep{szegedy2016rethinking} && 69.54 & 88.65 \\ 
	\texttt{googlenet}~\citep{szegedy2015going} && 69.78 & 89.53 \\ 
	\texttt{vgg16}~\citep{simonyan2015very} && 71.59 & 90.38 \\ 
	\texttt{mobilenet\_v2}~\citep{sandler2018mobilenetv2} && 71.88 & 90.29 \\ 
	\texttt{vgg19}~\citep{simonyan2015very} && 72.07 & 90.74 \\ 
	\texttt{resnet50}~\citep{he2016deep} && 76.13 & 92.86 \\ 
	\texttt{efficientnet\_b0}~\citep{tan2019efficientnet} && 76.43 & 93.05 \\ 
	\texttt{densenet161}~\citep{huang2017densely} && 77.14 & 93.56 \\ 
	\texttt{resnet101}~\citep{he2016deep} && 77.37 & 93.55 \\  
	\bottomrule
			\end{tabular}
		\end{center}
	\end{minipage}
\begin{minipage}{0.5\textwidth}
	\begin{center}
		\begin{tabular}{lccc}
			\toprule
			\textbf{Model} & \phantom{x} & \textbf{Top-1} & \textbf{Top-5} \\
			\midrule
			\texttt{efficientnet\_b1}~\citep{tan2019efficientnet} && 78.38 & 94.04 \\ 
			\texttt{wide resnet50\_2}~\citep{zagoruyko2016wide} && 78.47 & 94.09 \\ 
			\texttt{efficientnet\_b2}~\citep{tan2019efficientnet} && 79.81 & 94.73 \\ 
			\texttt{gluon\_resnet152\_v1d}~\citep{he2019bag} && 80.49 & 95.17 \\ 
			\texttt{inceptionresnetv2}~\citep{szegedy2017inception} && 80.49 & 95.27 \\ 
			\texttt{gluon\_resnet152\_v1s}~\citep{he2019bag} && 80.93 & 95.31 \\ 
			\texttt{senet154}~\citep{hu2018squeeze} && 81.25 & 95.30 \\ 
			\texttt{efficientnet\_b3}~\citep{tan2019efficientnet} && 81.53 & 95.65 \\ 
			\texttt{nasnetalarge}~\citep{zoph2018learning} && 82.54 & 96.01 \\ 
			\texttt{pnasnet5large}~\citep{liu2018progressive} && 82.79 & 96.16 \\ 
			\texttt{efficientnet\_b4}~\citep{tan2019efficientnet} && 83.03 & 96.34 \\ 
			\texttt{efficientnet\_b5}~\citep{tan2019efficientnet} && 83.78 & 96.71 \\ 
			\texttt{efficientnet\_b6}~\citep{tan2019efficientnet} && 84.13 & 96.96 \\ 
			\texttt{efficientnet\_b7}~\citep{tan2019efficientnet} && 84.58 & 97.00 \\ 
			\bottomrule
		\end{tabular}
	\end{center}
\end{minipage}
\end{table}

\clearpage
\section{Obtaining Image Annotations}
\label{app:methodology}

Our goal is to use human annotators to obtain labels for each distinct object  in ImageNet 
images (provided it corresponds to a valid ImageNet class). To make this classification 
task feasible, we first identify a small set of relevant candidate labels per image to present 
to annotators.

\subsection{Obtaining candidate labels}
\label{app:contains}
As discussed in Section~\ref{sec:contains}, we narrow down the candidate labels 
for each image by (1) restricting to the predictions of a set of pre-trained ImageNet 
models, and then (2) repeating the \contains{} task (cf. Section~\ref{sec:pipeline}) on 
human annotators using the labels from (1) to identify the most reasonable ones.

\subsubsection{Pre-filtering using model predictions} 
\begin{figure}[!h]
	
	\centering
	\includegraphics[width=0.4\textwidth]{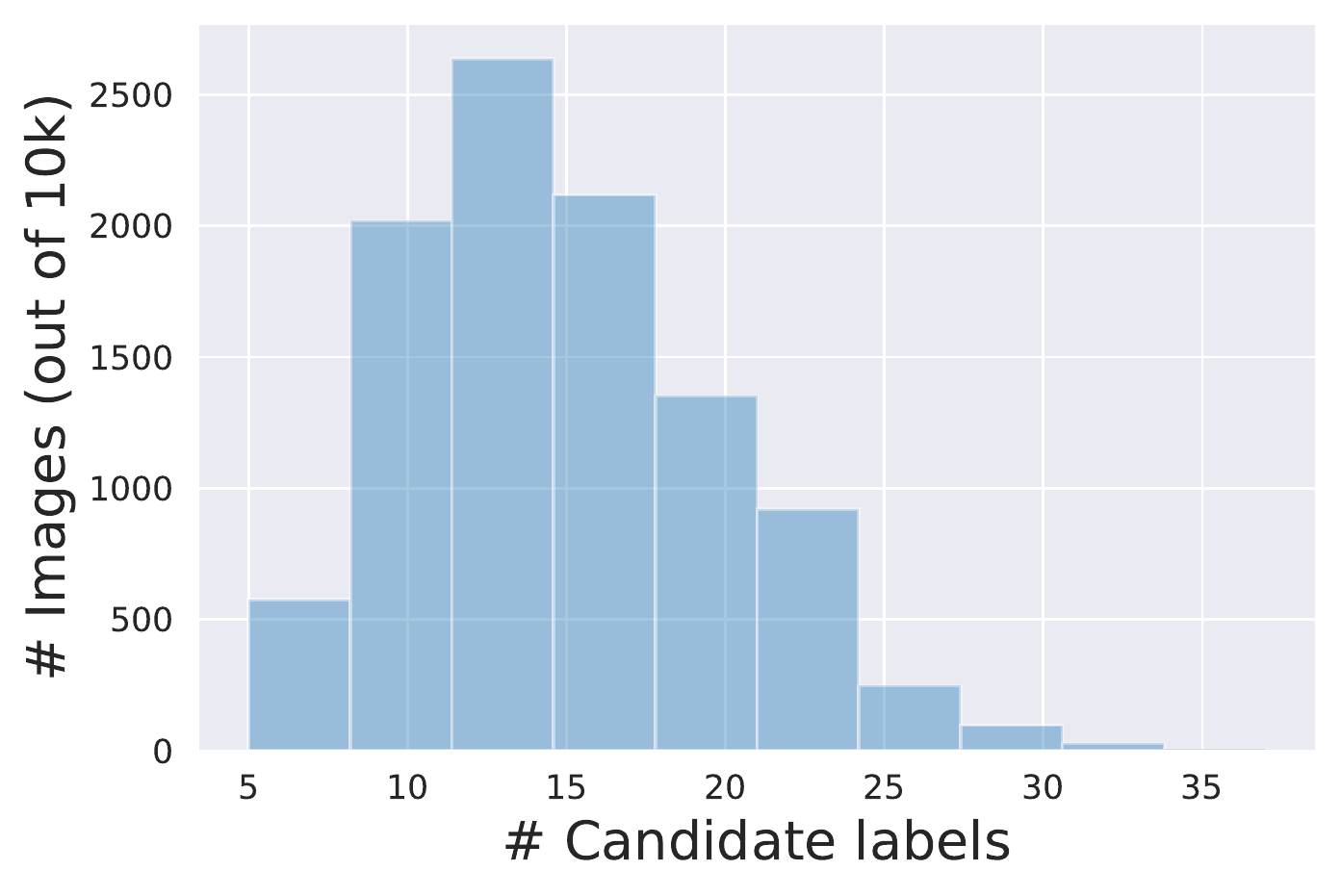}
	\caption{Distribution of labels per image obtained from the predictions of 
	ImageNet-trained models (plus the ImageNet label). We present these labels 
	(in separate grids) to annotators via the \contains{} task (cf. 
	Section~\ref{sec:contains})to identify a small set of 
	relevant candidate labels for the classification task in 
	Section~\ref{sec:followup}.}
	\label{fig:cand_label_dist}
\end{figure}

We use the top-5 predictions of
models with varying ImageNet (validation) 
accuracies (10 in total): \texttt{alexnet}, \texttt{resnet101}, \texttt{densenet161}, 
\texttt{resnet50}, \texttt{googlenet}, \texttt{efficientnet\_b7} \texttt{inception\_v3},
\texttt{vgg16}, \texttt{mobilenet\_v2}, \texttt{wide\_resnet50\_2} (cf.
Table~\ref{tab:models}) to identify a set of \emph{potential labels}.
Since model predictions tend to overlap, we end up with
$\sim 14$ potential labels per image on average (see full histogram in
Figure~\ref{fig:cand_label_dist}).
We {always} include the ImageNet label in the set of potential labels, 
even if it is absent in all the model predictions.

\subsubsection{Multi-label validation task}
\label{app:mturk}
We then use human annotators to go through these potential labels and identify the most 
reasonable ones via the \contains{} task.
Recall that in \contains{}  task, annotators are shown a grid of images  
and asked to select the ones that contain an object corresponding to the specified 
query label  (cf. Section~\ref{sec:pipeline})
In our case, each image appears in multiple such grids---one for each potential label.
By presenting these grids to multiple annotators, we can then obtain a selection 
frequency for every image-potential label pair, i.e., the number of annotators that perceive 
the label as being contained in the image (cf. Figure~\ref{fig:confident}). Using these 
selection frequencies, we 
identify the most relevant \emph{candidate labels} for each image.

\begin{figure}[!h]	
	\centering
	\includegraphics[width=0.75\textwidth]{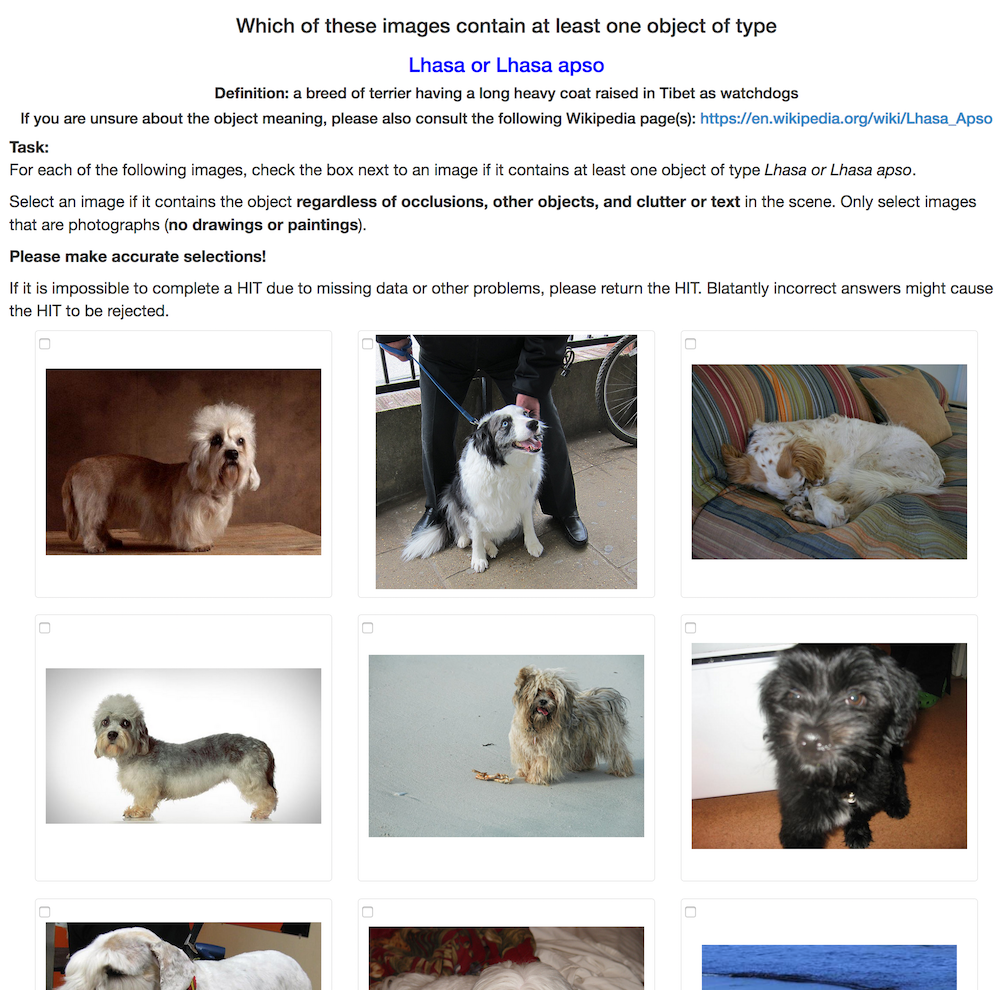}
	\caption{Sample interface of the \contains{} task we use for label validation:
		annotators are shown a grid of 48 images and asked to
		select all images that correspond to a specific label 
		(Section~\ref{sec:contains}). }
	\label{fig:contains_screen}
\end{figure}

\paragraph{Grid setup.} The grids used in our study contains 48 images, at least 5 of 
which are 
\emph{controls}---obtained by randomly sampling from validation set images labeled  as 
the {query} class. Along with the images, annotators are provided with a description of 
the query  label in terms of (a) WordNet synsets and (b) the relevant Wikipedia link---see 
Figure~\ref{fig:contains_screen} for an example.
(Our MTurk interface is based on a modified version of the code made publicly available by
\citet{recht2018imagenet}\footnote{\url{https://github.com/modestyachts/ImageNetV2}}.)
We find that a total of 3, 934 grids suffice to obtain selection frequencies for all 10k 
images used in our analysis (w.r.t. all  potential 
labels).
Every grid was shown to 9 annotators, compensated \$0.20 per task.

\paragraph{Quality control.}
We filtered low-quality responses on a per-annotator and per-task basis. First, we 
completely omitted results from annotators who selected less than 20\% of the control 
images on half or more of
the tasks they completed: a total of 10 annotators and the corresponding 513
tasks. Then we omitted tasks for which less than 40\% of the controls were
selected: at total of 3,104 tasks. Overall, we omitted 3,617 tasks in total out
of the total 35,406. As a result, the selection frequency of some
image-label pairs will be computed with fewer than 9 annotators. 

\subsubsection{Final candidate label selection} 
\begin{figure}[!h]
	
	\centering
	\includegraphics[width=0.41\textwidth]{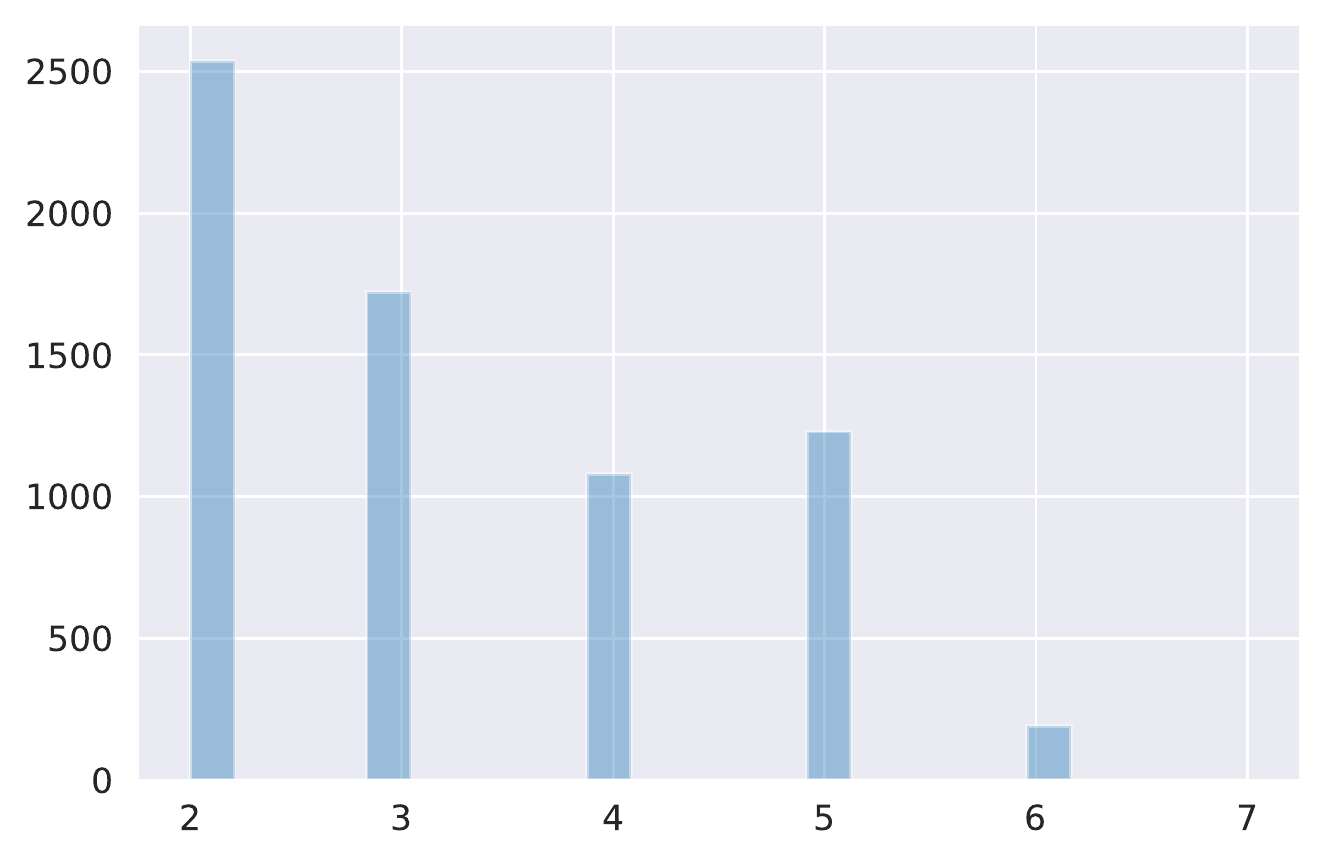}
	\caption{Distribution of the number of candidate labels used per image presented to 
	annotators during the 
		classification task in Section~\ref{sec:followup_specs}.}
	\label{fig:cand_label_dist_fu}
\end{figure}

We then obtain the most relevant candidate labels by selecting the potential 
labels with high human selection frequency. 
To construct this set, we consider (in order):
\begin{enumerate}
	\item The existing ImageNet label, irrespective of its selection frequency.
	\item All the highly selected potential labels: for which annotator selection 
	frequency is at least 0.5.
	\item All potential labels with non-zero selection frequency that are
	semantically very different from the ImageNet label---so as to
	include labels that may correspond to different objects.
	Concretely, we select candidate labels that are more than 5 nodes
	away from the ImageNet label in the WordNet graph.
	\item If an image has fewer than $5$ candidates, we also consider other potential 
	labels after sorting them based on their selection frequency (if non-zero). 
	\item To keep the number of candidates relatively small, we truncate the resulting set 
	size to $6$ if the excess labels have selection 
	frequencies lower than the ImageNet label, or the ImageNet label itself has 
	selection frequency $\leq 1/8$. During this truncation, we explicitly ensure 
	that the ImageNet label is retained.
\end{enumerate}
In Figure~\ref{fig:cand_label_dist_fu}, we visualize the distribution of number 
of candidate labels per image, over the set of images.

\subsection{Image classification}
\label{app:followup}
\begin{figure}[!h]
	
	\centering
	\includegraphics[width=0.7\textwidth]{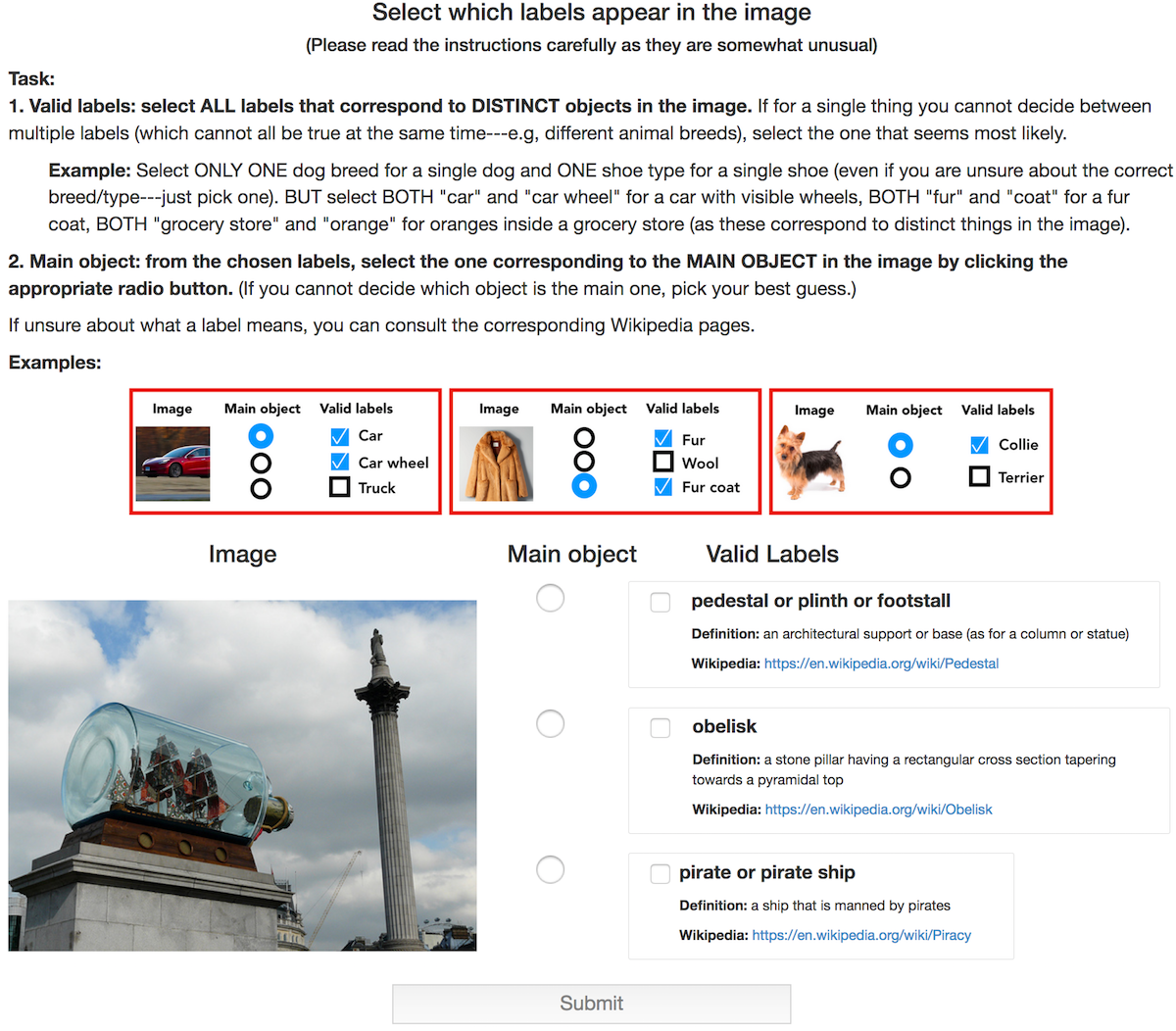}
    \caption{Screenshot of a sample image annotation task.
        (Section~\ref{sec:followup_specs}).
        Annotators are presented with an image and multiple candidate labels.
        They are asked to select all valid labels (selecting only one of
        mutually exclusive labels in the case of confusion) and indicate the
    main object of the image.}

	\label{fig:main_screen}
\end{figure}

The candidate labels are then presented to annotators during the 
\classify{} task (cf. Section~\ref{sec:followup_specs}).
Specifically, annotators are shown images, and their corresponding candidate 
labels and asked to select: a) all valid labels for
that image, b) a label for the main object of
the image---see Figure~\ref{fig:main_screen} for a sample task interface.
We instruct annotators to pick multiple labels as valid, only if they correspond
to different objects 
in the image and are not mutually exclusive.
In particular, in case of confusion about a specific object label, we explicitly
ask them to pick a single label making their best guess.
Each task was presented to 9 annotators, compensated at \$0.08 per
task.

\paragraph{Images included.} 
We only conduct this experiment on images that annotators identified as having  \emph{at 
least} one candidate label outside the existing ImageNet label (based on experiment in 
Appendix~\ref{app:contains}).
To this end, we omitted images for which the ImageNet label was clearly the most
likely: out of all the labels seen by 6 or more of the 9 annotators, the
original label had more than double the selection frequency of any other class. Note 
that since we discard some tasks as part of quality control, it is possible that for 
some image-label pairs, we have the results of fewer than 9 annotators.
Furthermore, we also omitted images which were not selected by any annotator as 
containing their ImageNet label ($150$ images
total)---cf. Appendix Figure~\ref{fig:mis_10k} for examples.
These likely corresponds to labeling mistakes in the dataset creation process and do
not reflect the systemic error we aim to study.
The remaining $6,761$ images that are part of our follow-up study have at least 1 label, 
in addition to the ImageNet label, that annotators think could be valid.

\paragraph{Quality control.} Performing stringent quality checks for this task is 
challenging since we do not have ground truth annotations to compare
against---which was after all the original task motivation.
Thus, we instead perform basic sanity checks for quality control---we ignore tasks where 
annotators did not 
select \emph{any} valid labels or selected a main label that they did not indicate as 
valid.  In addition, if the tasks of specific annotators are consistently flagged based 
on these criteria (more than a third of the tasks), we ignore all their
annotations. Overall, we omitted 1,269 out of the
total 59,580 tasks. \\

The responses of multiple annotators are aggregated as described in
Section~\ref{sec:followup}.

\clearpage
\section{Additional experimental results}
\label{app:dissect}

\subsection{Multi-object Images}
\label{app:clutter}
\begin{figure}[!h]
	\centering
	\includegraphics[width=1\textwidth]{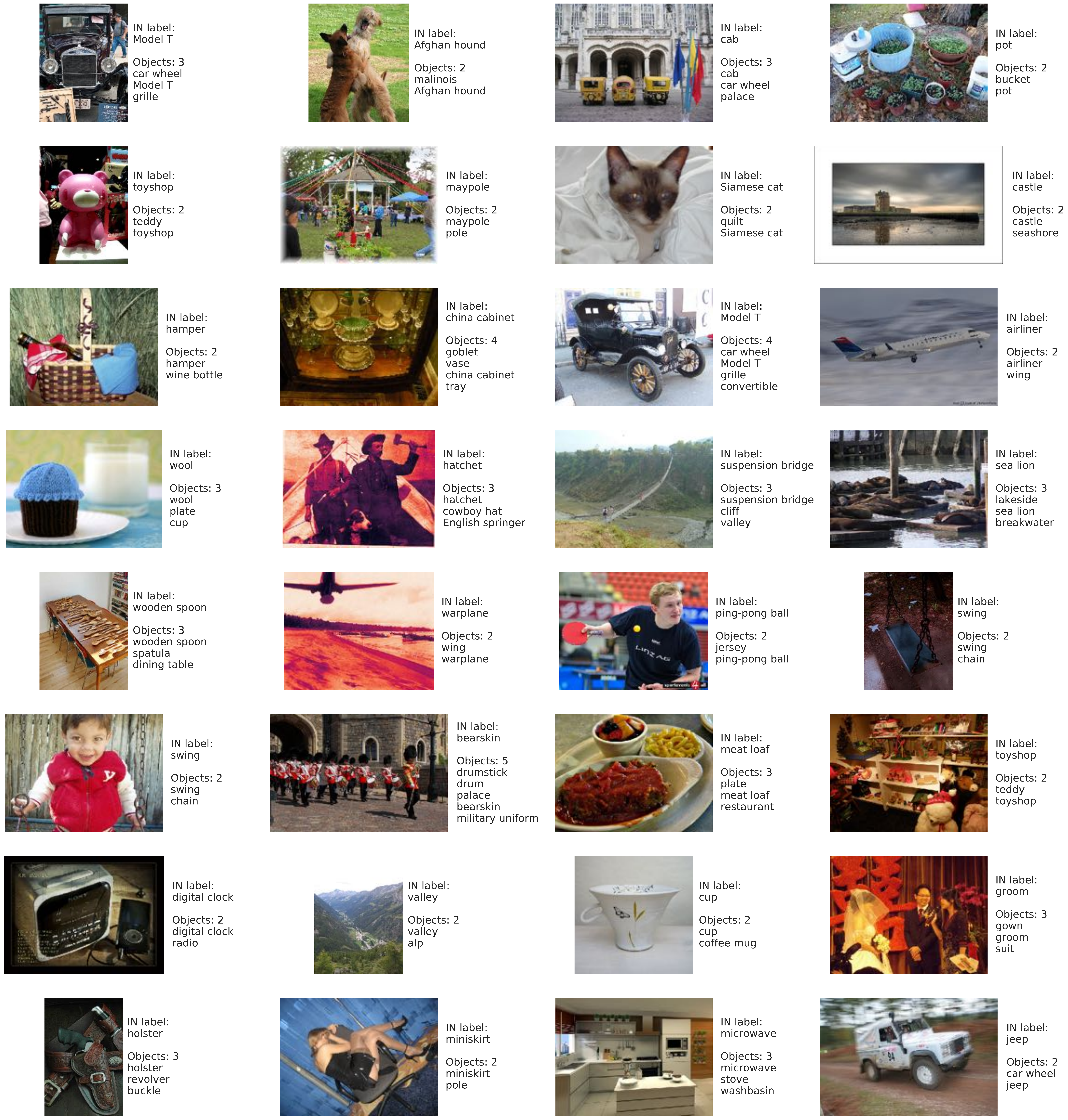}
	\caption{Sample ImageNet images with more than one valid label as per 
		human annotators (cf. Section~\ref{sec:contains}).}
	\label{fig:multi_examples}
\end{figure}

\begin{figure}[!h]
	\centering
	\includegraphics[width=1\textwidth]{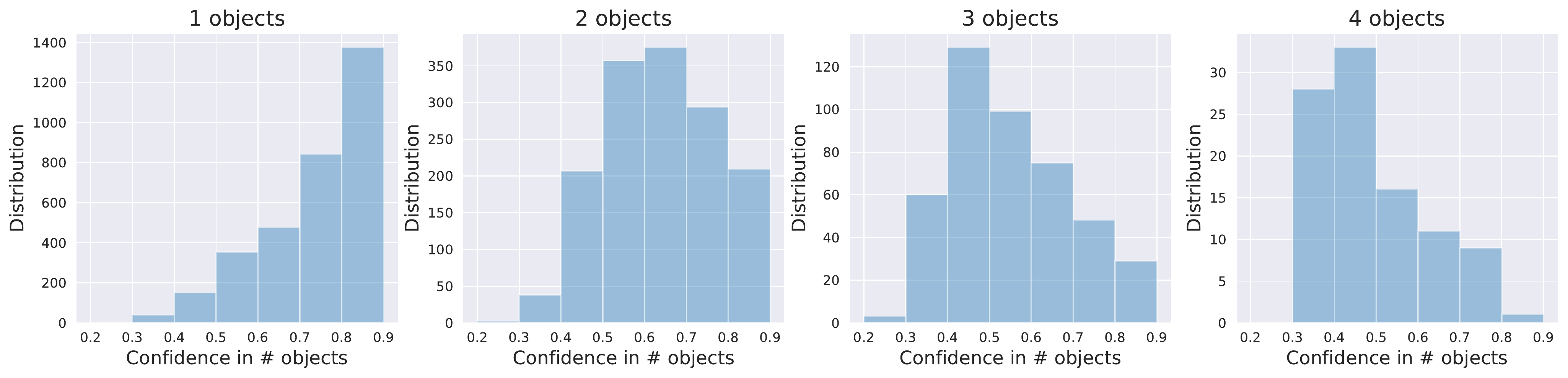}
	\caption{Annotator agreement for multi-object images. Recall that we 
	determine the number of objects in an image based on a majority vote over 
	annotators. Here, we define ``confidence'' as the fraction of annotators 
	that make up that majority, relative to the total number of annotators 
	shown the image (cf. Section~\ref{sec:followup}). We 
	visualize the distribution of annotator confidence, as a function of the 
	number of image objects.
		}
	\label{fig:count_confidence_hist}
\end{figure}

Additional (random) multi-object images are shown in 
Figure~\ref{fig:multi_examples}. We observe that 
annotators tend to agree on 
the 
number of objects present---see Figure~\ref{fig:count_confidence_hist}.

\clearpage

\begin{figure}[!h]
	\centering
	\centering
	\includegraphics[width=1\textwidth]{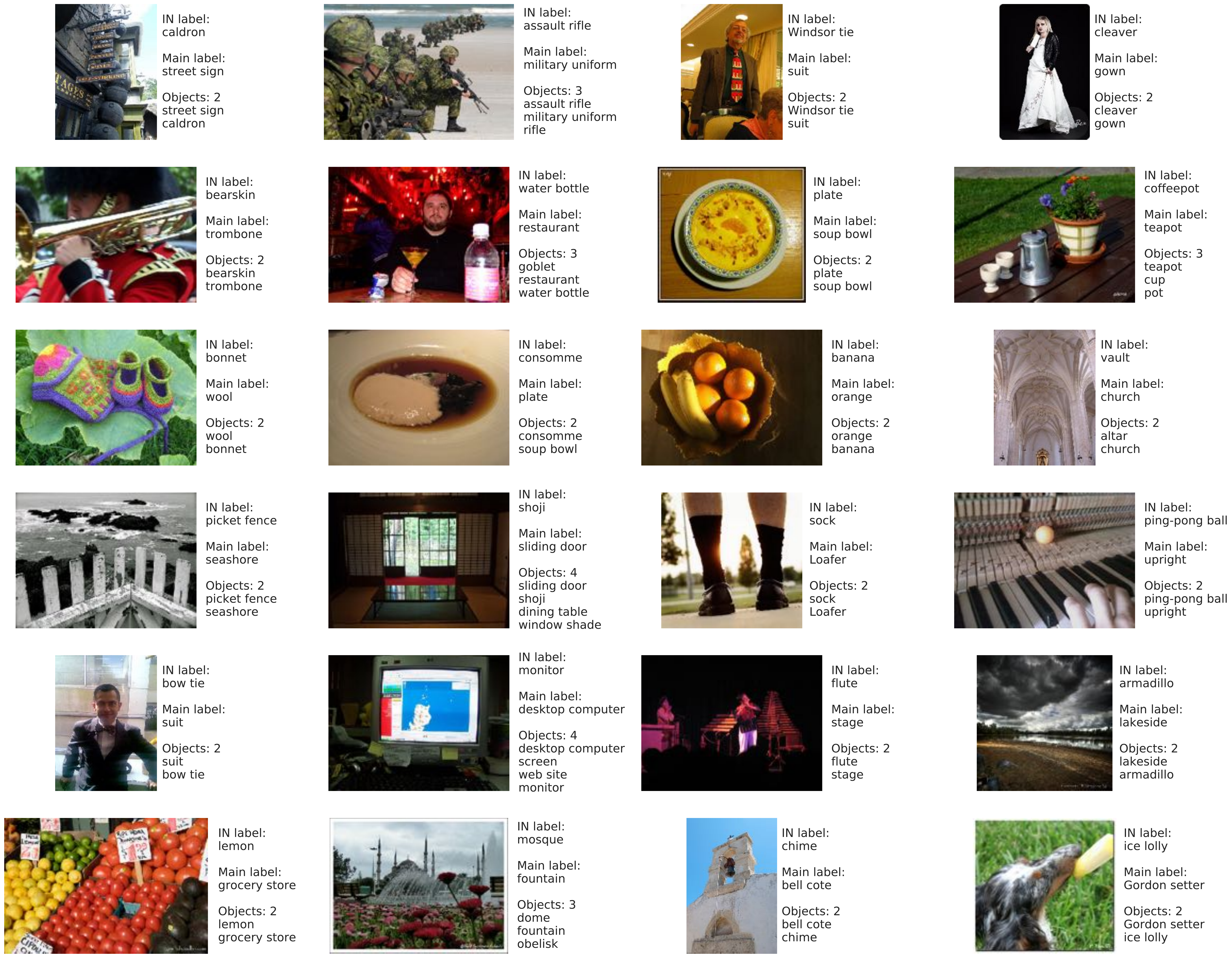}
	\caption{Sample images for which the main label as per 
		annotators 
		differs from the ImageNet label.}
	\label{fig:disagreement}
\end{figure}

\begin{figure}[!h]
	\centering
	\includegraphics[width=1\textwidth]{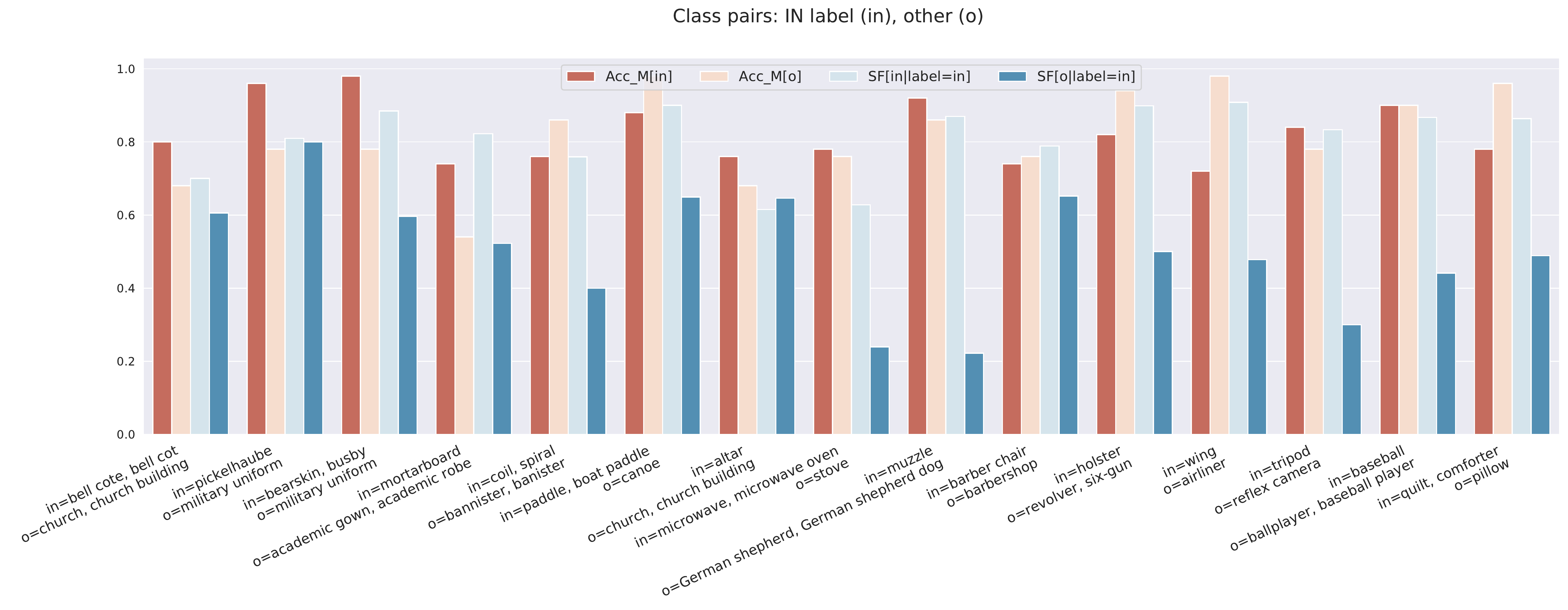}
	\caption{Classes for which human main label frequently differs from the 
	ImageNet 
		label. Here, although annotator selection frequency for the ImageNet 
		label is high, the 
		selection frequency for \emph{another} class---which humans 
		consider to be the main label---is also consistently high. Models still 
		predict the ImageNet label, possibly by 
		picking up on distinctive 
		features of the 
		objects.}
	\label{fig:bearskin_bars}
\end{figure}

In 
Figure~\ref{fig:disagreement}, we visualize 
instances where the ImageNet label does not match with what annotators 
deem to be the ``main object''.
In many of these cases, we find that models perform at predicting the 
ImageNet label, even though the images contain other, more salient objects 
according to annotators---Figure~\ref{fig:bearskin_bars}.
 
\begin{figure}[!h]
	\centering
	\includegraphics[width=1\textwidth]{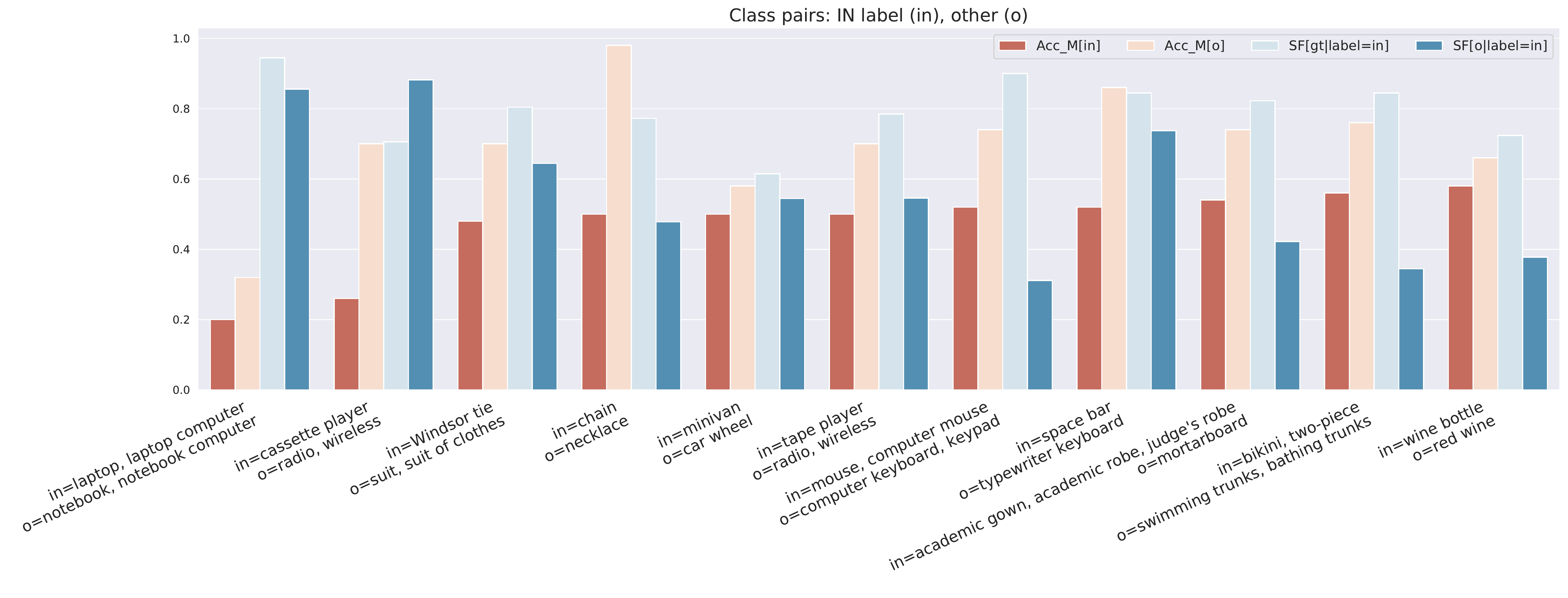}
	\caption{Classes where model accuracy is consistently low due to frequent object 
	co-occurences: in these cases, an object from the
		ImageNet class frequently co-occurs with (or is a sub-part of) objects from another 
		class. Here, 
		models 
		seem to be unable to disambiguate the two classes completely, and thus perform 
		poorly on one/both classes. Note that human selection frequency for the 
		ImageNet class is high, indicating that an object from that class is 
		present in the image.}
	\label{fig:cooccurrence_bars}
\end{figure}

In Figure~\ref{fig:class_coocc}, we visualize pairs of ImageNet classes that 
frequently co-occur in images---e.g., suit
and tie, space bar and keyboard.
For some of these co-occurring class pairs, model performance as a whole, seems to be
poor (cf. Figures~\ref{fig:cooccurrence_bars})---possibly because of an 
inherent significant overlap in the image distributions of the two classes. 

\begin{figure}[!h]
	\centering
	\includegraphics[width=0.9\textwidth]{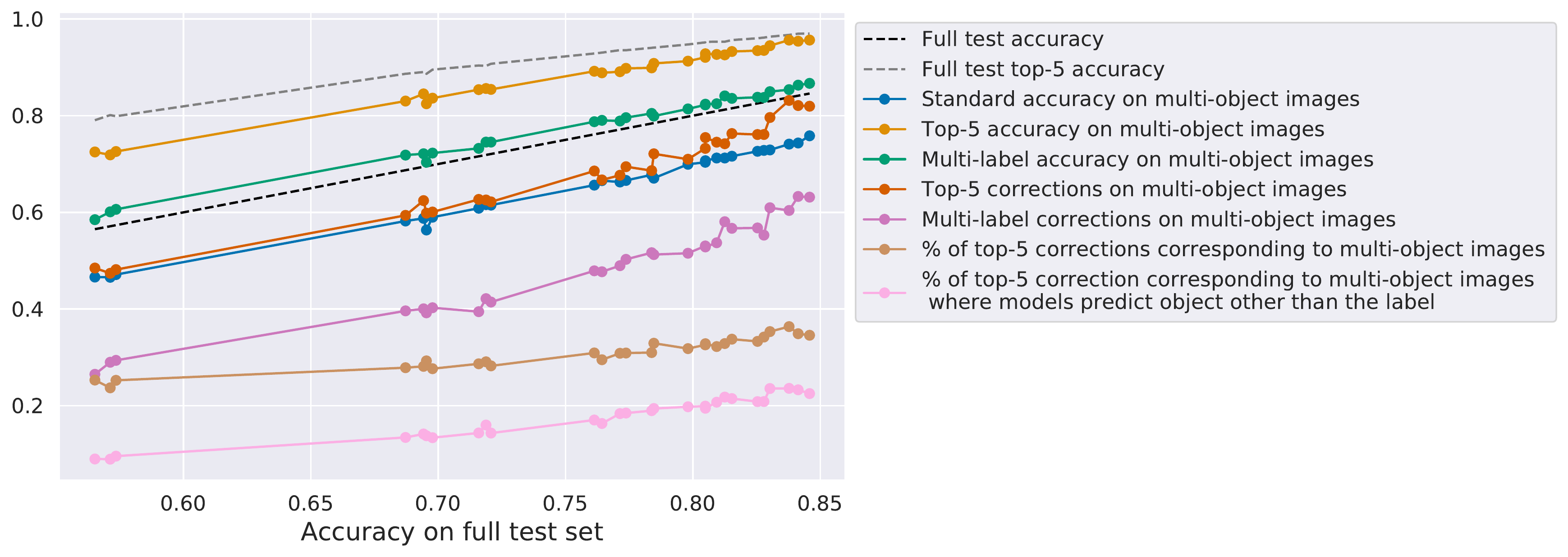}
	\caption{A closer look at top-5 accuracy: we visualize top-1, top-5 and multi-label (cf. 
	Section~\ref{sec:multi}) on multi-object images in ImageNet. 
	We also measure the fraction of top-5 corrections (ImageNet label is not top model 
	prediction, but is among top 5) that correspond to multi-object confusions---wherein 
	the ImageNet label and top prediction belong to distinct image objects as 
	per human annotators.
	We see that although top-5 accuracy is much higher than top-1, even for multi-object 
	images, it may be overestimating model performance. In particular, a relatively small 
	fraction of top-5 corrections actually correspond to the aforementioned multi-object 
	images.
}
	\label{fig:top5_acc}
\end{figure}

\paragraph{Top-5 accuracy in the multi-label context.} The issue of label
ambiguity that can arise in multi-object images was noted by the creators
of the ILSVRC challenge~\citep{russakovsky2015imagenet}. To tackle this 
issue, they 
proposed evaluating models based on top-5 accuracy.
Essentially, a model is deemed correct if any of the top 5 predicted labels  
match the ImageNet label.
We find that model top-5 accuracy is much higher than top-1 (or even our 
notion of multi-label) accuracy on multi-object images---see 
Figure~\ref{fig:top5_acc}.
However, a priori, it 
is not obvious whether this increase is actually because of adjusting for  
model 
confusion between distinct objects.  
In fact, one would expect that properly accounting for such
multi-object confusions should yield numbers similar to (and not markedly 
higher 
than) top-1 accuracy on single-object images.

To get a better understanding of this, we visualize the fraction of \emph{top-5 
corrections}---images for which the ImageNet label was not the top prediction of the 
model, but was in the top 5---that correspond to \emph{different objects} in the image.  
Specifically, we only consider images where the top model prediction and 
ImageNet label were selected by annotators as: (a) present in the image and 
(b) corresponding to different objects.  
We observe that  the fraction of  top-5 corrections that correspond to multi-object 
images is relatively small---about  $20\%$ for more recent models.
This suggests that top-5 accuracy may be overestimating model performance and, in a 
sense, masking model errors on single objects.
Overall, these findings highlight the need for designing better performance metrics that 
reflect the underlying dataset structure.

\clearpage
\subsection{Bias in label validation} 
\label{app:selection_frequency}

\begin{figure}[!h]
\begin{subfigure}{0.75\textwidth}
	\centering
	\includegraphics[width=1\textwidth]{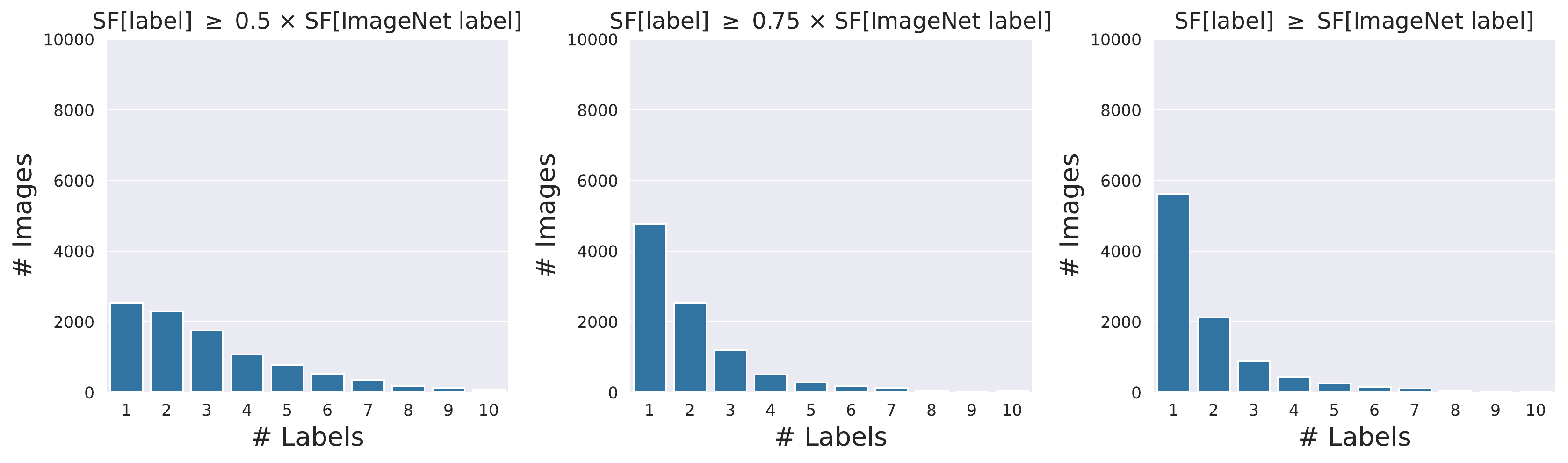}
	\caption{}
\end{subfigure}
\begin{subfigure}{0.2\textwidth}
	\centering
	\includegraphics[width=1\textwidth]{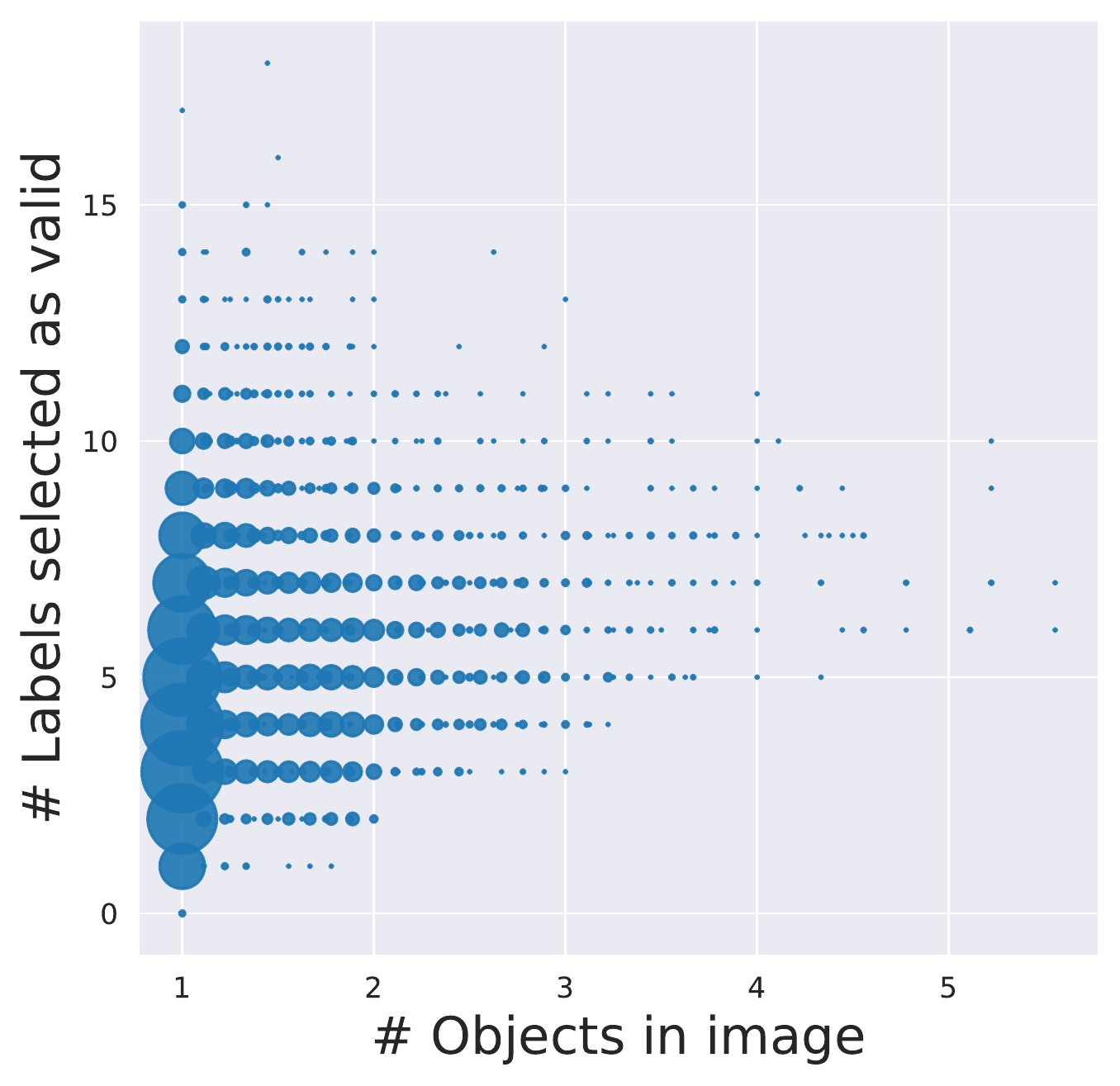}
	\caption{}
\end{subfigure}
	\caption{Effect of subsampling annotator population (5 instead of 9 annotators): (a)
	Number of labels annotators consider valid determined 
	based on the selection frequency of a label relative to that of the ImageNet label.
	Even in this annotator subpopulation, for $>$70\% of images, 
	another label is still selected at least half as often as they select the ImageNet label
		(\emph{leftmost}).
	(b) Number of labels that at least one (of five) annotators selected as
	valid for an image (cf.\ Section~\ref{sec:contains}) versus the
	number of objects in the image (cf.\ Section~\ref{sec:followup}).
	(Dot size is proportional to the number of images in each 2D bin.)
	Even when annotators consider the image as containing only a single object,
	they often select multiple labels as valid. }
	\label{fig:confident_bs}
\end{figure}

\paragraph{Potential biases in selection frequency estimates.}
In the course of obtaining fine-grained image annotations, we collect selection 
frequencies for several potential image labels (including the ImageNet label) 
using the \contains{} task (cf. Section~\ref{sec:contains}).
Recall however, that during the ImageNet creation process, every image was
already validated w.r.t. the ImageNet label (also via the \contains{} task) by a
different pool of annotators, and only images with high selection frequency
actually made it into the dataset.
This fact will result in a \emph{bias} for our new selection frequency
measurements~\cite{engstrom2020identifying}.
At a high level, if we measure a low selection frequency for the ImageNet label
of an image, it is more likely that we are observing an underestimate, rather
than the actual selection frequency being low. 
In order to understand whether this bias significantly affects our
findings, we reproduce the relevant plots in Figure~\ref{fig:confident_bs} using
only a subset of workers (this should exacerbate the bias allowing us to detect
it).
We find however, that the difference is quite small, not changing any of the
conclusions.
Moreover, since most of our analysis is based on the per-image annotation task
for which this specific bias does not apply, we can effectively ignore it
in our study.

\begin{figure}[!h]
	\centering
	\includegraphics[width=.7\textwidth]{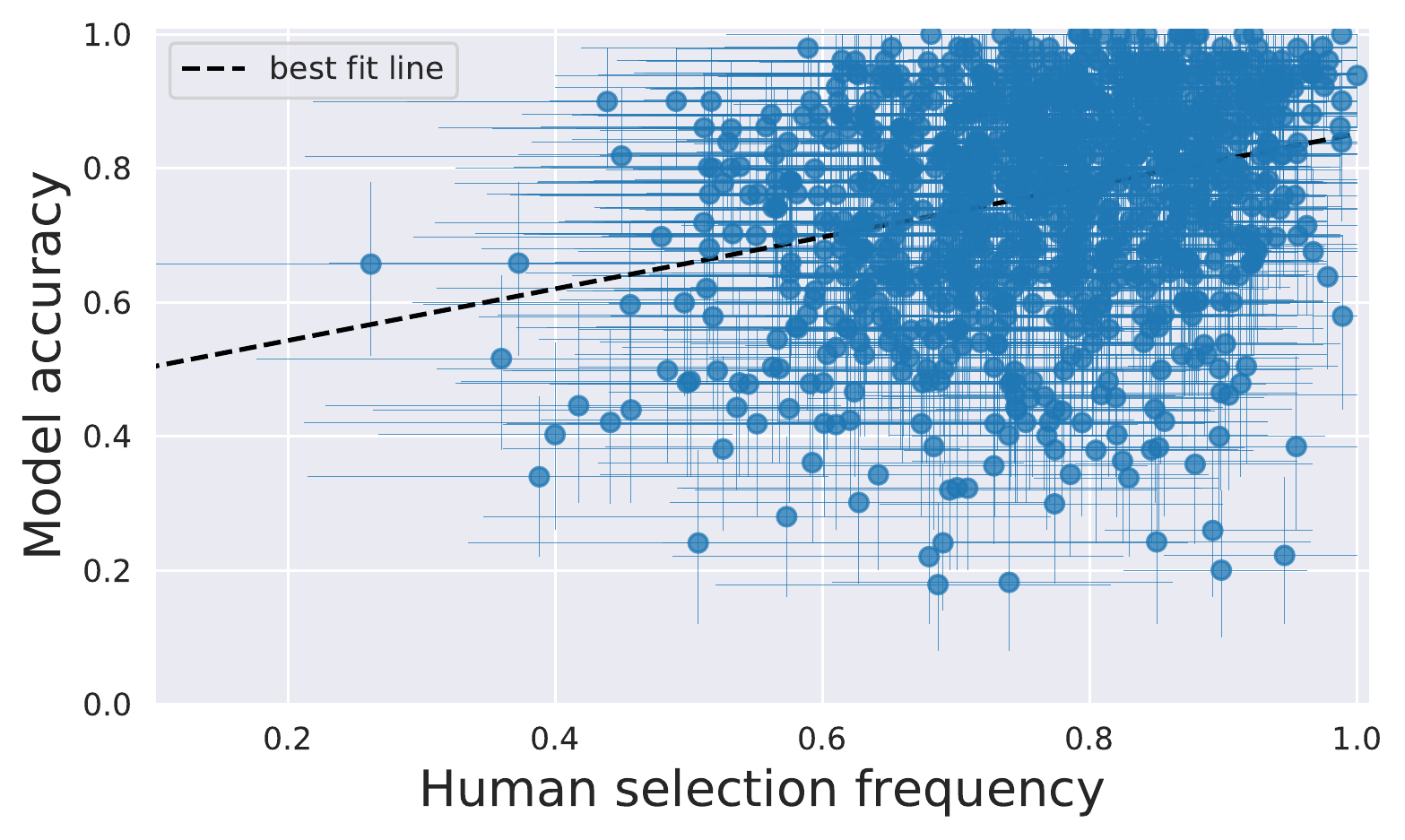}
	\caption{Relationship between per-class selection frequency and model (ResNet-50)
	accuracy. We observe that while in general higher selection frequency does 
	correlate with better  accuracy, this is not uniformly true. This suggests that 
	using 
	selection frequency as a proxy for how easy an image is in terms of
    \task{} 
	may not be perfect---especially if the dataset contains fine-grained classes or 
	multi-object 
	images.}
	\label{fig:scatter}
\end{figure}

\paragraph{Selection frequency and model accuracy.}
In typical dataset creation pipelines, selection frequency (and other similar 
metrics) is 
often used to filter out images that are not easily recognizable by humans. We now consider how 
the selection frequency of a class relates to model accuracy on that class---see 
Figure~\ref{fig:scatter}. While we see that, in general, classes with high human selection 
frequency are also easier for models to classify, this is not uniformly true---in particular, 
there seem to be classes that significantly deviate from this trend. 
Given our observations about the nuances in ImageNet data, this may be justified---for 
instance, multi-object images would have high human selection frequency 
for the ImageNet label, but may not be easy to learn.

\clearpage

\subsection{Mislabeled examples} 
\label{app:mislabeled}
\begin{figure}[!h]
	\centering
	\includegraphics[width=0.96\textwidth]{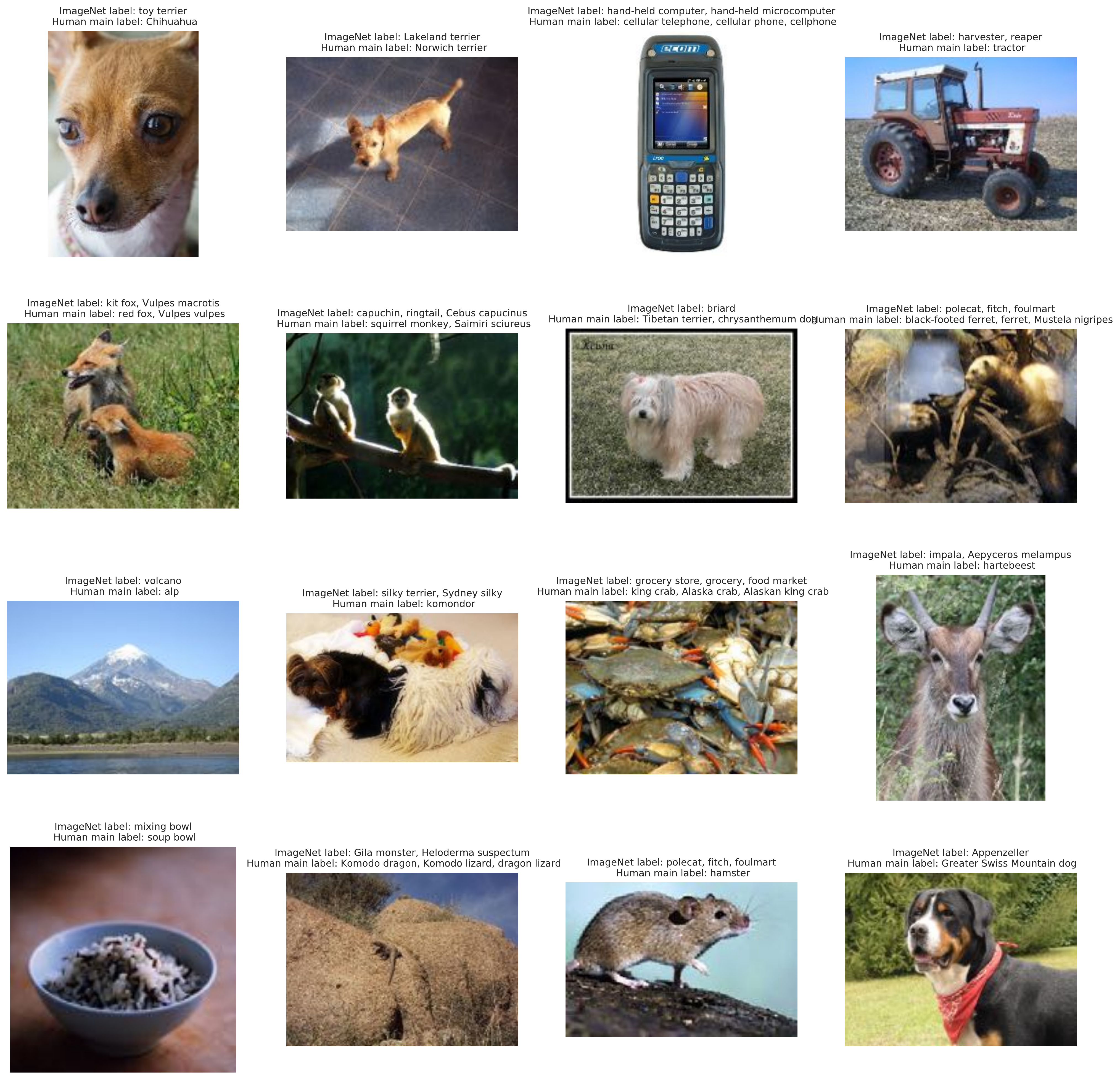}
	\caption{Possibly mislabeled images: human selection frequency for the 
	ImageNet label is $0$ (cf. Section~\ref{sec:contains}). Also 
	depicted is the label most frequently selected by the annotators as 
	contained in the image (\emph{sel}).}
	\label{fig:mis_10k}
\end{figure}

\begin{figure}[!h]
	\centering
	\includegraphics[width=1\textwidth]{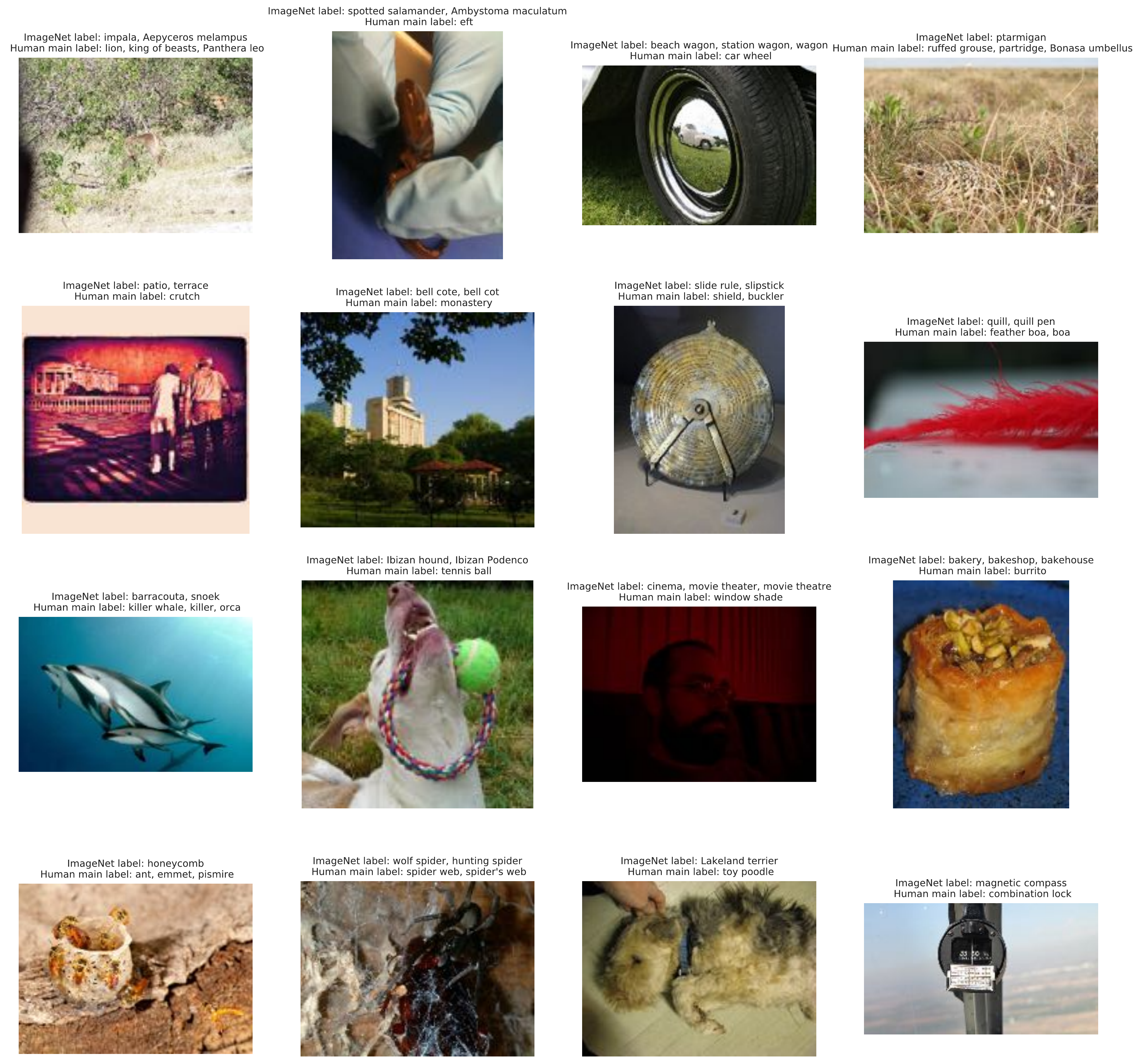}
	\caption{Possibly mislabeled  images: ImageNet label is not 
	selected by any of the annotators during fine-grained image annotation process 
	described in Section~\ref{sec:followup_specs}. 
		Also 
		shown in the title is label that was most frequently selected by the annotators as 
		denoting the main object in the image (\emph{sel}).}
	\label{fig:mis_fu}
\end{figure}

In the course of our human studies in Section~\ref{sec:contains}, we also identify a set of 
possibly mislabeled ImageNet images. Specifically, we find images for which:

\begin{itemize}
	\item Selection frequency for the ImageNet 
	label is $0$, i.e., no annotator selected the  label to be contained in the 
	image (cf. Section~\ref{sec:contains}). We identify 150 (of 10k) such 
	images--- cf. Appendix 
	Figure~\ref{fig:mis_10k} for examples.
	\item The ImageNet label was not selected at all (for 
	any object) during the detailed image annotation phase in 
	Section~\ref{sec:followup_specs}. We identify 119 (of 10k) such images--- 
	cf. Appendix Figure~\ref{fig:mis_fu} for examples.
\end{itemize}

\clearpage

\subsection{Confusion Matrices}
\label{app:confusion}

The $(i, j)th$ entry of the human/model confusion matrix denotes how often an image 
with ImageNet label $i$ is predicted as class $j$. 
We consider the model prediction to simply be the top-1 label. We determine 
the ``human prediction'' in two ways, as the class: (1) with 
highest annotator selection frequency in the \contains{} task,  and (2) 
which is the most likely choice for the main label in the \classify{} task. In 
Appendix 
Figure~\ref{fig:conf_r50} we compare model and human confusion matrices 
(for both notions of human prediction). Unless otherwise specified, all 
other confusion 
matrices in the paper are based on the 
main label (using method (2)).

\begin{figure*}[!h]
	\includegraphics[width=1\textwidth]{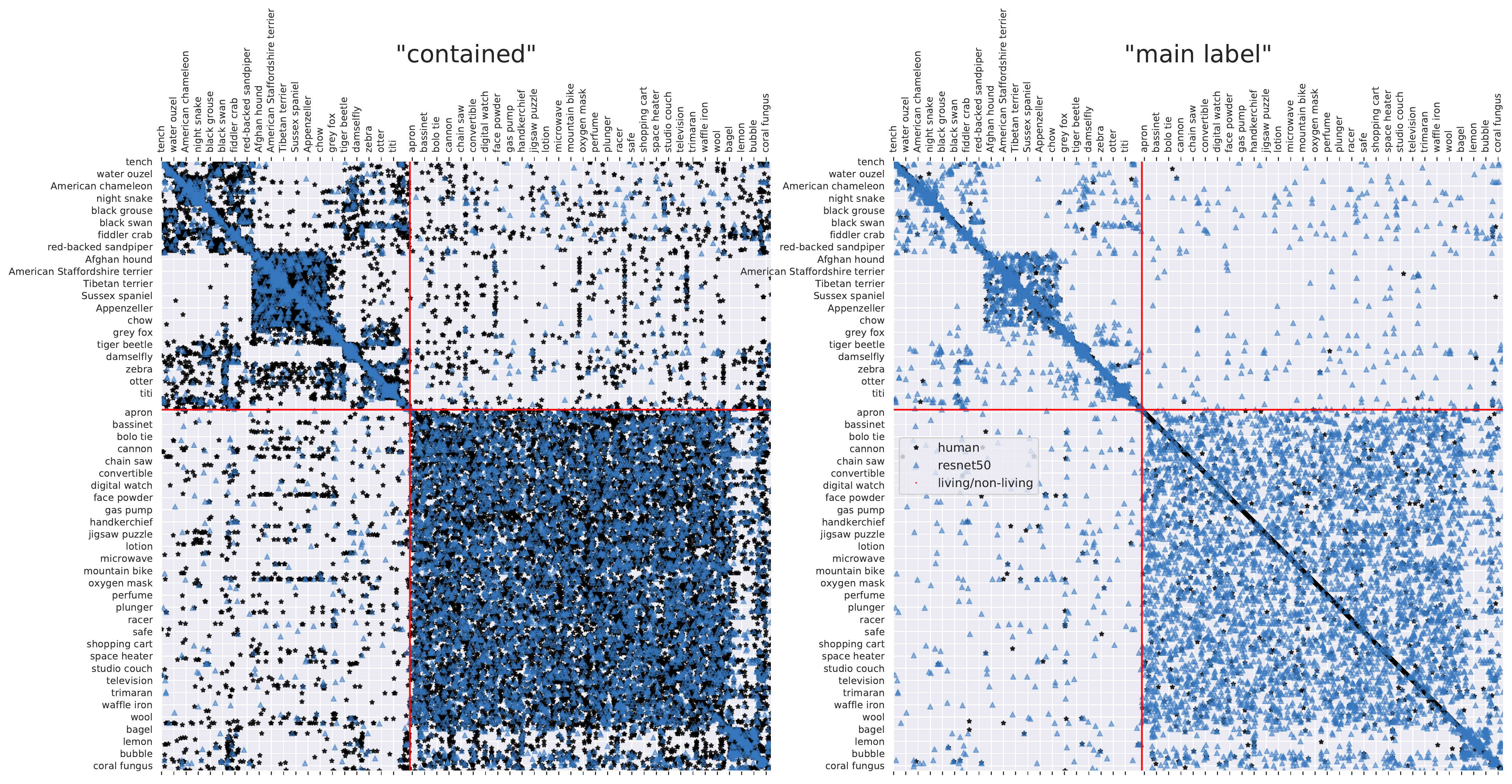}
	\caption{Comparison of model (\emph{blue}; ResNet-50) and human (\emph{black}) 
	confusion matrices for all 1000 ImageNet classes. At a high-level, model and human 
	confusion patterns seem somewhat aligned although humans seem 
	to be particularly worse when it comes to fine-grained classes. }
	\label{fig:conf_r50}
\end{figure*}

We compare the inter-superclass confusion matrices for humans and various models in 
Appendix Figure~\ref{fig:high_level_conf_models}. We see that as models get better, their 
confusion patterns look similar to human annotators. Moreover, as we saw previously in 
Figure~\ref{fig:conf_r50}, there are blocks of superclasses where confusion seems 
particularly prominent---likely due to frequent object co-occurrences (cf. 
Figure~\ref{fig:cooccurrence}). We find that intra-superclass confusions are more 
prominent for humans---see Appendix Figure~\ref{fig:high_level_conf_b7_intra}). This is in 
line with our findings from  Section~\ref{sec:contains}, where we observe that human 
annotators often select  multiple labels for an image, even when they think it contains one 
object.


\begin{figure*}[!h]
	\begin{subfigure}[b]{1\textwidth}
		\includegraphics[width=\textwidth]{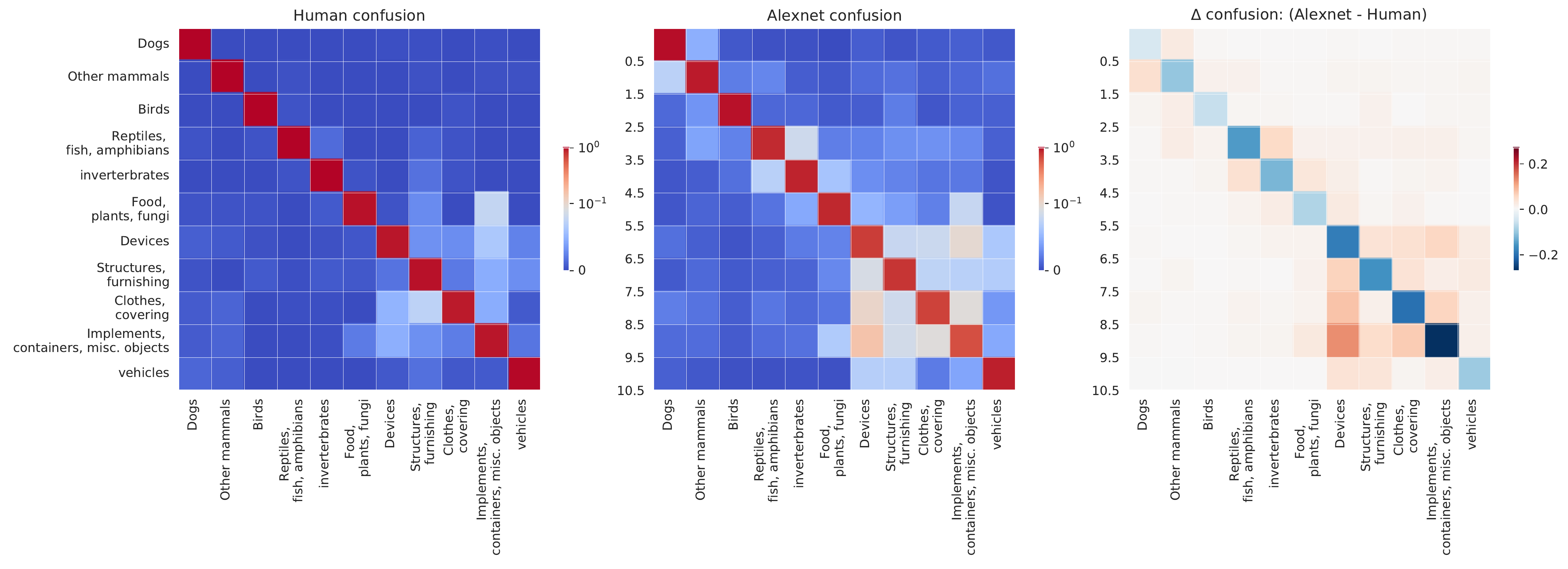}
		\caption{AlexNet}
	\end{subfigure}
	\begin{subfigure}[b]{1\textwidth}
		\includegraphics[width=\textwidth]{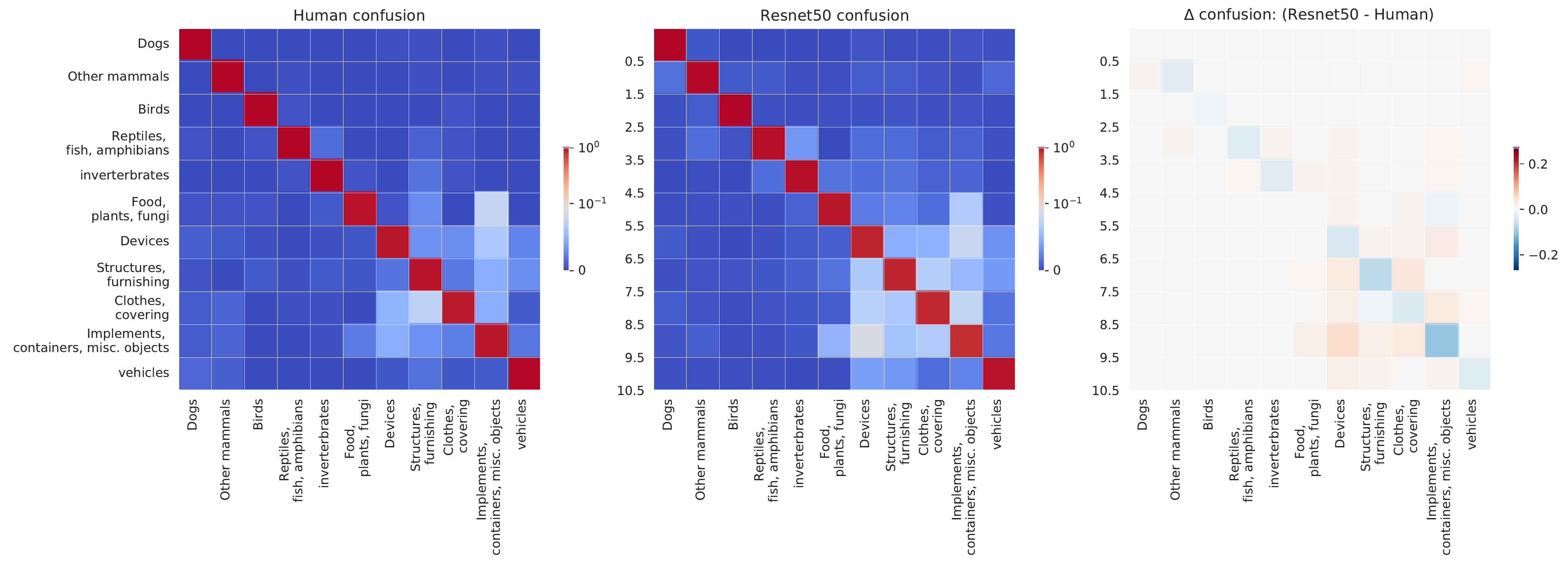}
		\caption{ResNet-50}
	\end{subfigure}
	\begin{subfigure}[b]{1\textwidth}
		\includegraphics[width=\textwidth]{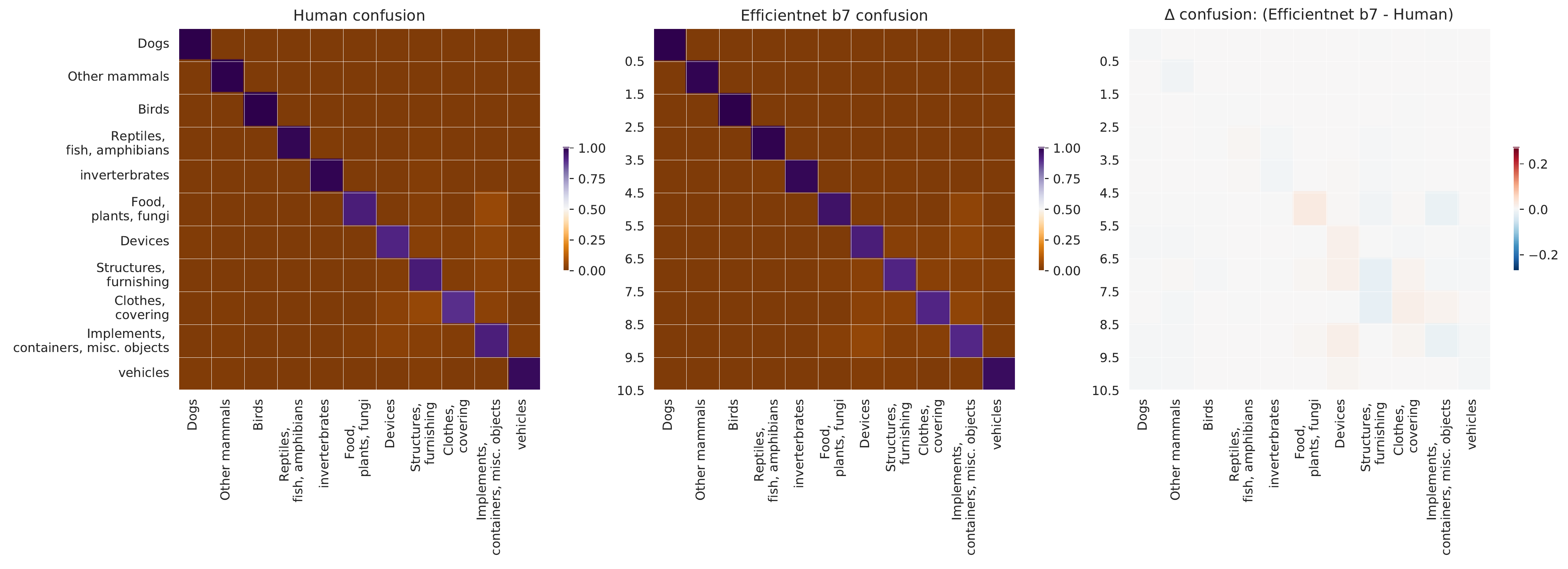}
		\caption{EfficientNet B7}
	\end{subfigure}
	\caption{Inter-superclass confusion matrices for models and humans. We observe that 
	as models get more accurate, their confusion also tends to align better with humans. In 
	fact, more recent models seem to be better than humans, especially when it comes 
	to fine-grained classes. Moreover, we see that for these models, the confusion patterns 
	seem to mirror the object co-occurrences in Figure~\ref{fig:cooccurrence}. This 
	suggests that part of errors of current models may stem from (legitimate) 
	confusions due to multi-label images. }
	\label{fig:high_level_conf_models}
\end{figure*}

\begin{figure*}[!t]
	\begin{center}
		\begin{subfigure}[b]{1\textwidth}
			\centering
			\includegraphics[width=\textwidth]{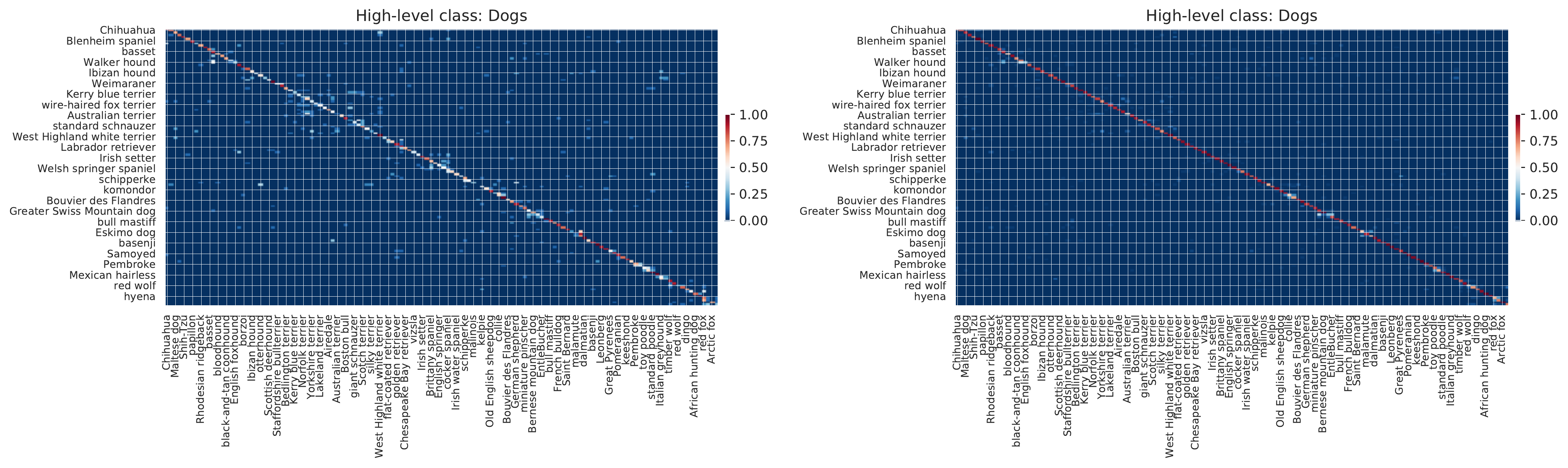}
		\end{subfigure}
		\vspace*{2pt}
		\begin{subfigure}[b]{1\textwidth}
			\centering
			\includegraphics[width=\textwidth]{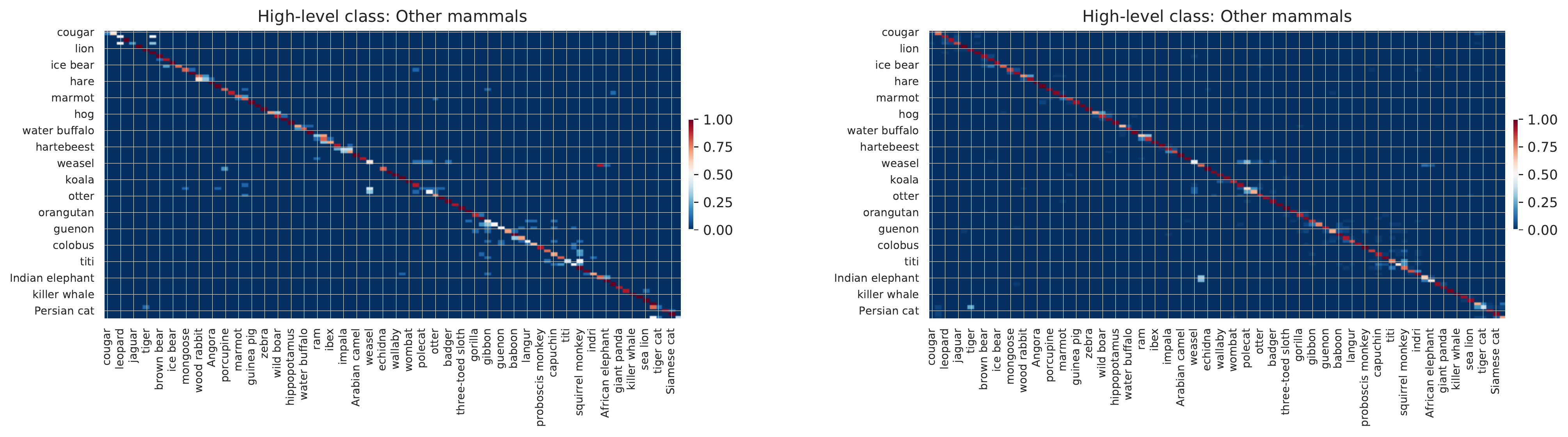}
			
		\end{subfigure}
		\begin{subfigure}[b]{1\textwidth}
			\centering
			\includegraphics[width=\textwidth]{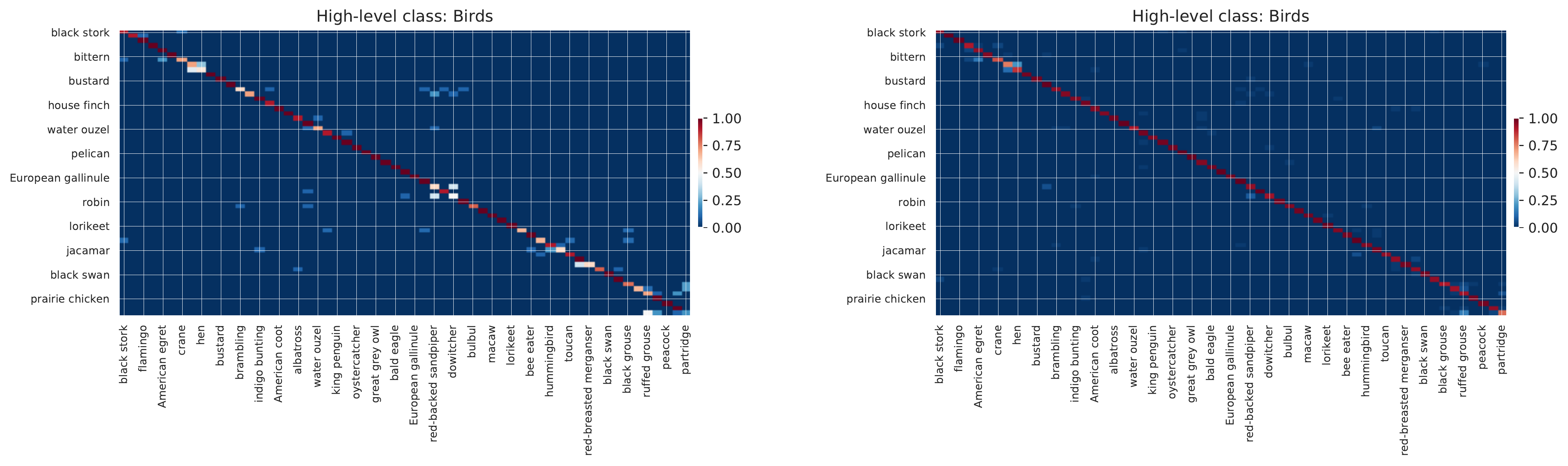}
		\end{subfigure}
		\vspace*{2pt}
		\begin{subfigure}[b]{1\textwidth}
			\centering
			\includegraphics[width=\textwidth]{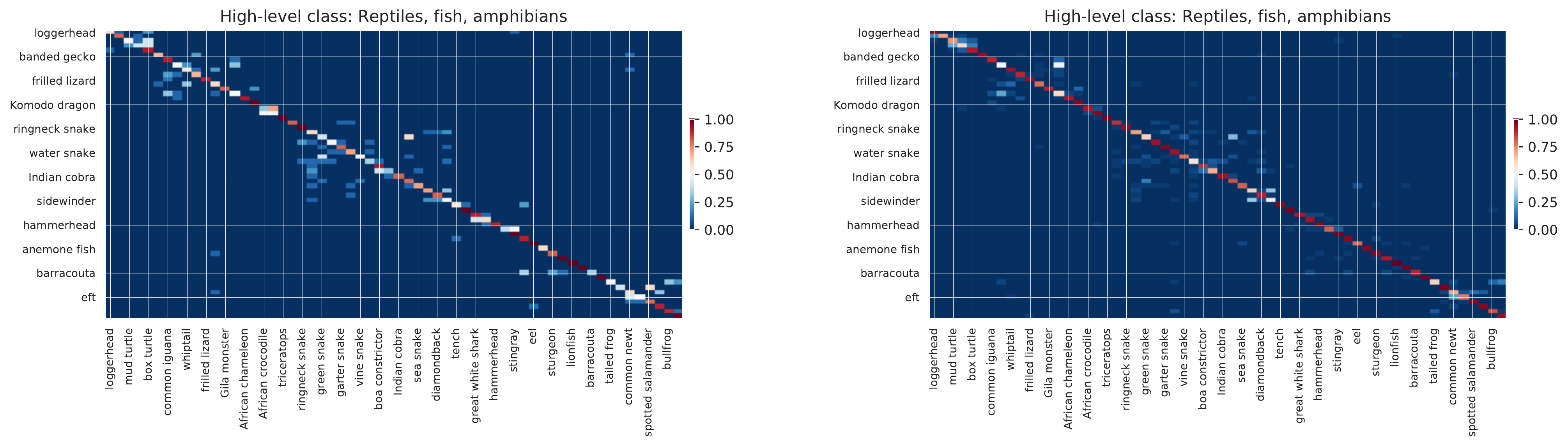}
			
		\end{subfigure}
	\end{center}
\end{figure*}

\begin{figure*}[!t]\ContinuedFloat
	\begin{center}
		\begin{subfigure}[b]{1\textwidth}
			\centering
			\includegraphics[width=\textwidth]{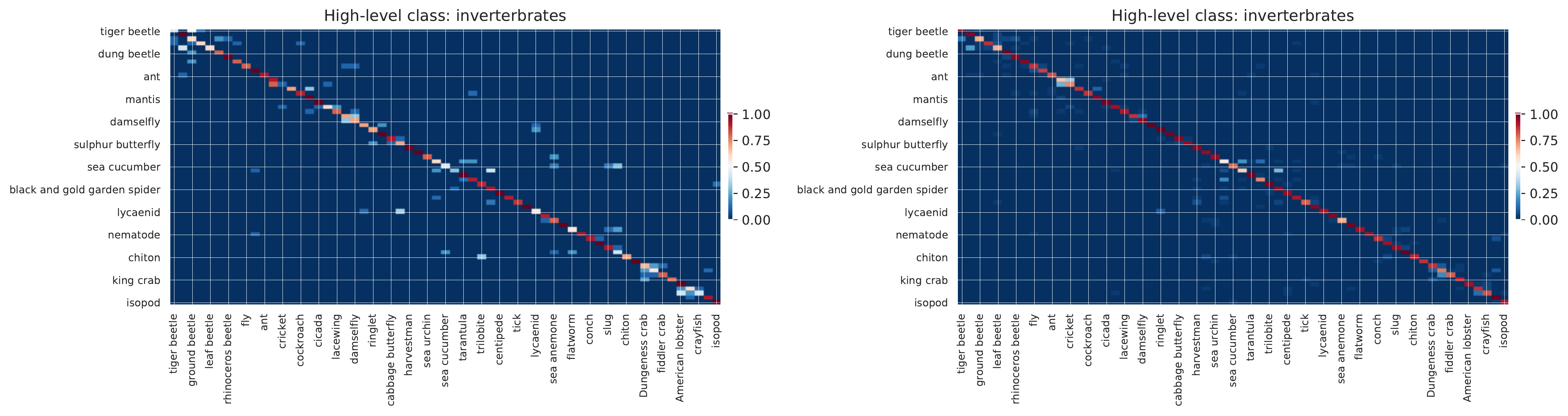}
		\end{subfigure}
		\vspace*{2pt}
		\begin{subfigure}[b]{1\textwidth}
			\centering
			\includegraphics[width=\textwidth]{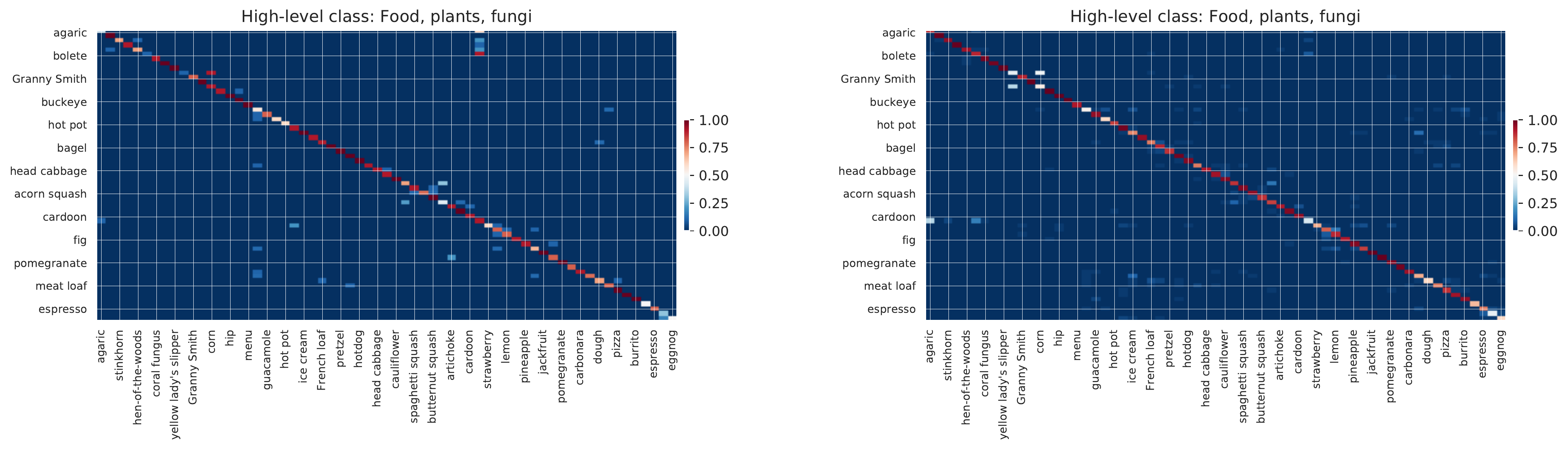}
			
		\end{subfigure}
		\begin{subfigure}[b]{1\textwidth}
			\centering
			\includegraphics[width=\textwidth]{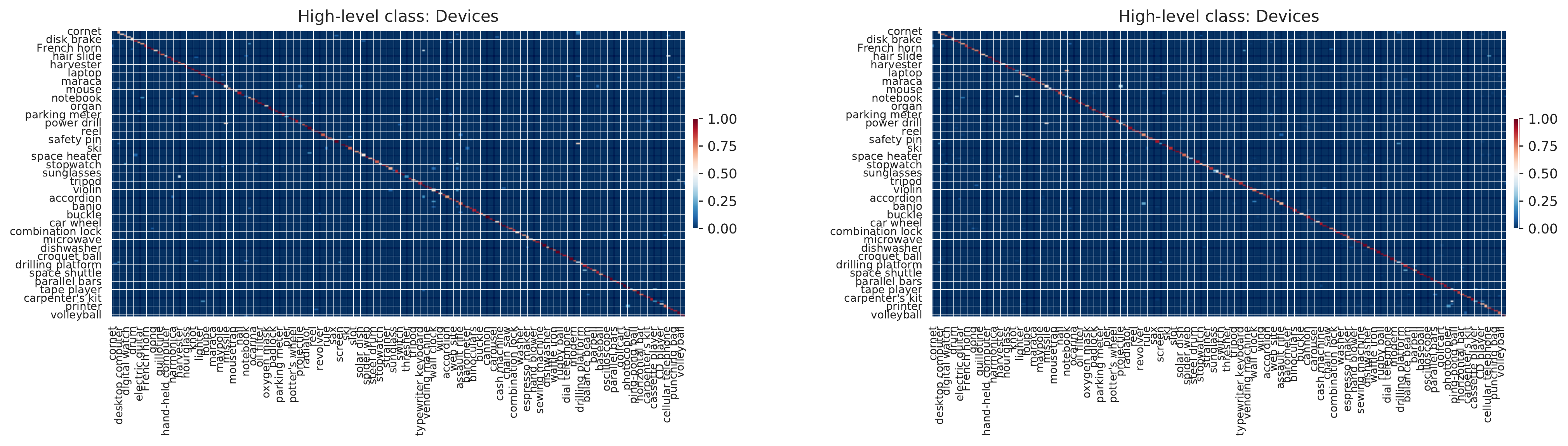}
		\end{subfigure}
		\vspace*{2pt}
		\begin{subfigure}[b]{1\textwidth}
			\centering
			\includegraphics[width=\textwidth]{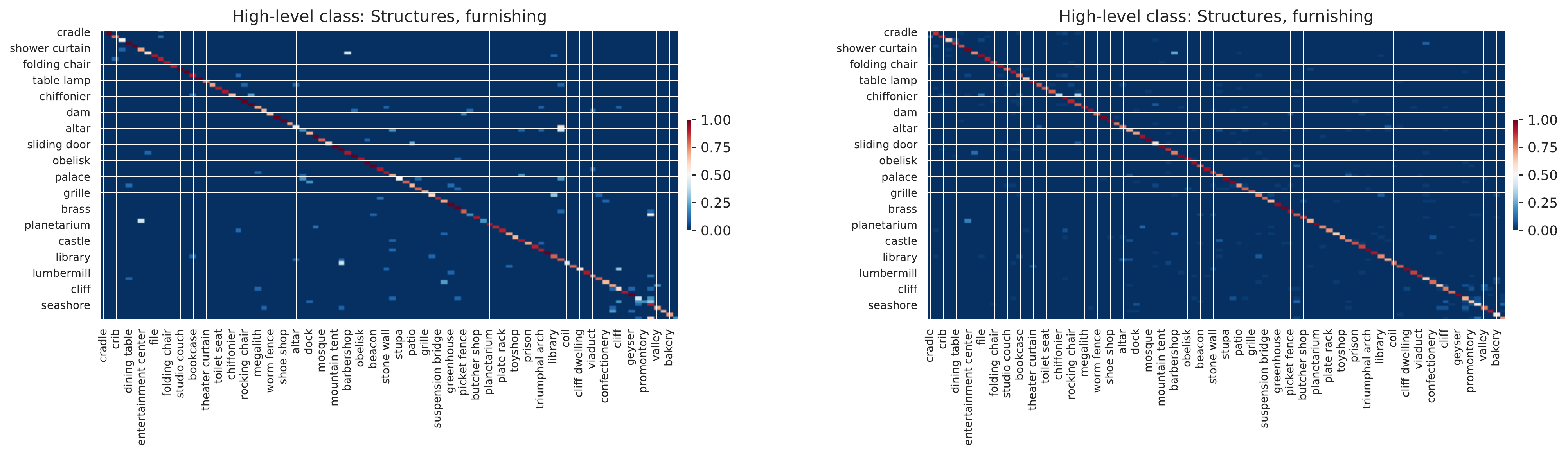}
			
		\end{subfigure}
	\end{center}
\end{figure*}

\begin{figure*}[!t]\ContinuedFloat
	\begin{center}
		\begin{subfigure}[b]{1\textwidth}
			\centering
			\includegraphics[width=\textwidth]{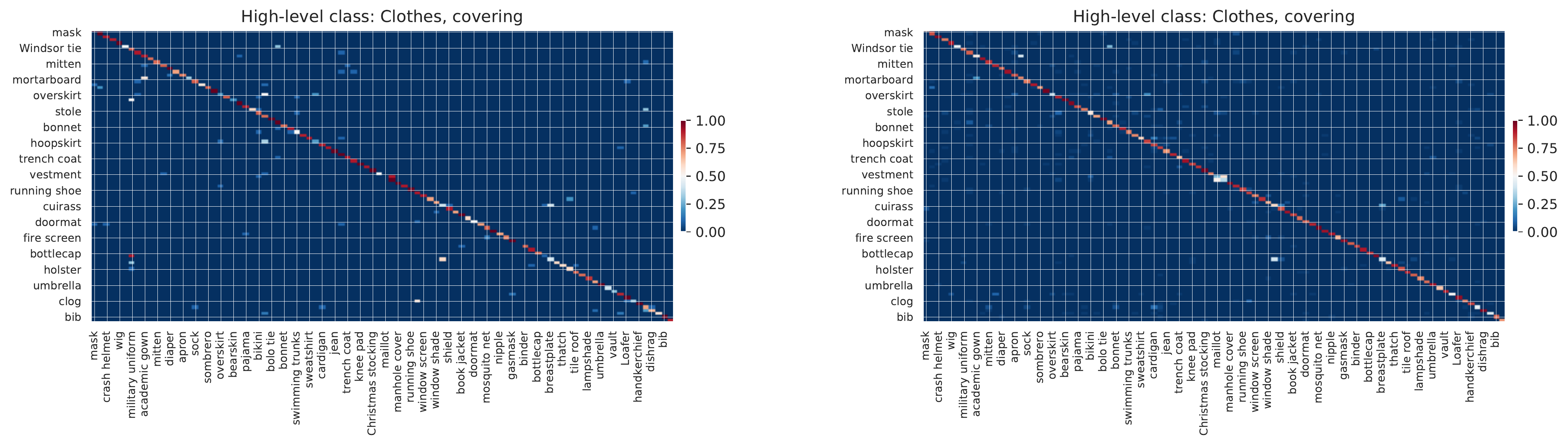}
		\end{subfigure}
		\vspace*{2pt}
		\begin{subfigure}[b]{1\textwidth}
			\centering
			\includegraphics[width=\textwidth]{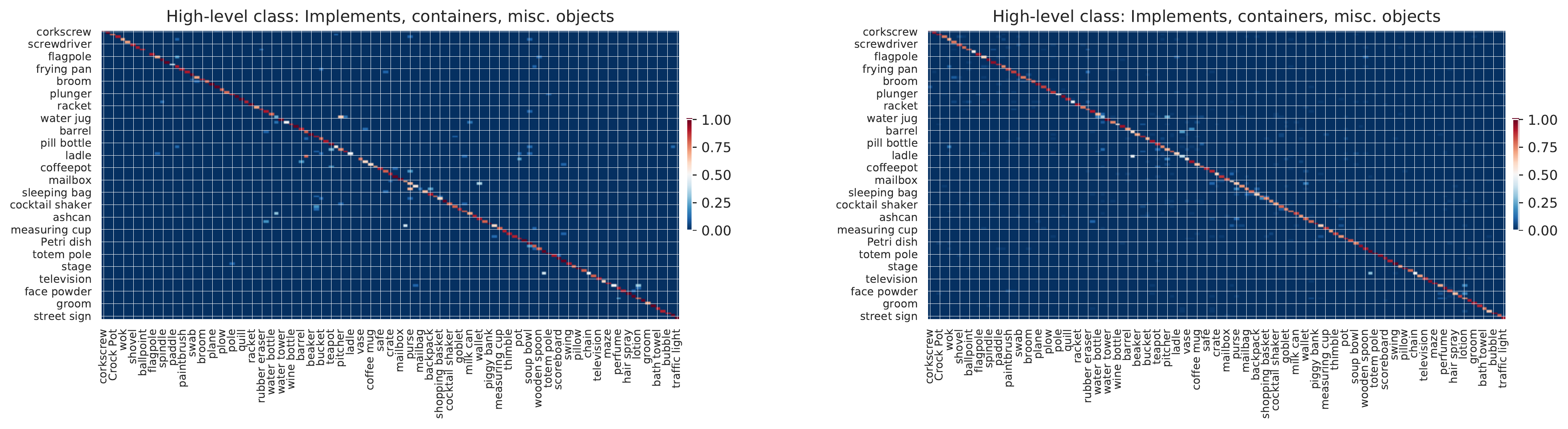}
			
		\end{subfigure}
		
		\begin{subfigure}[b]{1\textwidth}
			\centering
			\includegraphics[width=\textwidth]{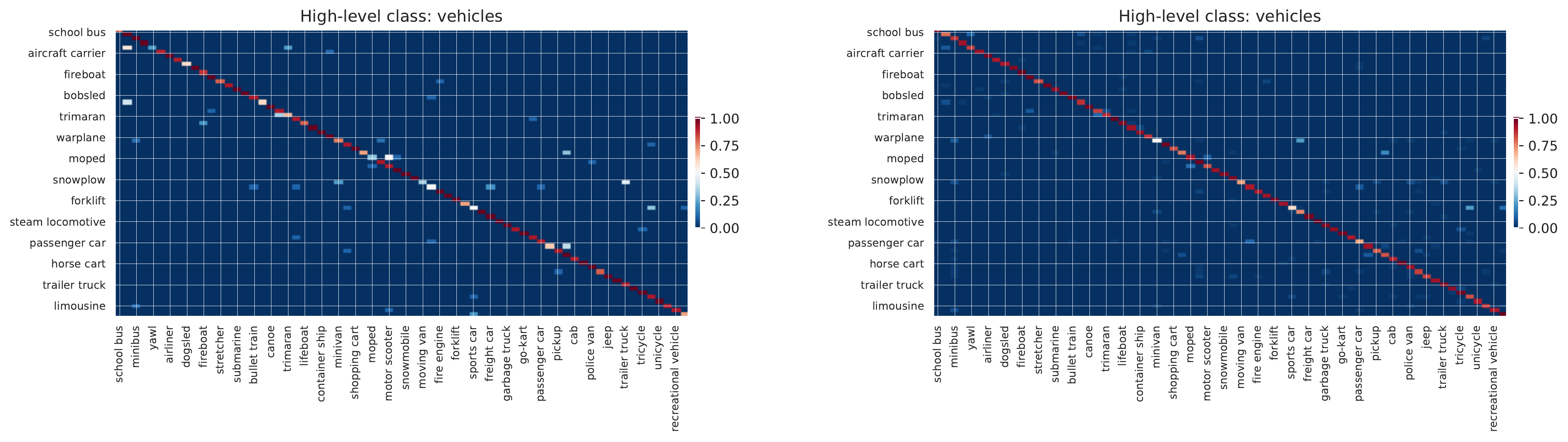}
			
		\end{subfigure}
	\end{center}
	\caption{Intra-superclass confusion matrices for (\emph{left}) humans 
	and (\emph{right}) an EfficientNet B7.}
	\label{fig:high_level_conf_b7_intra}
\end{figure*}

%


\end{document}